%% file: paper.tex
\documentclass[twoside,11pt]{article}

\usepackage{jmlr2e}

\usepackage[utf8]{inputenc}

\usepackage{amsmath}
\usepackage{url}
\usepackage{subfigure}
\usepackage{floatflt}
\usepackage{color}

\newcommand{\eq}{\!=\!}
\renewcommand{\neq}{\!\ne\!}
\newcommand{\given}{\;|\;}

\ShortHeadings{How to Explain Individual Classification Decisions}{D. Baehrens, T. Schroeter, S. Harmeling, M. Kawanabe, K. Hansen, K.-R. Müller}
\firstpageno{1}

\begin{document}
\title{How to Explain Individual Classification Decisions}

\author{%
\name David Baehrens\thanks{The first three authors contributed
equally.} \email baehrens@cs.tu-berlin.de \\
\name Timon Schroeter$^*$        \email timon@cs.tu-berlin.de \\
\addr{Technische Universit\"at Berlin} \\
\addr{Franklinstr. 28/29, FR 6-9}    \\
\addr{10587 Berlin, Germany}         \\
\name Stefan Harmeling$^*$      \email stefan.harmeling@tuebingen.mpg.de \\
\addr{MPI for Biological Cybernetics} \\
\addr{Spemannstr. 38}               \\
\addr{72076 T\"ubingen, Germany}        \\
\name Motoaki Kawanabe      \email motoaki.kawanabe@first.fraunhofer.de \\
\addr{Fraunhofer Institute FIRST.IDA} \\
\addr{Kekulestr.7}                    \\
\addr{12489 Berlin, Germany}          \\
\addr{and} \\
\addr{Technische Universit\"at Berlin} \\
\addr{Franklinstr. 28/29, FR 6-9}    \\
\addr{10587 Berlin, Germany}         \\
\name Katja Hansen          \email khansen@cs.tu-berlin.de \\
\name Klaus-Robert M\"uller \email klaus-robert.mueller@tu-berlin.de \\
\addr{Technische Universit\"at Berlin} \\
\addr{Franklinstr. 28/29, FR 6-9}    \\
\addr{10587 Berlin, Germany}         \\
}
\editor{Carl Edward Rasmussen}
\maketitle
\begin{abstract}%
After building a classifier with modern tools of machine learning we
typically have a black box at hand that is able to predict well for
unseen data. Thus, we get an answer to the question {\em what} is
the most likely label of a given unseen data point.  However, most methods will provide no answer \emph{why} the model predicted the particular label for a single instance and what features were most influential for that particular instance.  The only method that is currently able to provide such explanations are decision trees.  This paper proposes a procedure which (based on a set of assumptions) allows to explain the decisions of \emph{any} classification method.
\end{abstract}

\begin{keywords}
explaining, nonlinear, black box model, kernel methods, Ames mutagenicity
\end{keywords} 

\input{introduction}

\input{definition}

\input{theory_gpc}
\input{estimating}

\input{iris_knn}

\input{usps_svm}

\input{tox_gpc}

\input{related_work}

\input{discussion}

\input{conclusion}

\input{acknowledgments}

\def\thesection{\Alph{section}}
\setcounter{section}{0}
\input{appendix}

\bibliography{ida,t_misc,explaining,tox,harmeling}

\end{document}

%% file: introduction.tex
\section{Introduction}
\label{sec:introduction}

Automatic nonlinear classification is a common and powerful tool in
data analysis.  Machine learning research has created methods that are
practically useful and that can classify unseen data after being
trained on a limited training set of labeled examples.

Nevertheless, most of the algorithms do not {\em explain} their decision. 
However in practical data analysis it is essential to obtain an
instance based explanation, i.e. we would like to
gain an understanding what input features made the nonlinear machine
give its answer for each individual data point.

Typically, explanations are provided jointly for all instances of the
training set, for example feature selection methods 
(including Automatic Relevance Determination) find out
which inputs are salient for a good generalization \citep*[see for a review][]{GuyEli03}. While this can
give a coarse impression about the global usefulness of each input dimension,
it is still an ensemble view and does not provide an answer on an
instance basis.\footnote{This point is illustrated in Figure \ref{fig:toy_gpc} (Section \ref{sec:definition}).
Applying feature selection methods to the training set (a) will
lead to the (correct) conclusion that both dimensions are equally important
for accurate classification.
As an alternative to this ensemble view, one may ask:
Which features (or combinations thereof) are most influential in the
vicinity of each particular instance. 
As can be seen in Figure \ref{fig:toy_gpc}~(c), the answer depends on where the respective
instance is located. On the hypotenuse and at the corners of the triangle,
both features contribute jointly, whereas along each of the remaining two edges 
the classification depends almost completely on just one of the features.}
In the neural network literature also solely an
ensemble view was taken in algorithms like input pruning
\citep*[e.g.][]{Bis95,LeCunBottou98}.
The only classification which does provide individual explanations are decision trees \citep*[e.g.][]{Hastie01}.

This paper proposes a simple framework that provides local explanation vectors applicable to \emph{any} classification method in order to help understanding
prediction results for single data instances. The local explanation yields the features being relevant for the prediction at the very points of interest in the data space and is able to spot local peculiarities which are neglected in the global view e.g. due to cancellation effects.

The paper is organized as follows: We define local explanation vectors as class probability gradients in Section \ref{sec:definition} and give an illustration for Gaussian Process Classification (GPC).
Some methods output a prediction without a direct probability interpretation. 
For these we propose in Section \ref{sec:estimating} a way to estimate local explanations. 
In Section \ref{sec:iris_knn} we will apply our methodology to learn distinguishing properties of Iris flowers by estimating explanation vectors for a k-NN classifier applied to the classic Iris data set.  Section \ref{sec:usps_svm} will discuss how our approach applied to a SVM classifier allows us to explain how digits "two" are distinguished from digit "8" in the USPS data set.
In Section \ref{sec:tox_gpc} we discuss a more real-world application scenario where the proposed explanation capabilities prove useful in drug discovery: Human experts regularly decide how to modify existing lead compounds in order to obtain new compounds with improved properties. Models capable of explaining predictions can help in the process of choosing promising modifications. Our automatically generated explanations match with chemical domain knowledge about toxifying functional groups of the compounds in question.
Section \ref{sec:related_work} contrasts our approach with related work and Section \ref{sec:discussion} discusses characteristic properties and limitations of our approach, before we conclude the paper in Section \ref{sec:conclusion}.

%% file: definition.tex
\section{Definitions of Explanation Vectors}
\label{sec:definition}




In this Section we will give definitions for our approach of local explanation vectors in the 
classification setting. We start with a theoretical definition for multi-class Bayes classification and then give a specialized definition being more practical for the binary case.

For the multi-class case, suppose we are given data points $x_1, \ldots,x_n\in\Re^d$ with labels
$y_1, \ldots, y_n\in\{1, \ldots, C\}$ and we intend to learn a function
that predicts the labels of unlabeled data points.  Assuming that
the data could be modeled as being IID-sampled from some unknown
joint distribution $P(X, Y)$, in theory, we can define the Bayes
classifier,
\begin{align*}
  g^*(x) = \arg\min_{c\in\{1,\ldots,C\}} P(Y\neq c\given X\eq x)
\end{align*}
which is optimal for the 0-1 loss function \citep*[see][]{DevGyoLug96}.

For the Bayes classifier we define the {\em explanation vector} of a
data point $x_0$ to be the derivative with respect to $x$ at $x=x_0$
of the conditional probability of $Y\neq g^*(x_0)$ given $X=x$, or
formally,
\begin{definition}
  \label{eq:5}
  \[\zeta(x_0) := \left.\frac{\partial}{\partial x} \;P(Y\neq
  g^*(x_0)\given X\eq x)\right\rvert_{x=x_0}\]
\end{definition}
Note that $\zeta(x_0)$ is a $d$-dimensional vector just like $x_0$ is.
The classifier $g^*$ partitions the data space $\Re^d$ into up to $C$
parts on which $g^*$ is constant. We assume that the conditional 
distribution $P(Y=c \given X=x)$ is first-order differentiable w.r.t.
$x$ for all classes $c$ and over the entire input space. For instance,
the assumption holds, if
$P(X\eq x\given Y\eq c)$ is for all $c$ first-order differentiable
in $x$ and the supports of the class densities overlap around the boarder
for all the neighboring pairs in the partition by the Bayes
classifier. 
The vector $\zeta(x_0)$
defines on each of those parts a vector field that characterizes the
flow away from the corresponding class.  Thus entries in $\zeta(x_0)$
with large absolute values highlight features that will influence the
class label decision of $x_0$.  A positive sign of such an entry
implies that increasing that feature would lower the probability that
$x_0$ is assigned to $g^*(x_0)$.  
Ignoring the orientations of the
explanation vectors, $\zeta$ forms a continuously changing
(orientation-less) vector field along which the class labels change.
This vector field lets us {\em locally} understand the Bayes
classifier.

We remark that $\zeta(x_0)$ becomes a zero vector, e.g. 
when $P(Y\neq g^*(x_0)\given X\eq x)\rvert_{x=x_0}$ is equal to
one in some neighborhood of $x_0$.
Our explanation vector fits well to probabilistic classifiers 
such as Gaussian Process Classification (GPC), where the conditional
distribution $P(Y=c \given X=x)$ is usually not completely flat 
in some regions. 
In the case of deterministic classifiers, despite of this issue,
Parzen window estimators with appropriate widths (Section
\ref{sec:estimating}) can
provide meaningful explanation vectors for many samples in practice 
(see also Section \ref{sec:discussion}).

For the case of binary classification we directly define local explanation vectors
as local gradients of the probability function $p(x) = P(Y = 1 \given X = x)$ of the learned model for the positive class.

So
for a probability function $p~:~\Re^d \rightarrow [0,1]$ of a classification model learned from examples $\{(x_1, y_1), \dots, (x_n, y_n)\} \in \Re^d \times \{-1,+1\}$ the explanation vector for a classified test point $x_0$ is the local gradient of $p$ at $x_0$:

\begin{definition}
  \label{def:ev_cls}
  \[\eta_p(x_0) := \nabla p(x) \rvert_{x=x_0}\]
\end{definition}

By this definition the explanation $\eta$ is again a $d$-dimensional vector just like the test point $x_0$ is. 
The sign of each of its individual entries indicates whether the prediction would increase or decrease when the corresponding feature of $x_0$ is increased locally and each entry's absolute value give the amount of influence in the change in prediction. 
As a vector $\eta$ gives the direction of the steepest ascent from the test point to higher probabilities for the positive class. For binary classification the negative version $-\eta_p(x_0)$ indicates the changes in features needed to increase the probability for the negative class which may be especially useful for $x_0$ predicted in the positive class.



For an example we apply Definition \ref{def:ev_cls} to model predictions learned by Gaussian Process Classification (GPC), see \citet*{Rasmussenbook06}. GPC is used here for three reasons:\\ (i) In our real-world application we are interested in classifying data from drug discovery, which is an area where Gaussian processes have proven to show state-of-the-art performance, see e.g.~\citet*{Obrezanova_Csanyi_2007_gp,SchSchMikLaaSueGanHeiMue07a,SchSchMikLaaSueGanHeiMue07b,SchSchMikLaaSueGanHeiMue07c,SchSchMikLauLaaSueGanHeiMue06,SchSchMikHanLaaLieReiHeiMue08,Obrezanova_Gola_2008_gp}. It is natural to expect a model with high prediction accuracy on a complex problem to capture relevant structure of the data which is worth explaining and may give domain specific insights in addition to the values predicted. For an evaluation of the explaining capabilities of our approach on a complex problem from chemoinformatics see Section \ref{sec:tox_gpc}. \\ (ii) GPC does model the class probability function used in Definition \ref{def:ev_cls} directly. For other classification methods such as Support Vector Machines which do not provide a probability function as its output in Section \ref{sec:estimating} we give an example for an estimation method starting from Definition \ref{eq:5}.\\
(iii) The local gradients of the probability function can be calculated analytically for differentiable kernel as we discuss next.

Let $\overline{f}(x) = \sum_{i=1}^n  \alpha_i k(x,x_i)$ be a GP model
trained on sample points $x_1, \dots, x_n \in \Re^d$ where $k$ is a
kernel function and $\alpha_i$ are the learned weights of each sample
point. For a test point $x_0 \in \Re^d$ let $\mathrm{var}_f(x_0)$ be
the variance of $f(x_0)$ under the GP posterior for $f$. 
Because the posterior cannot be calculated analytically for GP
classification models, we used an approximation by expectation
propagation (EP) \cite*[][]{KusRas05}. In the case of the probit likelihood term defined by the
error function, the probability for being of the positive class $p(x_0)$ can be
computed easily from this approximated posterior as
\begin{align*}
  p(x_0) = \frac{1}{2}\mathrm{erfc}\left(\frac{-\overline{f}(x_0)}{\sqrt{2}*\sqrt{1+\mathrm{var}_f(x_0)}}\right), 
\end{align*}
where $\mathrm{erfc}$ denotes the complementary error function
\cite*[see Equation 6 in][]{SchSchMikHanLaaLieReiHeiMue08}.

Then the local gradient of $p(x_0)$ is given by

\begin{align*}
& \nabla p(x)\rvert_{x=x_0}\\
=& \nabla \left.\frac{1}{2}\mathrm{erfc}\left(\frac{-\overline{f}(x)}{\sqrt{2}*\sqrt{1+\mathrm{var}_f(x)}}\right) \right\rvert_{x=x_0} \\
=& \nabla \left. \frac{1}{2}\left(1 - \mathrm{erf}\left(\frac{-\overline{f}(x)}{\sqrt{2}*\sqrt{1+\mathrm{var}_f(x)}}\right) \right) \right\rvert_{x=x_0} \\
=& -\frac{1}{2} \nabla \left.\mathrm{erf}\left(\frac{-\overline{f}(x)}{\sqrt{2}*\sqrt{1+\mathrm{var}_f(x)}}\right) \right\rvert_{x=x_0} \\
=& -\frac{\exp\left(\frac{-\overline{f}(x_0)^2}{2(1+\mathrm{var}_f(x_0))}\right)}{\sqrt{\pi}}  \nabla \left. \left(\frac{-\overline{f}(x)}{\sqrt{2}*\sqrt{1+\mathrm{var}_f(x)}}\right) \right\rvert_{x=x_0} \\
=& -\frac{\exp\left(\frac{-\overline{f}(x_0)^2}{2(1+\mathrm{var}_f(x_0))}\right)}{\sqrt{\pi}}  \left(-\frac{1}{\sqrt{2}} \nabla \left. \left(\frac{\overline{f}(x)}{\sqrt{1+\mathrm{var}_f(x)}}\right) \right\rvert_{x=x_0} \right)\\
=& \frac{\exp\left(\frac{-\overline{f}(x_0)^2}{2(1+\mathrm{var}_f(x_0))}\right)}{\sqrt{2\pi}}  \left(\frac{\nabla \overline{f}(x) \rvert_{x=x_0}}{\sqrt{1+\mathrm{var}_f(x_0)}} \right. 
 + \left. \overline{f}(x_0) \left(\nabla\left.\mathrm{var}_f(x) \right\rvert_{x=x_0} * -\frac{1}{2}\left( 1 + \mathrm{var}_f(x_0) \right)^{-\frac{3}{2}} \right) \right)\\
=& \frac{\exp\left(\frac{-\overline{f}(x_0)^2}{2(1+\mathrm{var}_f(x_0))}\right)}{\sqrt{2\pi}} \left(\frac{\nabla \overline{f}(x) \rvert_{x=x_0}}{\sqrt{1+\mathrm{var}_f(x_0)}} \right. 
 - \left.\frac{1}{2}\frac{\overline{f}(x_0)}{\left(1+\mathrm{var}_f(x_0)\right)^{\frac{3}{2}}}\nabla\mathrm{var}_f(x) \rvert_{x=x_0} \right).
\end{align*}
As a kernel function choose e.g. the RBF-kernel $k(x_0,x_1) = \exp(-w(x_0 - x_1)^2)$, which has the derivative $(\partial/\partial x_{0,j})k(x_0,x_1) = -2w \exp(-w(x_0 - x_1)^2) (x_{0,j} - x_{1,j})$ for $j \in \{1, \dots, d\}$. Then the elements of the local gradient $\nabla \overline{f}(x) \rvert_{x=x_0}$ are

\begin{align*}
\frac{\partial \overline{f}}{\partial x_{0,j}} & = -2w \sum_{i=1}^n  \alpha_i \exp(-w(x_0 - x_i)^2) (x_{0,j} - x_{i,j}) \;\;\text{ for } j \in \{1, \dots, d\}.
\end{align*}

For $\mathrm{var}_f(x_0) = k(x_0,x_0) - k_*^T(K + \Sigma)^{-1} k_*$ the derivative is given by\footnote{Here $k_* = (k(x_0,x_1),\dots,k(x_0,x_n))^T$ is the evaluation of the kernel function between the test point $x_0$ and every training point. $\Sigma$ is the diagonal matrix of the variance site parameter. For details see \citet*[Chapter 3]{Rasmussenbook06}}
\begin{align*}
\nabla \mathrm{var}_f(x) \rvert_{x=x_0} 
 = \frac{\partial \mathrm{var}_f}{\partial x_{0,j}} 
 = \left(\frac{\partial}{\partial x_{0,j}}k(x_0,x_0)\right) - 2*k_*^T(K + \Sigma)^{-1}\frac{\partial}{\partial x_{0,j}}k_*
 \;\;\text{ for } j \in \{1, \dots, d\}.
\end{align*}


\begin{figure}[ht]
\centering
\subfigure[Object]{\includegraphics[width=0.49\textwidth, keepaspectratio]{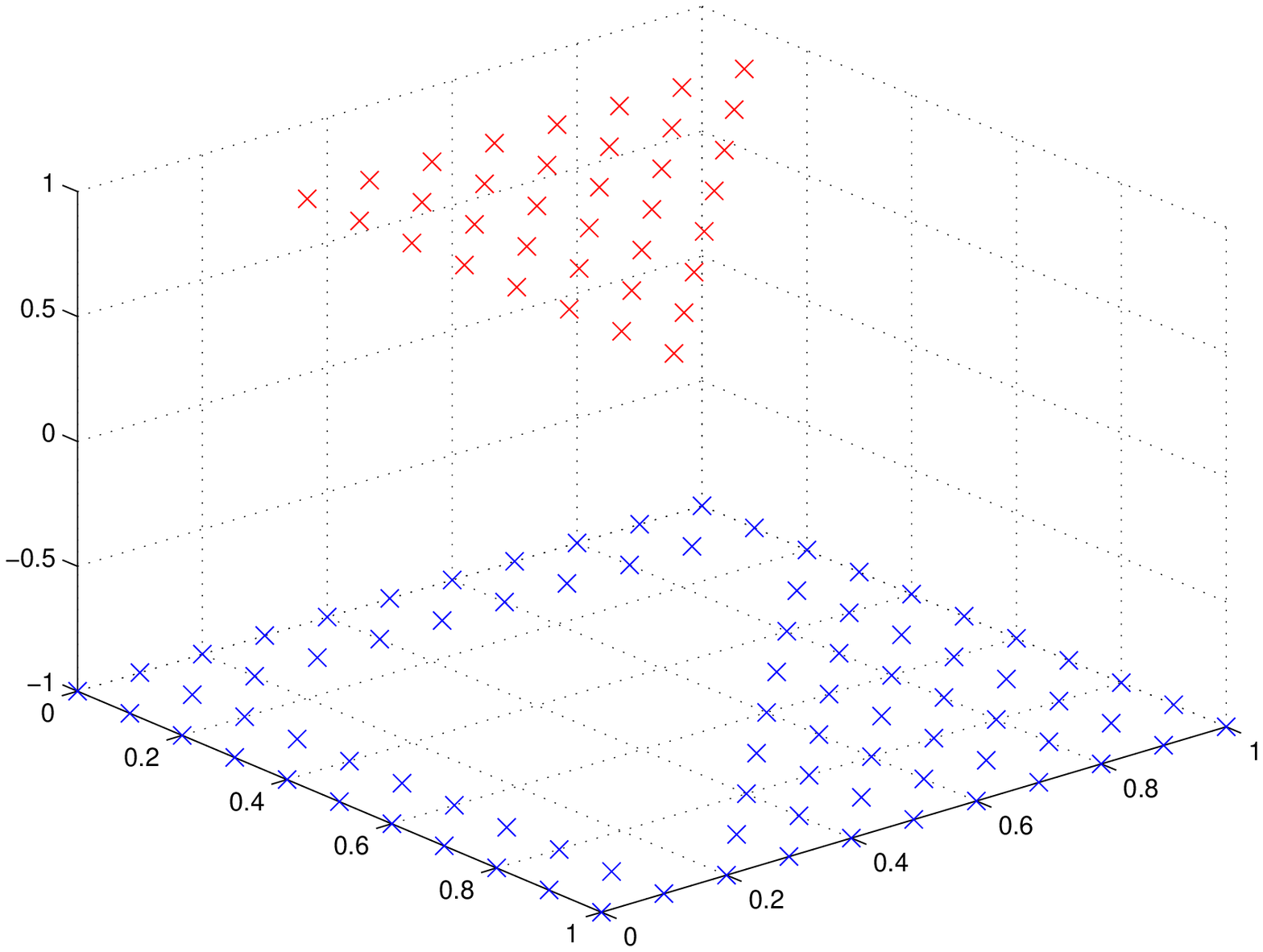}}
\subfigure[Model]{\includegraphics[width=0.49\textwidth, keepaspectratio]{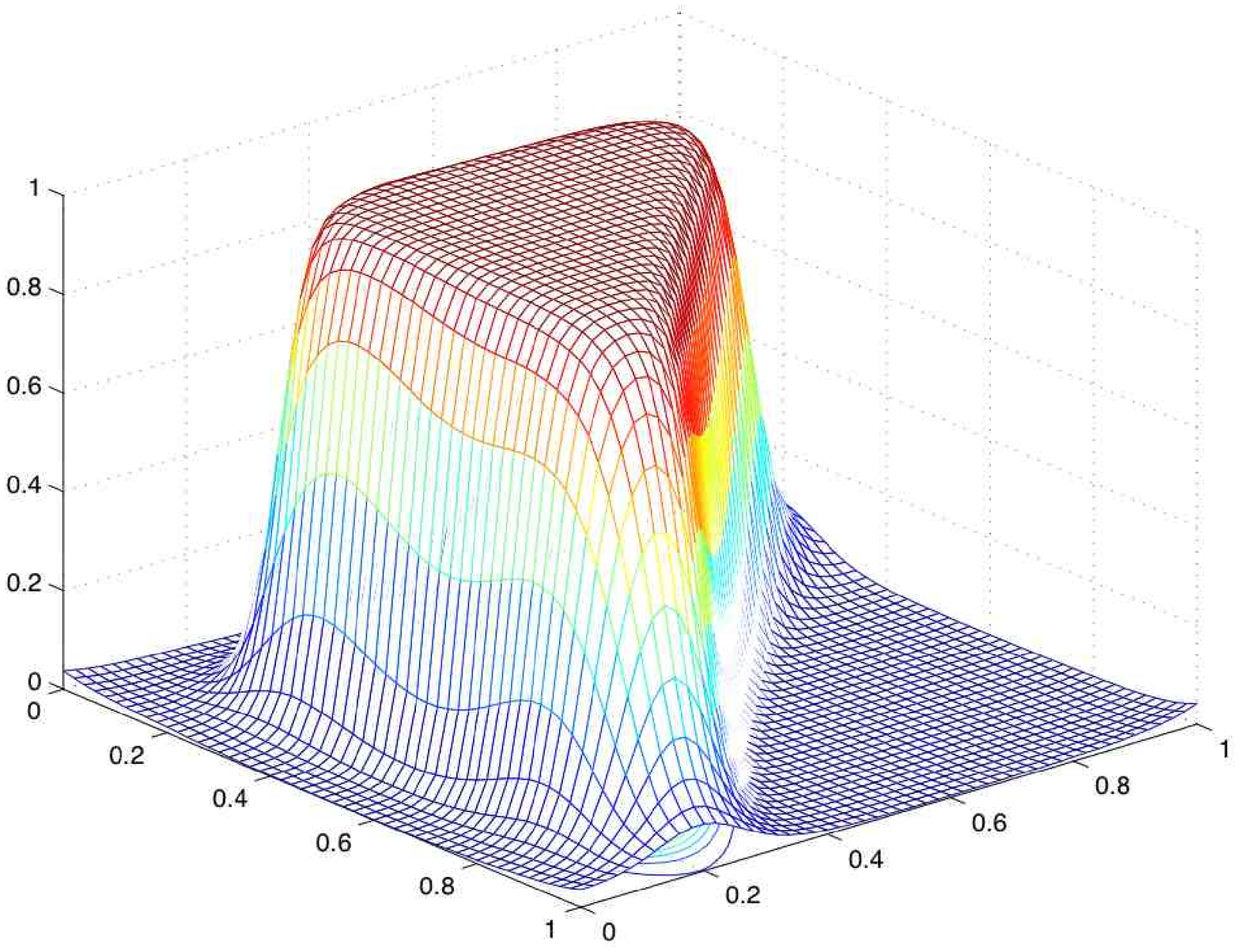}}
\subfigure[Local explanation vectors]{\includegraphics[width=0.49\textwidth, keepaspectratio]{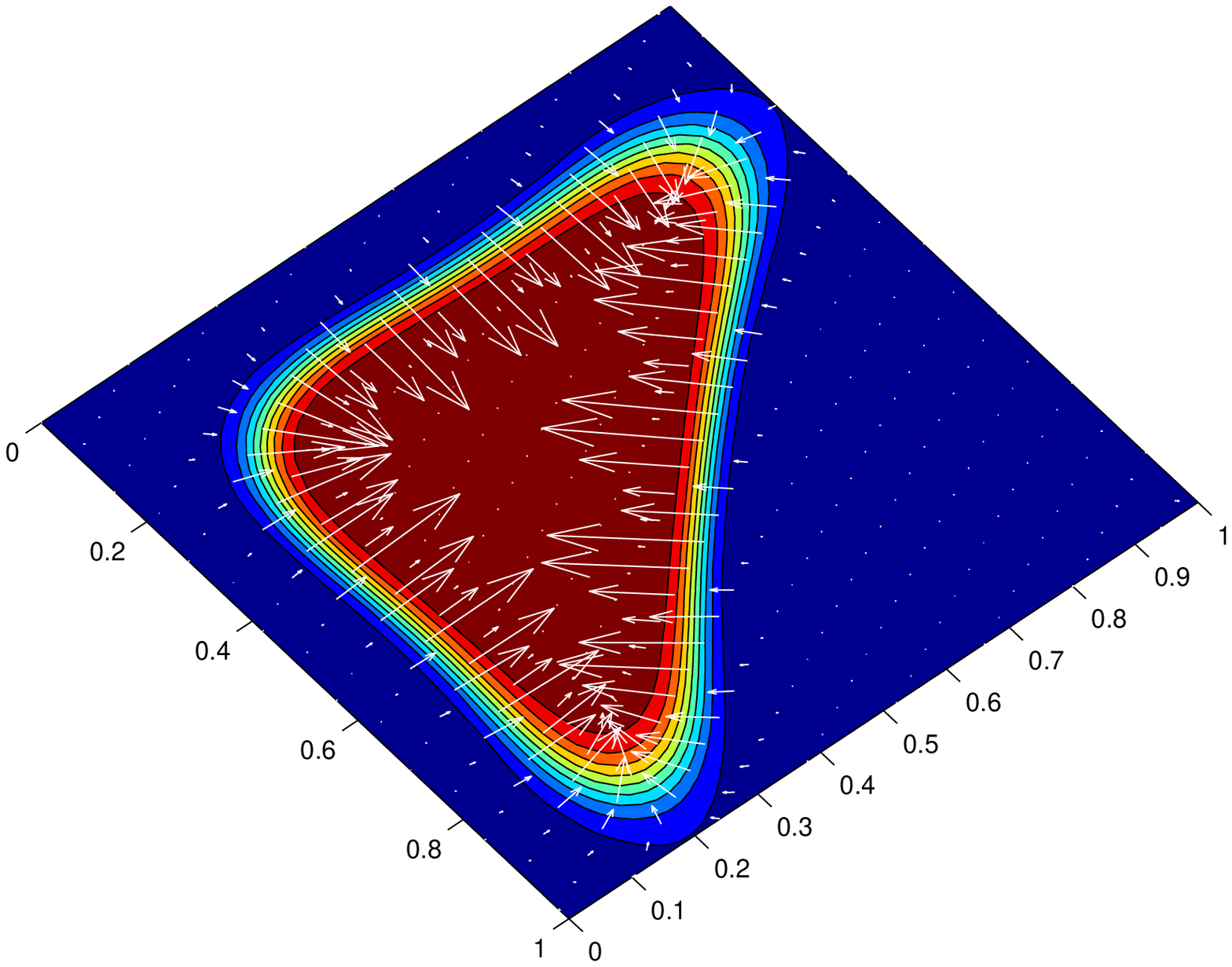}}
\subfigure[Direction of explanation vectors]{\includegraphics[width=0.49\textwidth, keepaspectratio]{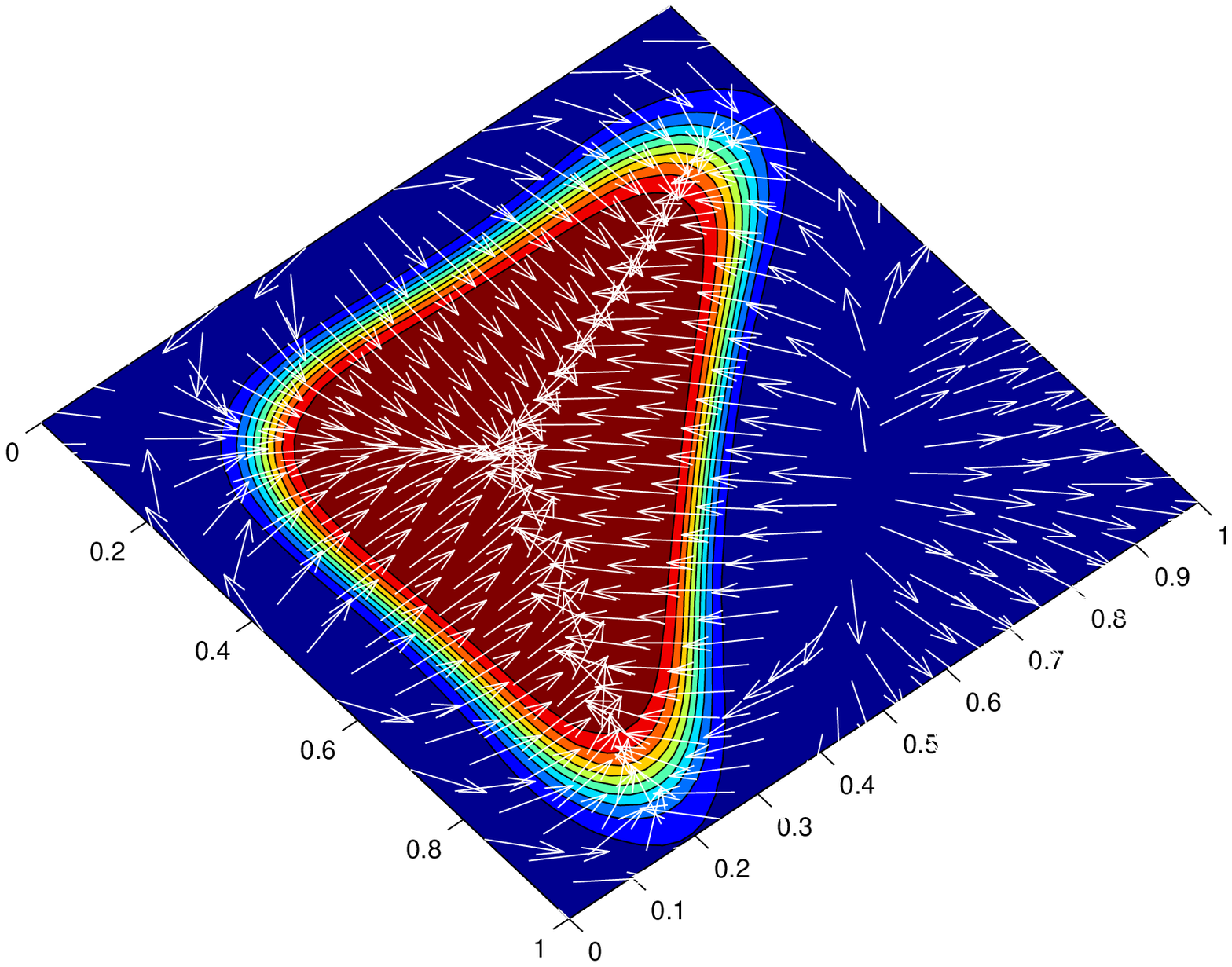}}
\caption{Explaining simple object classification with Gaussian Processes}
\label{fig:toy_gpc}
\end{figure}

Panel (a) of Figure \ref{fig:toy_gpc} shows the training data of a simple object classification task and panel (b) shows the model learned using GPC\footnote{Hyperparameters were tuned by a gradient ascend on the evidence.}. The data is labeled $-1$ for the blue points and $+1$ for the red points. As illustrated in panel (b) the model is a probability function for the positive class which gives every data point a probability of being in this class. Panel (c) shows the probability gradient of the model together with the local gradient explanation vectors. On the hypotenuse and at the corners of the triangle explanations from both features interact towards the triangle class while along the edges the importance of one of the two feature dimensions singles out. At the transition from the negative to the positive class the length of the local gradient vectors represents the increased importance of the relevant features. In panel (d) we see that explanations close to the edges of the plot (especially in the right hand side corner) point away from the positive class. However, panel (c) shows that their magnitude is very small. For discussion of this issue, see Section \ref{sec:discussion}.

%% file: estimating.tex
\section{Estimating Explanation Vectors}
\label{sec:estimating}

\begin{figure}[htbp]
  \centering
  \includegraphics[width=.5\textwidth]{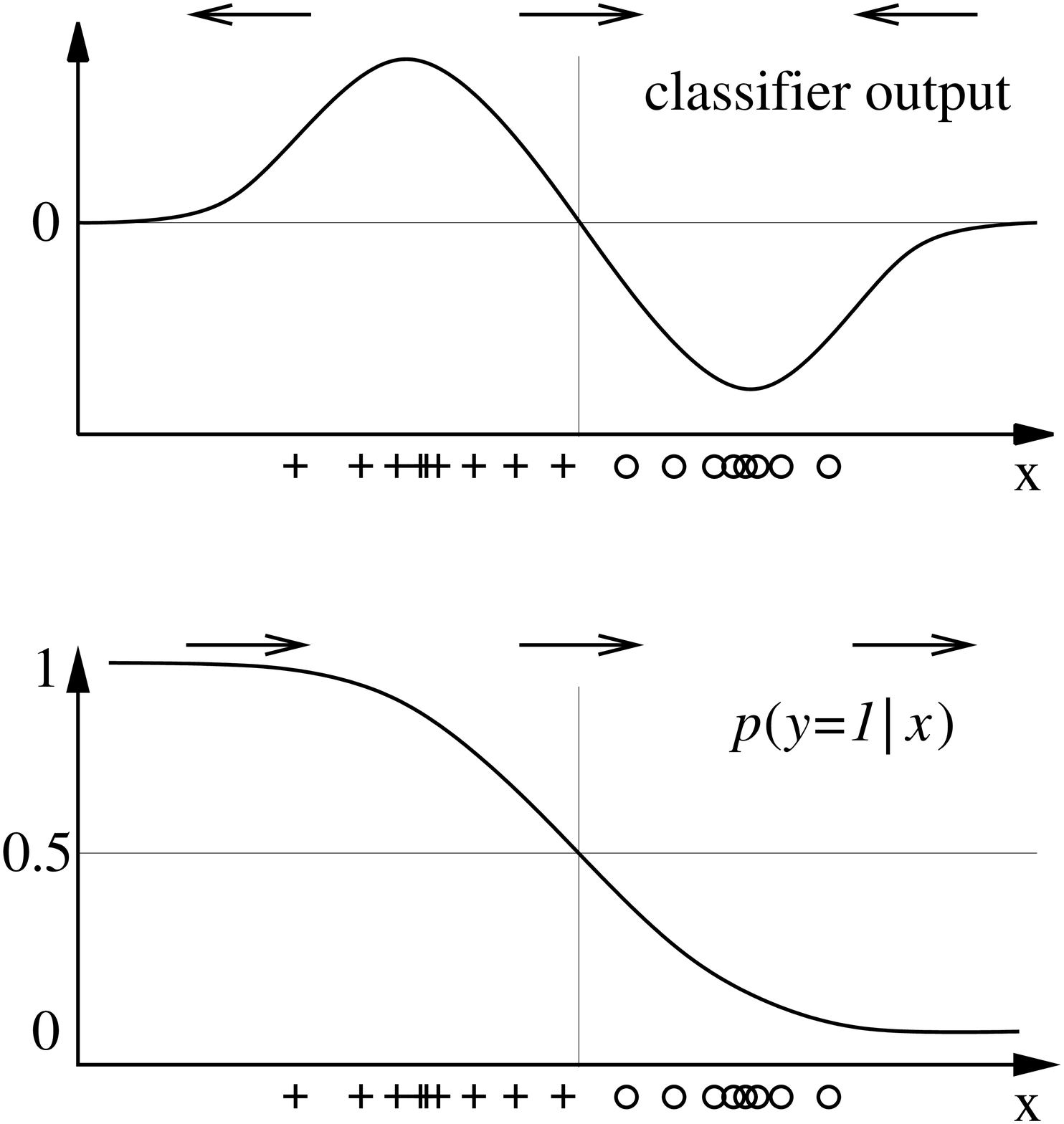}
  \caption{Classifier output of an SVM (top) compared to $p(y\eq 1|x)$ (bottom).}
  \label{fig:svm}
\end{figure}

Several classifier
methods estimate directly the decision rule, which often has no
interpretation as a probability function which is used in our Definition
\ref{def:ev_cls} in Section \ref{sec:definition}.
For example Support Vector Machines
estimate a decision function of the form
\[ f(x) = \sum_{i=1}^n \alpha_i k(x_i, x) + b, \] $\alpha_i, b \in \Re$.
Suppose we have two classes (each with one cluster) in one
dimension (see Figure~\ref{fig:svm}) and train a SVM with RBF kernel.
For points outside the data
clusters $f(x)$ tends to zero.  Thus, the derivative of $f(x)$ (shown
as arrows above the curves) for points on the very left or on the very
right side of the axis will point to the wrong side.
In the following, we will explain how explanations can be obtained
for such classifiers. 

In practice we do not have access to the true underlying distribution
$P(X,Y)$. Consequently, we have no access
to the Bayes classifier as defined in Section \ref{sec:definition}.
Instead we can
apply sophisticated learning machinery like Support Vector Machines
\citep*{Vap95,SchoelkSmo02,MueIEEETNN01} that estimates some classifier $g$ that tries
to mimic $g^*$.  For test data points $z_1, \ldots, z_m\in\Re^d$ which
are assumed to be sampled from the same unknown distribution as the
training data, $g$ estimates labels $g(z_1), \ldots, g(z_m)$.  Now,
instead of trying to explain $g^*$ to which we have no access, we will
define explanation vectors that help us understand the classifier $g$
on the test data points.

Since we do not assume that we have access to some intermediate
real-valued classifier output here (of which $g$ might be a thresholded
version and which further might not be an estimate of $P(Y\eq c\given
X\eq x)$), we suggest to approximate $g$ by another classifier
$\hat{g}$ the actual form of which resembles the Bayes classifier.
There are several choices for $\hat{g}$, e.g. GPC, logistic regression and Parzen windows.\footnote{For Support Vector Machines \citet*{Platt99probabilisticoutputs} fits a sigmoid function to map the outputs to probabilities. In the following, we will present a more general method for estimating explanation vectors.}
In this paper we apply
Parzen windows to the training points to estimate the
weighted class densities $P(Y\eq c) \cdot P(X\given Y\eq c)$,
\begin{align}
\label{eq:4}
  \hat{p}_\sigma(x, y\eq c) = \frac{1}{n}\sum_{i \in I_c} k_\sigma(x-x_i)
\end{align}
for the index set $I_c = \{i~|~g(x_i)=c\}$ and with $k_\sigma(z)$ being a Gaussian kernel $k_\sigma(z) = \exp(-0.5\; z\!^\top
\!z/\sigma^2)/\sqrt{2\pi\sigma^2}$ (as always other kernels are also
possible). This estimates $P(Y\eq c\given X\eq x)$ for
all $c$,
\begin{align}
  \label{eq:3}
  \hat{p}_\sigma(y\eq c|x) = \frac{\hat{p}_\sigma(x,y\eq
  c)}{\hat{p}_\sigma(x,y\eq c)+\hat{p}_\sigma(x,y\neq c)} \approx
\frac{\sum_{i \in I_c} k_\sigma(x-x_i).}{\sum_i k_\sigma(x-x_i),}
\end{align}
and thus an estimate of the Bayes classifier (that mimics $g$),
\begin{align*}
  \hat{g}_\sigma(x) = \arg\min_{c\in\{1,\ldots,C\}} \hat{p}_\sigma(y\neq c\given x).
\end{align*}
This approach has the advantage, that we can use our estimated classifier $g$ to generate
any amount of labeled data for for constructing $\hat{g}$.
The single hyper-parameter $\sigma$ is chosen, such that $\hat{g}$
approximates $g$ (which we want to explain), i.e.
\begin{align*}
  \hat{\sigma} := \arg\min_\sigma \sum_{j=1}^m \;I\left\{ g(z_j) \neq \hat{g}_\sigma(z_j) \right\},
\end{align*}
where $I\{\cdots\}$ is the indicator function.  
$\sigma$ is assigned the constant value $\hat{\sigma}$ from here on and omitted as a subscript.
For $\hat{g}$ it is straightforward to define explanation vectors:
\begin{definition}
  \label{eq:17}
\begin{gather*}
  \notag\hat{\zeta}(z) := \left.\frac{\partial}{\partial x}
    \;\hat{p}(y\neq g(z)\given x)\right\rvert_{x=z}
  = 
\frac{\Big(\sum_{i \notin I_{g(z)}}k(z-x_i)\Big)\Big(\sum_{i \in I_{g(z)}} k(z-x_i)(z-x_i)\Big)}%
  {\sigma^2 \;\big(\sum_{i=1}^n k(z-x_i)\big)^2}\\
  -\frac{\Big(\sum_{i \notin I_{g(z)}}k(z-x_i) (z-x_i)\Big)\Big(\sum_{i \in I_{g(z)}} k(z-x_i)\Big)}
  {\sigma^2 \;\big(\sum_{i=1}^n k(z-x_i)\big)^2}
\end{gather*}
\end{definition}
which is easily derived using Eq.~(\ref{eq:3}) and the derivative of
Eq.~(\ref{eq:4}), see Appendix \ref{sec:appendix_estimating_derivation}.  Note that we use $g$ instead of
$\hat{g}$.
This
choice ensures that the orientation of $\hat{\zeta}(z)$ fits to the
labels assigned by $g$, which allows better interpretations.

In summary, we imitate the classifier $g$ which we would like to
explain locally, by a Parzen window classifier $\hat{g}$ that has the
same form as the Bayes estimator and for which we can thus easily
estimate the explanation vectors using Definition~\ref{eq:17}. Practically
there are some caveats: the mimicking classifier $\hat{g}$ has to be
estimated from $g$ even in high dimensions; this needs to be done with
care.
However, in principle we
have an arbitrary amount of training data available for constructing
$\hat{g}$ since we may use our estimated classifier $g$ to generate
labeled data.

%% file: iris_knn.tex
\section{Explaining Iris Flower Classification by $k$-Nearest Neighbors}
\label{sec:iris_knn}

The Iris flower data set \cite[introduced in][]{fis1936iris} describes
150 flowers from the genus Iris by 4 features: sepal length, sepal
width, petal length, and petal width, which are easily measured
properties of certain leaves of the corolla of the flower.  There are
three clusters in that data which correspond to three different
species: Iris setosa, Iris virginica, and Iris versicolor.

Let us consider the problem of classifying the data points of Iris
versicolor (class 0) against the other two species (class 1).  We
applied some standard classification machinery to this problem as
detailed in the following:
\begin{itemize}
\item Class 0 consists of all examples of Iris versicolor.
\item Class 1 consists of all examples of Iris setosa and Iris virginica.
\item Randomly split 150 data points into 100 training and 50 test
  examples.
\item Normalize training and test set using the mean and variance of
  the training set.
\item Apply $k$-nearest neighbor classification with $k=4$ (chosen by
  leave-one-out cross validation on the training data).
\item Training error is 3\% (i.e.~3 mistakes in 100).
\item Test error is 8\% (i.e.~4 mistakes in 50).
\end{itemize}
In order to estimate explanation vectors we mimic the classification
results with a Parzen window classifier.  The best fit (3\% error) is
obtained with a kernel width of $\sigma=0.26$ (chosen by
  leave-one-out cross validation on the training data).

\begin{figure}
\centering
\includegraphics[width=\textwidth]{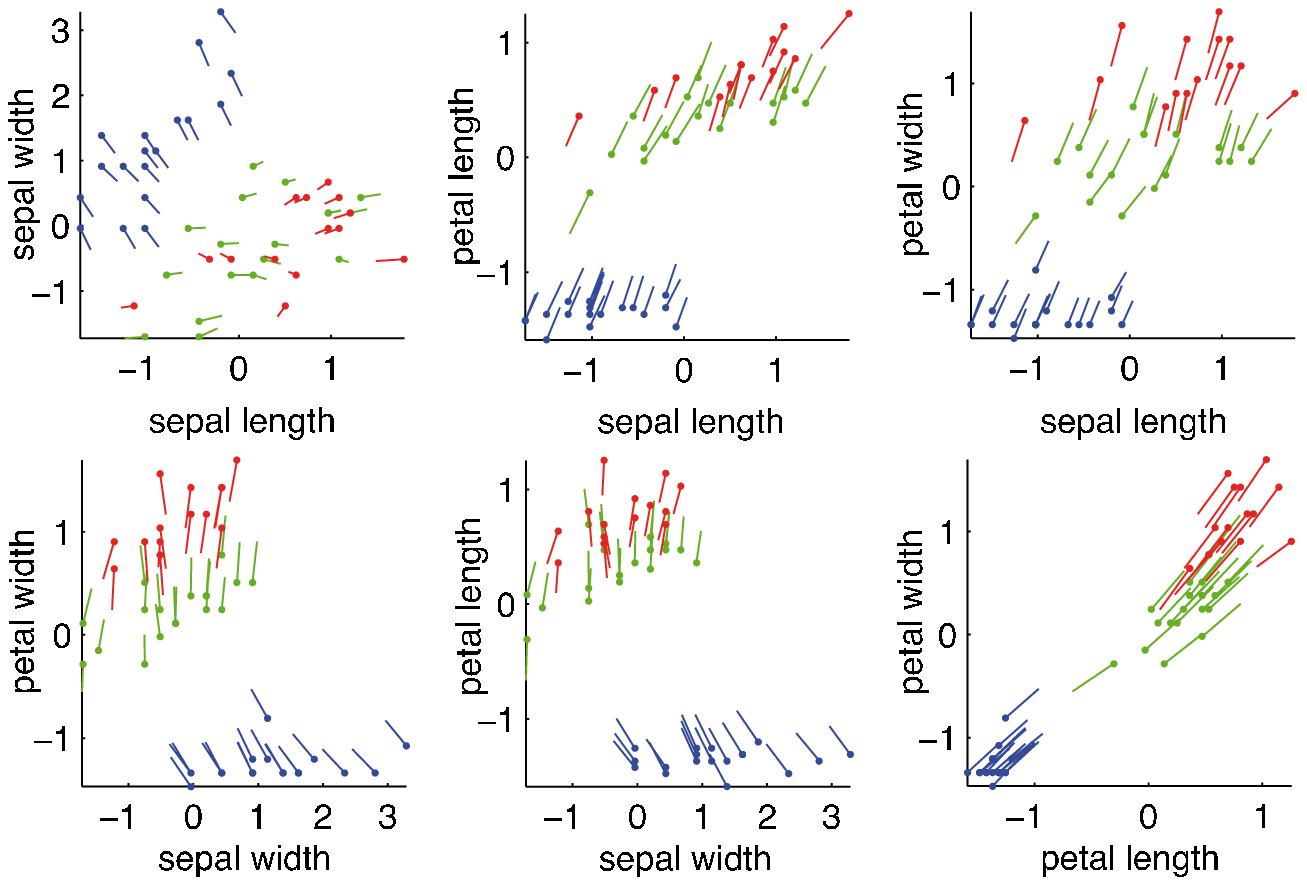}
\caption{\label{fig:iris_explain}Scatter plots of the explanation
    vectors for the 
test data.
    Shown are all explanation vectors for both classes:
    class 1 containing Iris setosa (shown in blue) and Iris virginica
    (shown in red) versus class 0 containing only one species Iris
    versicolor (shown in green).  Note that the explanations why an
    Iris flower is not an Iris versicolor is different for Iris setosa
    and Iris virginica.}
\end{figure}

Since the explanation vectors live in the input space we can visualize
them with scatter plots of the initially measured features.
The resulting \emph{explanations} (i.e.~vectors) for the test set are shown in
Figure~\ref{fig:iris_explain}.
The blue dots correspond to
explanation vectors for Iris setosa and the red dots for Iris virginica
(both class 1).  Both groups of dots point to the green dots of Iris
versicolor.  The most important feature is the combination of petal
length and petal width (see the corresponding panel), the product of
which corresponds roughly to the area of the petals.  However, the
resulting explanations for the two species in class 1 are different:
\begin{itemize}
\item Iris setosa (class 1) is different from Iris versicolor (class
  0) because its petal area is \emph{smaller}.
\item Iris virginica (class 1) is different from Iris versicolor (class
  0) because its petal area is \emph{larger}.
\end{itemize}
Also the dimensions of the sepal (another part of the blossom) is
relevant, but not as distinguishing.



%% file: usps_svm.tex
\section{Explaining USPS Digit Classification by Support Vector Machine}
\label{sec:usps_svm}
\begin{figure}
  \includegraphics[width=.49\textwidth]{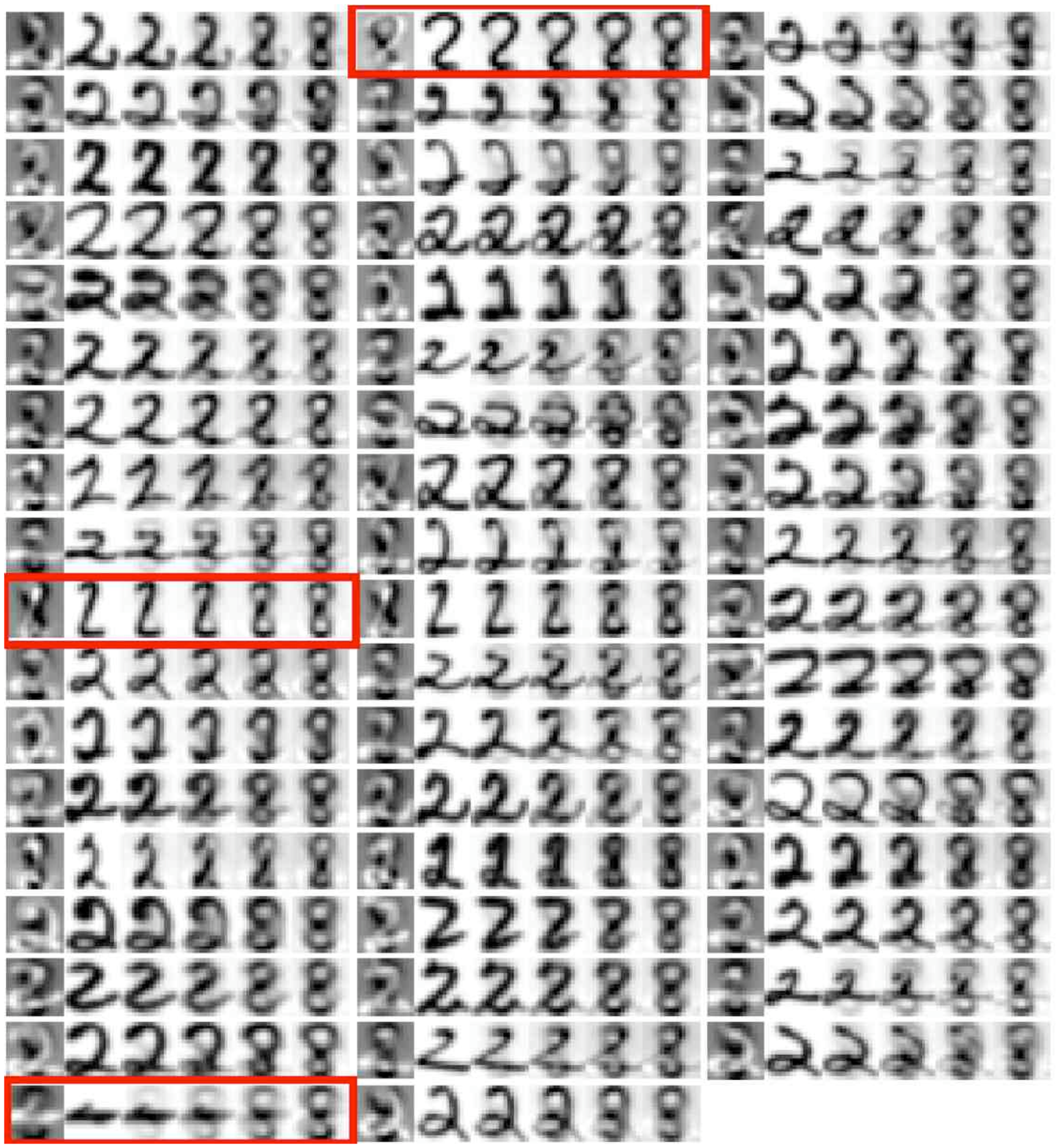}
  \includegraphics[width=.49\textwidth]{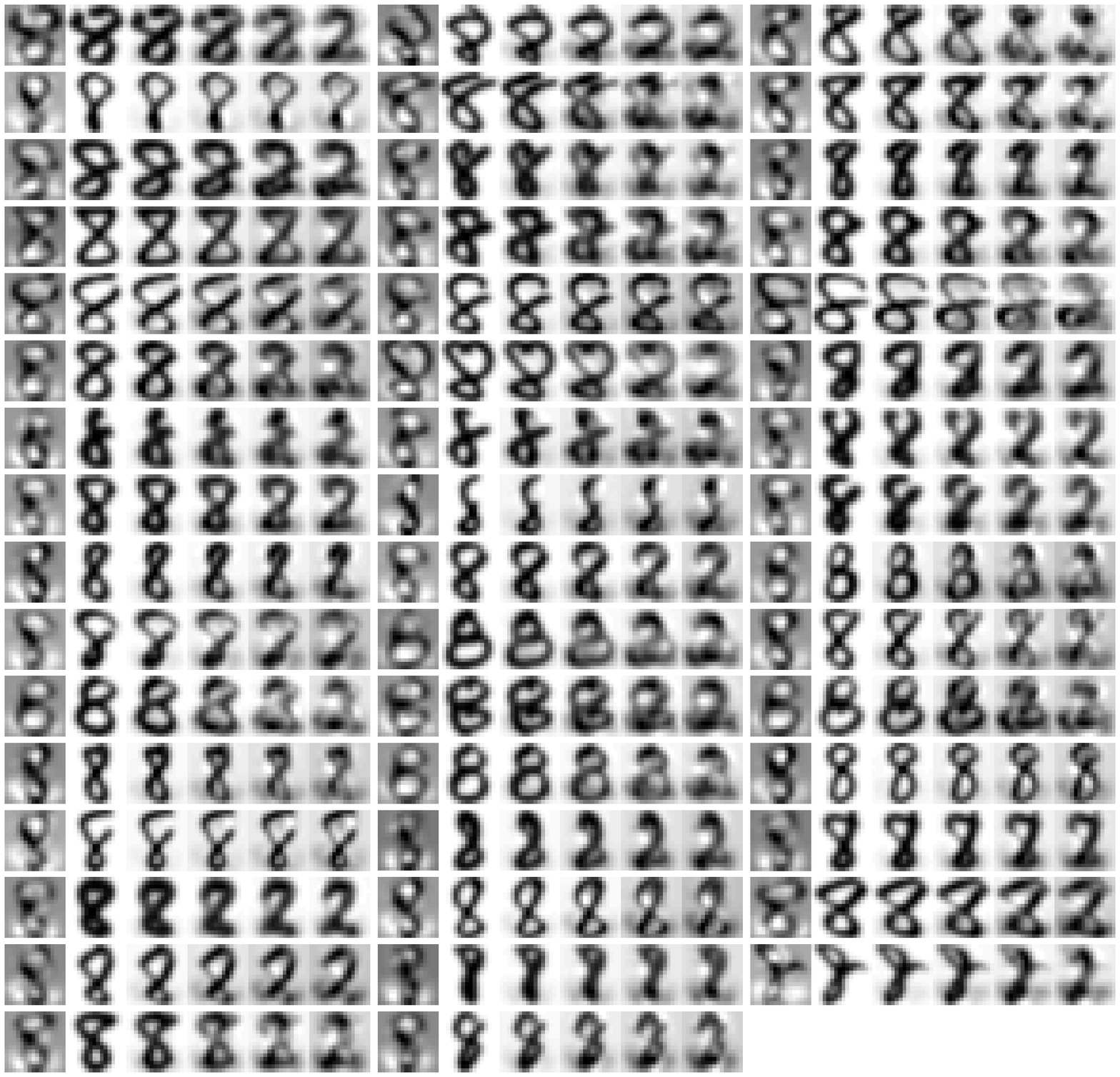}
  \caption{USPS digits (training set):
    'twos' (left) and 'eights' (right) with correct classification. For
    each digit from left to right: (i) explanation vector (with black
    being negative, white being positive), (ii) the original digit,
    (iii-end) artificial digits along the explanation vector towards
    the other class.}
  \label{fig:usps_correct_tr}
\end{figure}
\begin{figure}
    \includegraphics[width=.49\textwidth]{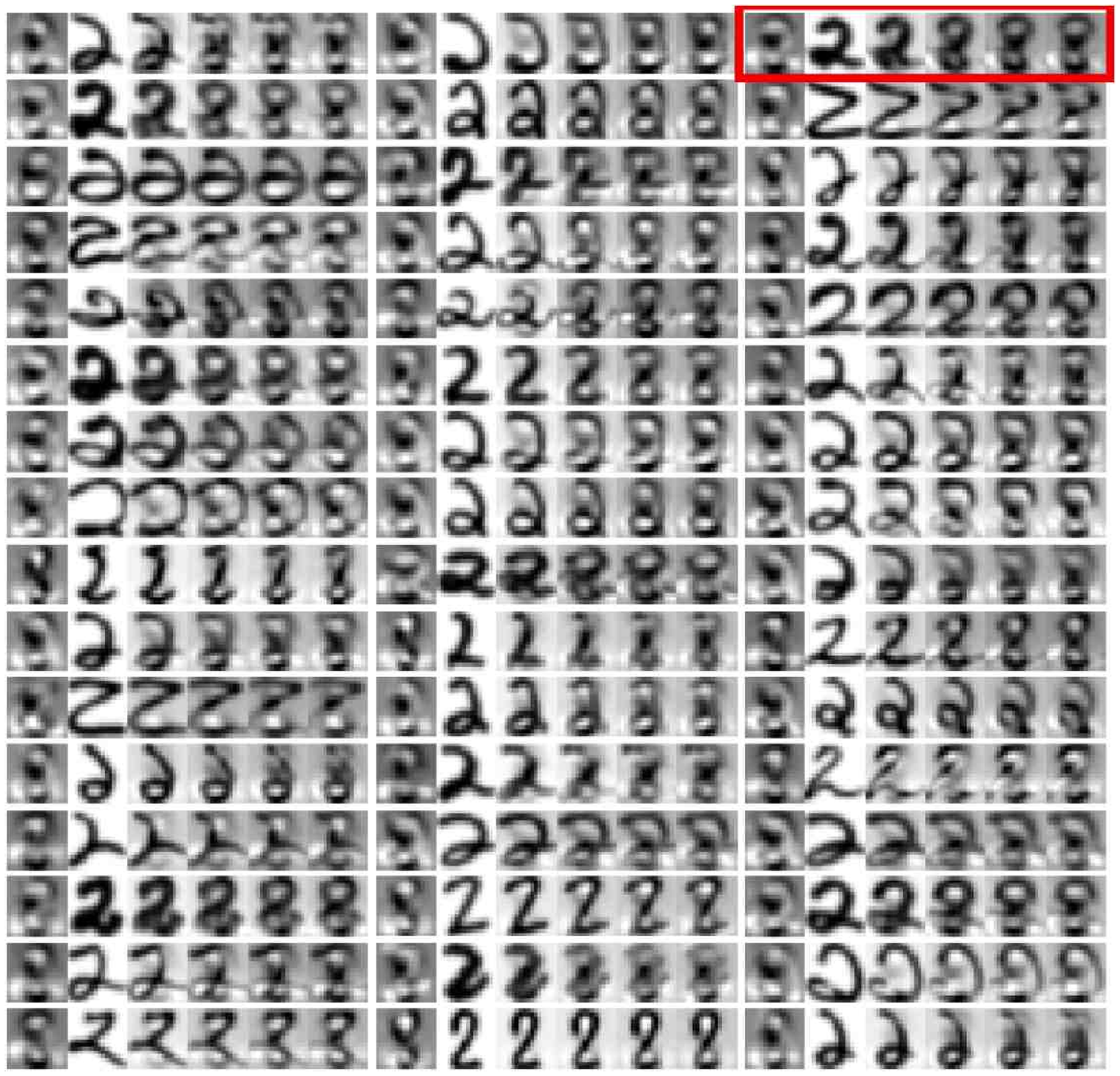}
    \includegraphics[width=.49\textwidth]{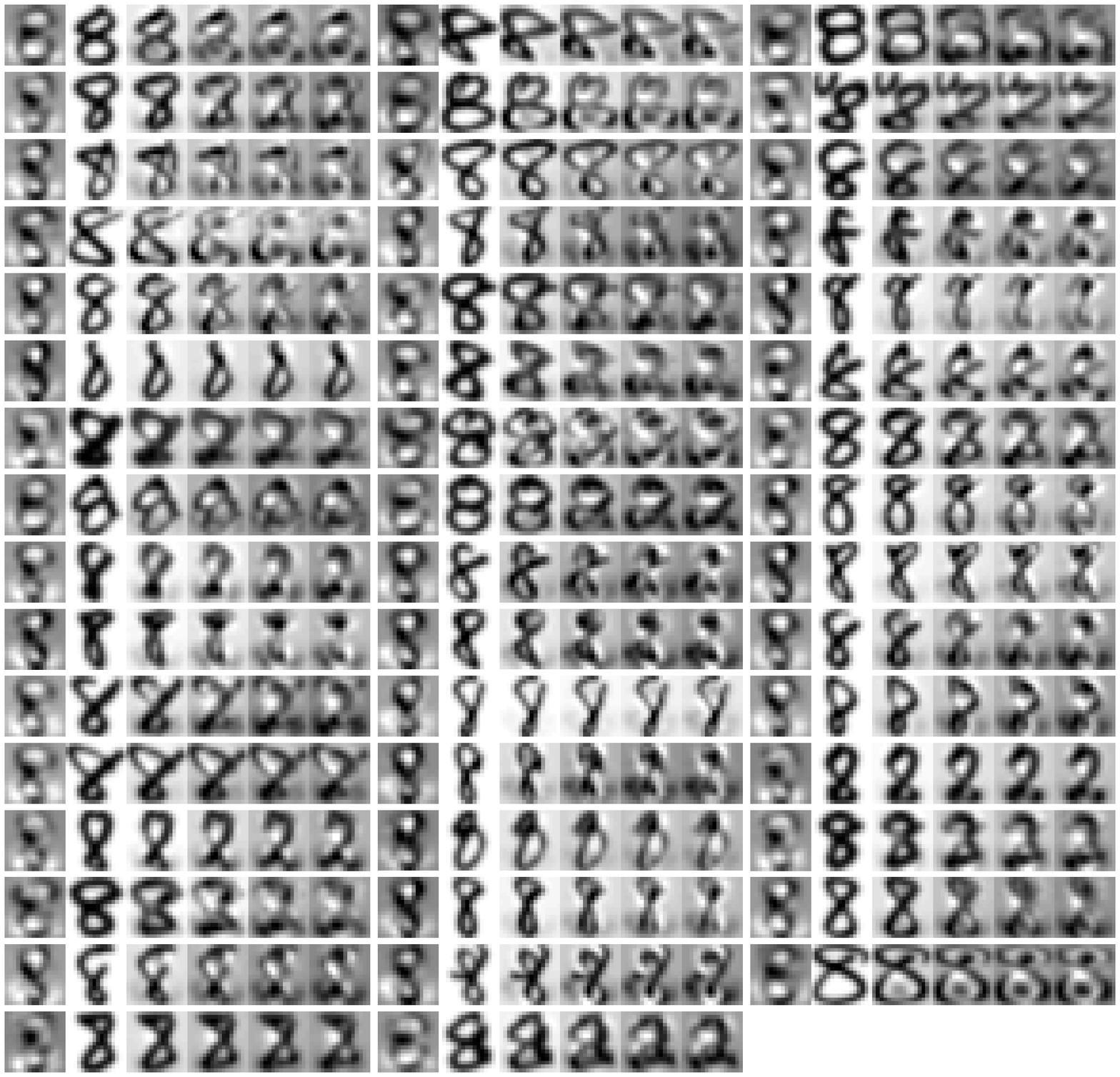}
  \caption{USPS digits (test set bottom part):
    'twos' (left) and 'eights' (right) with correct classification. For
    each digit from left to right: (i) explanation vector (with black
    being negative, white being positive), (ii) the original digit,
    (iii-end) artificial digits along the explanation vector towards
    the other class.}
  \label{fig:usps_correct_te}
\end{figure}
%
We now apply the framework of estimating explanation vectors to a high dimensional data
set, the USPS digits. The classification problem that we designed for
illustration purposes is detailed in the following list:
\begin{itemize}
\item all digits are $16\times 16$ images which are reshaped to
  $256\times 1$ dimensional column vectors
\item classifier: SVM from \citet*{Sch2002} with
  RBF-kernel-width $\sigma = 1$ and regularization constant $C=10$ (chosen by
  grid search in cross validation on the training data).
\item training set: 47 'twos', 53 'eights'; training error 0.00
\item test set: 48 'twos', 52 'eights'; test error 0.05
\end{itemize}
We approximated the estimated class labels obtained by the SVM with
the Parzen window classifier (Parzen window size $\sigma=10.2505$,
chosen by grid search in cross validation on the training data).
The SVM and the Parzen window classifier only disagreed on 2\% of
the test examples, so a good fit was achieved.  Figures
\ref{fig:usps_correct_tr} and \ref{fig:usps_correct_te} show our
results.  All parts show three examples per row.  For each example we
display from left to right: (i) the explanation vector, (ii) the
original digit, (iii-end) artificial digits along the explanation
vector towards the other class.\footnote{For the sake of simplicity,
no intermediate
updates were performed, i.e. artificial digits were generated
by taking equal sized steps in the direction given by 
the original explanation vector calculated fot the original digit.}
These artificial digits should help
to understand and interpret the explanation vector.  Let us first have
a look at the results on the training set:
\begin{description}
\item[Figure~\ref{fig:usps_correct_tr} (left panel):] Let
us focus on the top example framed in red.
The line that forms
the 'two' is part of some 'eight' from the data set.  Thus the parts
of the lines that are missing show up in the explanation vector: if
the dark parts (which correspond to the missing lines) are added to
the 'two' digit then it will be classified as an 'eight'.  Or in
other words, because of the lack of those parts the digit was
classified as a 'two' and not as an 'eight'.  A similar explanation
holds for the middle example framed in red of the same Figure.
Not all examples transform easily to 'eights': besides adding parts of
black lines, some existing black spots (that the digit has to be a
'two') must be removed.  This is reflected in the explanation vector
by white spots/lines.  Curious is the bottom 'two' framed in red,
which is actually a dash and is in the data set by mistake.  However,
its explanation vector shows nicely which parts have to be added and
which have to be removed.
\item[Figure~\ref{fig:usps_correct_tr} (right panel):] we
see similar results for the 'eights' class.  The explanation vectors
again tell us how the 'eights' must change to become classified as
'twos'.  However, sometimes the transformation does not reach the
'twos'.  This is probably due to the fact that some of the 'eights'
are inside the cloud of 'eights'.
\end{description}
On the test set the explanation vectors are not as pronounced as on
the training set.  However, they show similar tendencies:
\begin{description}
\item[Figure~\ref{fig:usps_correct_te} (left panel):] we
see the correctly classified 'twos'.  Let's focus on the example framed in red.
Again the explanation vector shows us how to edit
the image of the 'two' to make it some of the 'eights', i.e.~exactly
what parts of the digit have been important for the classification
result.  For several other 'twos' the explanation vectors do not
directly lead to the 'eights' but weight the different parts of
the digits which have been relevant for the classification.
\item[Figure~\ref{fig:usps_correct_te} (right panel):]
similarly to the training data, we see that also these explanation
vectors are not bringing all 'eights' to 'two'.  Their explanation
vectors mainly suggest to remove most of the eights (the dark parts)
and add some in
the lower part (the light parts, which look like a white shadow).
\end{description}
Overall, our findings can be summarized, that the explanation vectors
tell us how to edit our example digits to change the assigned class
label.  Hereby, we get a better understanding of the reasons why the
chosen classifier classified the way it did.

%% file: tox_gpc.tex
\section{Explaining Mutagenicity Classification by Gaussian Processes}
\label{sec:tox_gpc}
In the following Section we describe an application of our local gradient explanation methodology to a complex real world data set. Our aim is to find structure specific to the problem domain that has \emph{not} been fed into training  explicitly but is captured implicitly by the GPC model in the high-dimensional feature space used to determine its prediction. 
We investigate the task of predicting Ames mutagenic activity of chemical compounds.
Not being mutagenic (i.e. not able to cause mutations in the DNA) is an important requirement for compounds under investigation in drug discovery and design. The Ames test \citep*{Ames_1972} is a standard experimental setup for measuring mutagenicity.
The following experiments are based on a  set of Ames test results for 6512 chemical compounds that we published  previously.\footnote{See \citet*{OurTox} for results of modeling this set using different machine learning methods. The data itself is available online at \url{http://ml.cs.tu-berlin.de/toxbenchmark}} 

\begin{figure}
\centering
\includegraphics[width=0.5\textwidth,keepaspectratio]{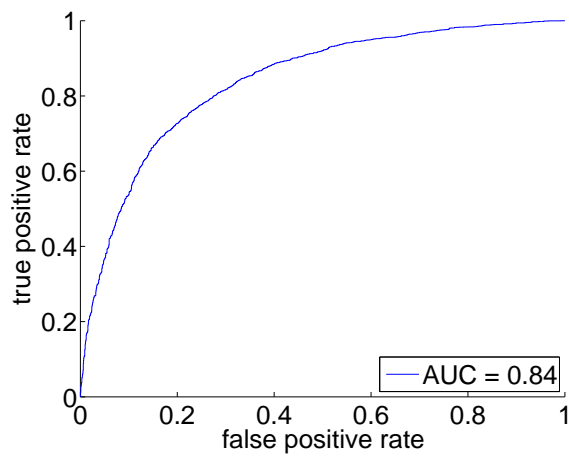}
\caption{Receiver operating curve of GPC model for mutagenicity prediction}
\label{fig:tox_gpc_error}
\end{figure}

GPC was applied as detailed in the following:
\begin{itemize}
\item Class 0 consists of non-mutagenic compounds
\item Class 1 consists of mutagenic compounds
\item Randomly split 6512 data points into 2000 training and 4512 test
  examples such that:
\begin{itemize}
\item The training set consists of equally many class 0 and class 1 examples.
\item For the steroid compound class the balance in the train and test set is enforced.
\end{itemize}
\item 10 additional random splits were investigated individually. This confirmed the results presented below.
\item Each example (chemical compound) is represented by  a vector of counts of 142 molecular substructures calculated using the \textsc{Dragon} software \citep*{Dragon_Manual_Online}.
\item Normalize training and test set using the mean and variance of
  the training set.
\item Apply GPC model with RBF kernel
\item Performance (84 \% area under curve) confirms our previous results \citep*{OurTox}. Error rates can be obtained from Figure \ref{fig:tox_gpc_error}.
\end{itemize}
Together with the prediction we calculated the explanation vector (as introduced in Section \ref{sec:definition} with Definition \ref{def:ev_cls}) for each test point. The remainder of this Section is an evaluation of these local explanations.

\begin{figure}
\centering
\subfigure[aromatic nitro]{\includegraphics[width=0.3\textwidth,keepaspectratio]{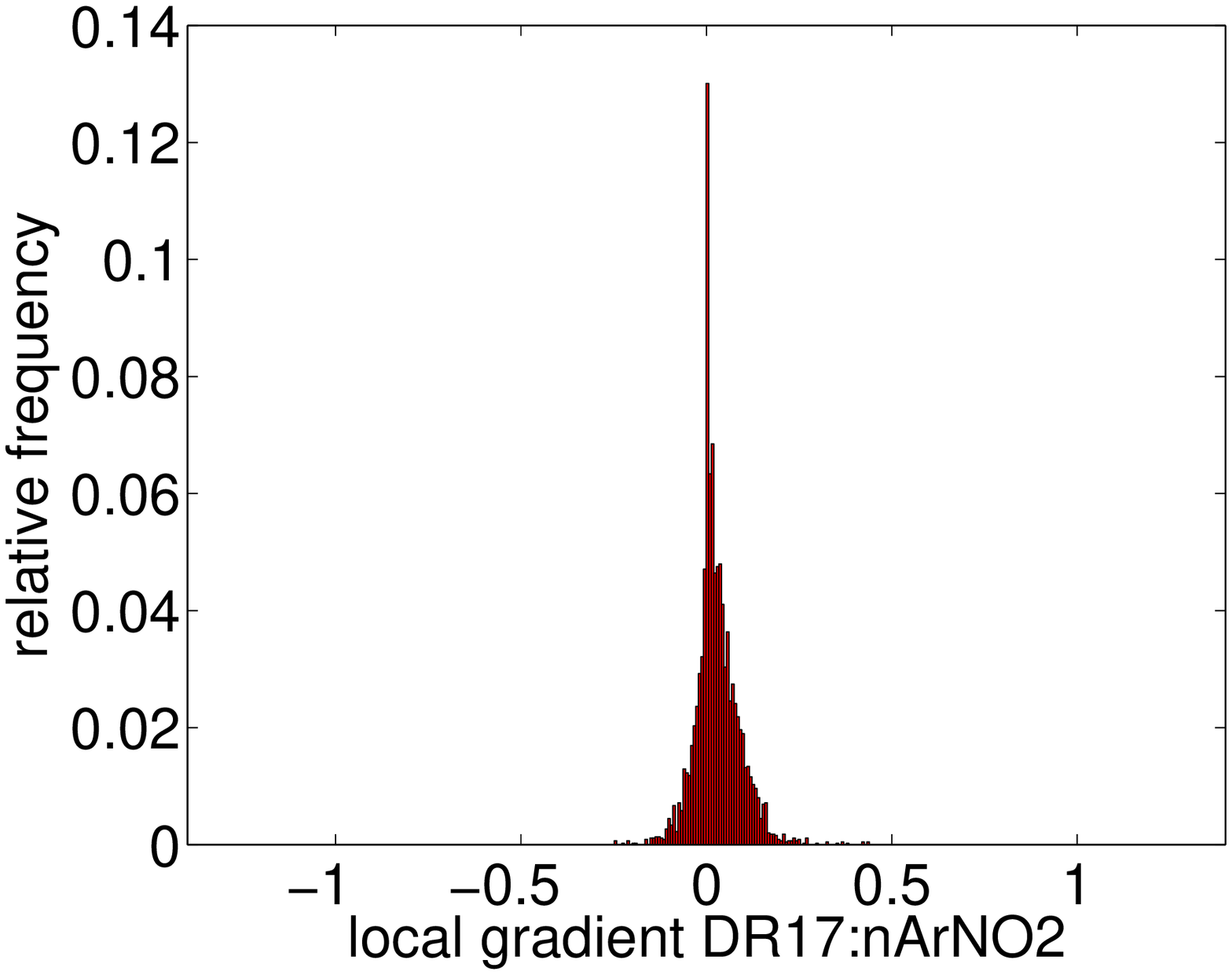}}
\subfigure[aromatic amine]{\includegraphics[width=0.3\textwidth,keepaspectratio]{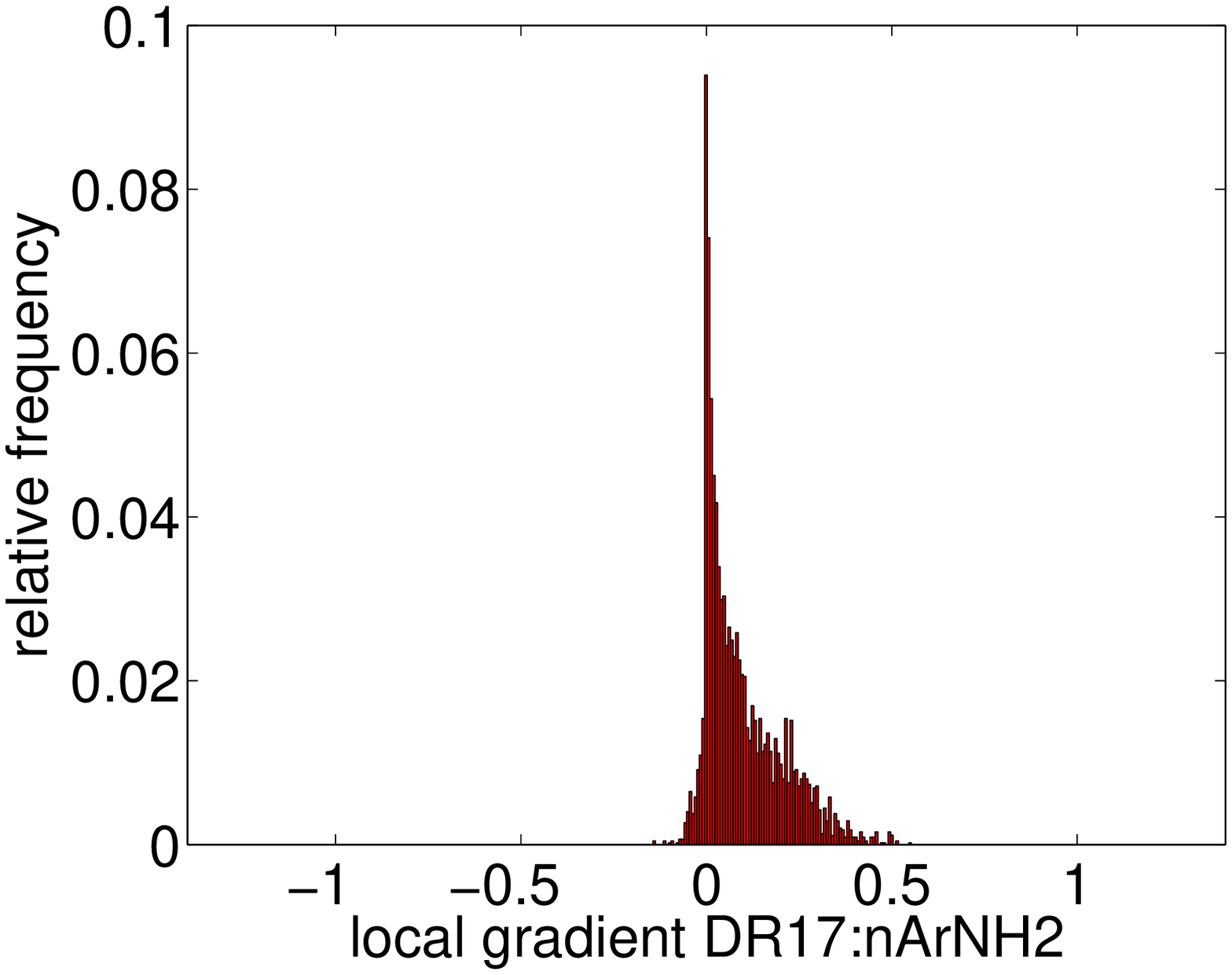}}
\subfigure[aromatic nitroso]{\includegraphics[width=0.3\textwidth,keepaspectratio]{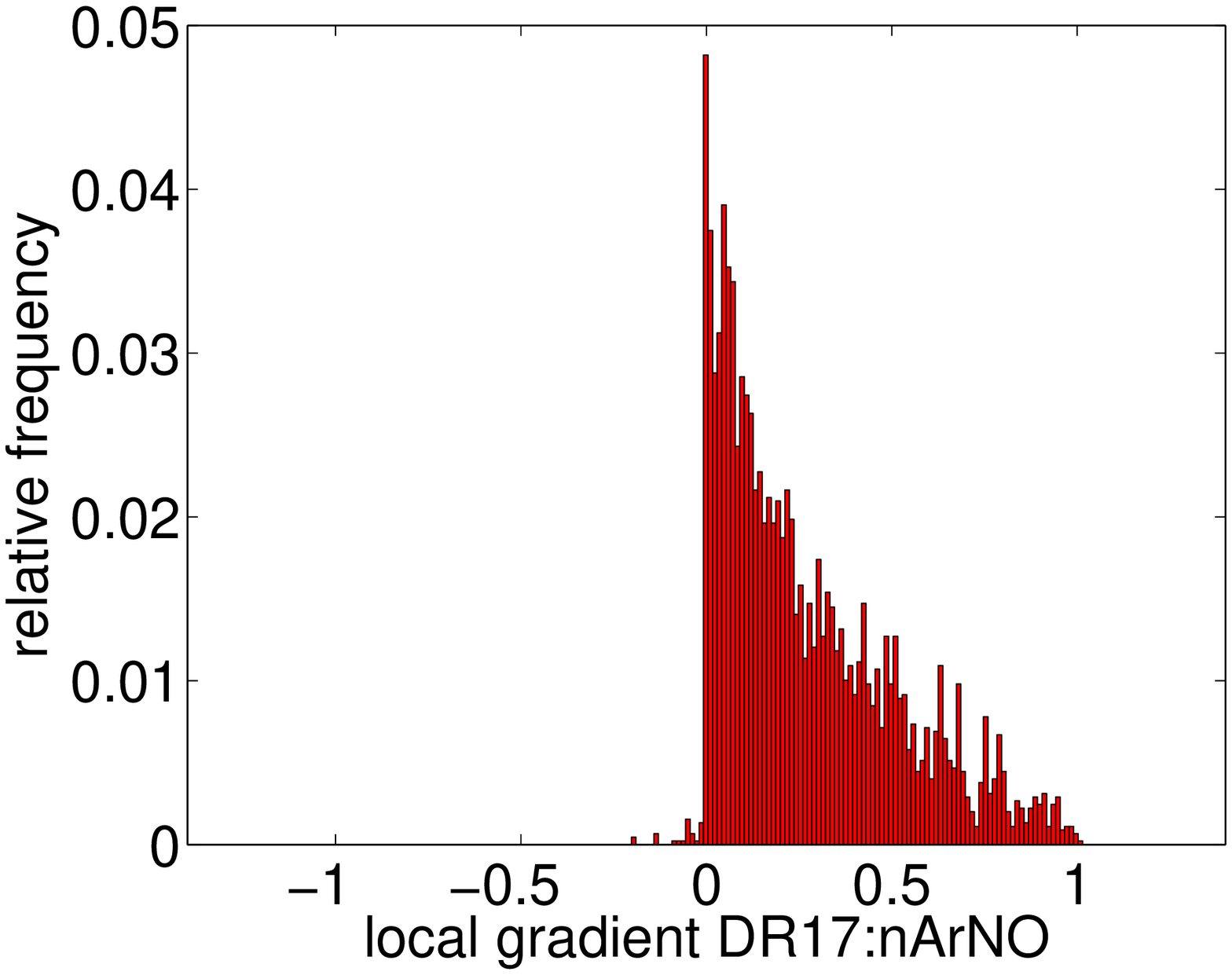}}
\subfigure[aliphatic nitrosamine]{\includegraphics[width=0.3\textwidth,keepaspectratio]{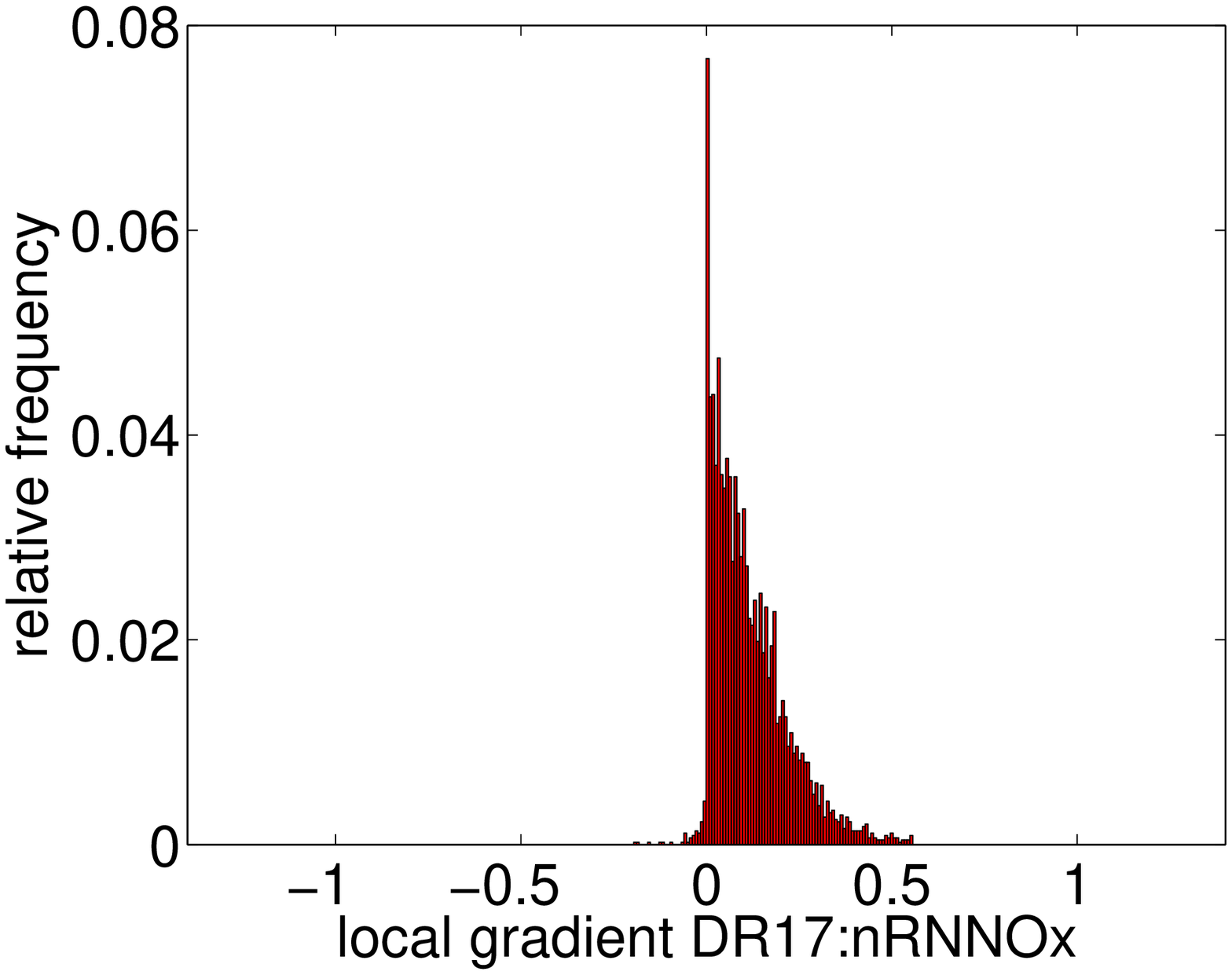}}
\subfigure[aromatic nitrosamine]{\includegraphics[width=0.3\textwidth,keepaspectratio]{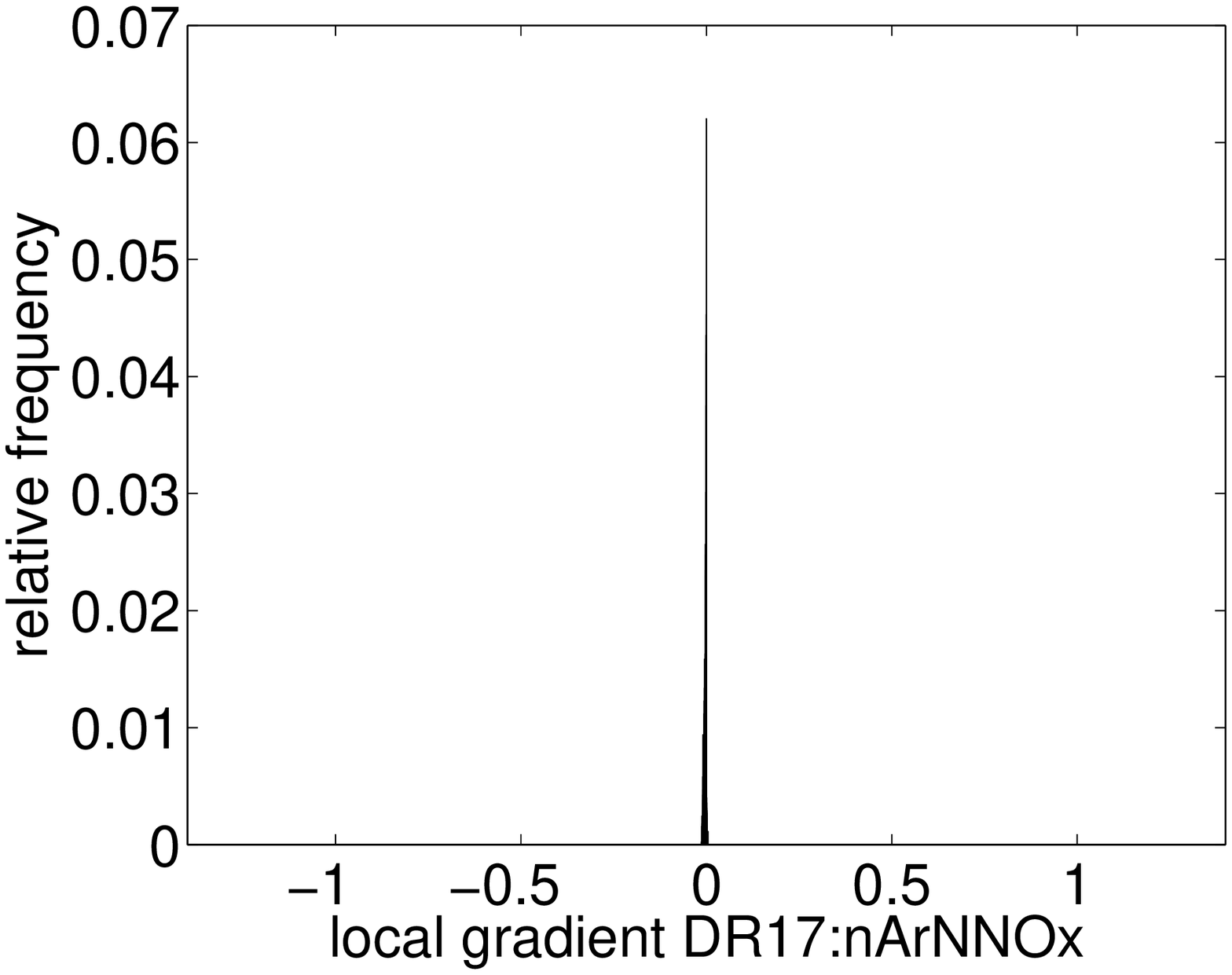}}
\subfigure[epoxide]{\includegraphics[width=0.3\textwidth,keepaspectratio]{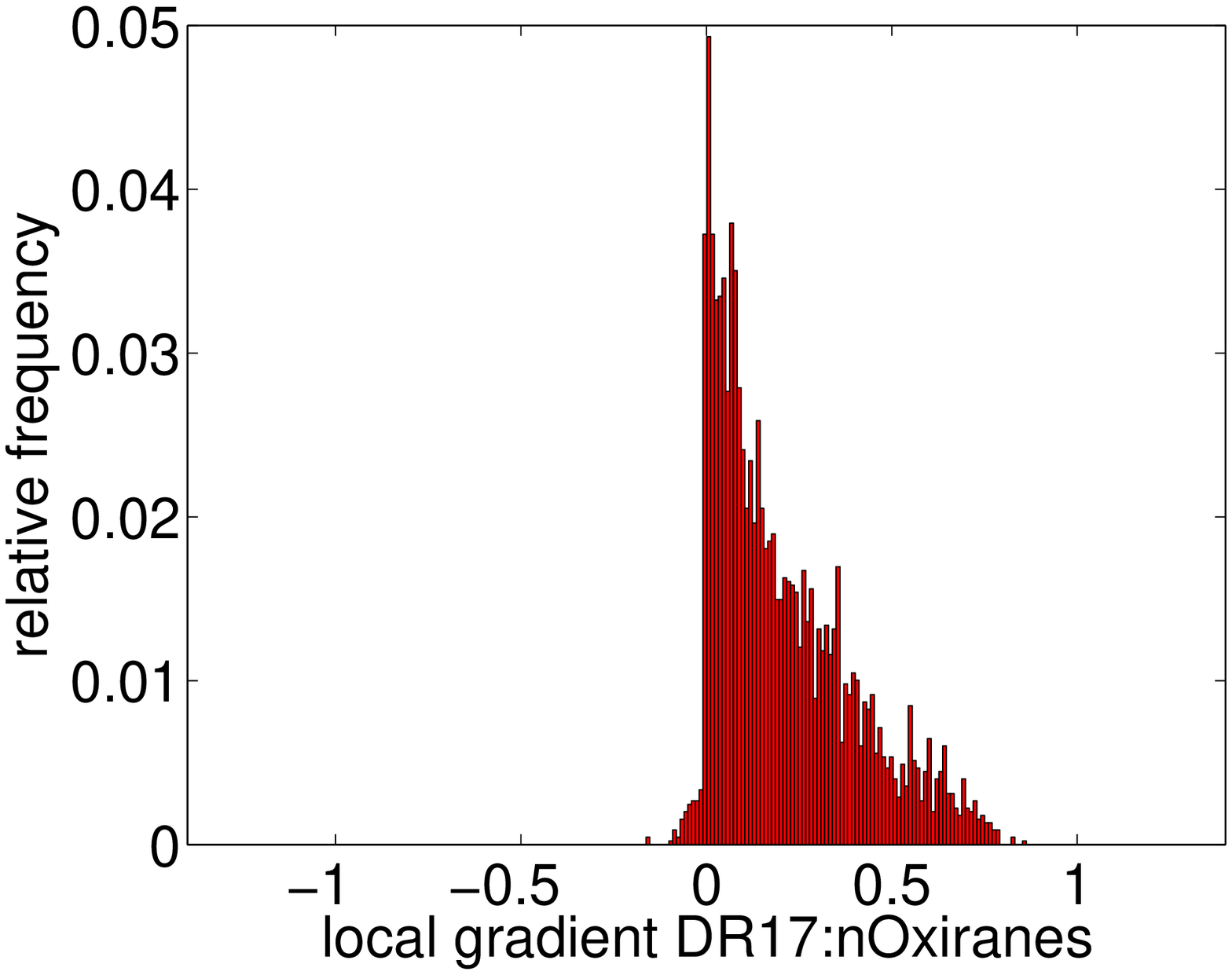}}
\subfigure[aziridine]{\includegraphics[width=0.3\textwidth,keepaspectratio]{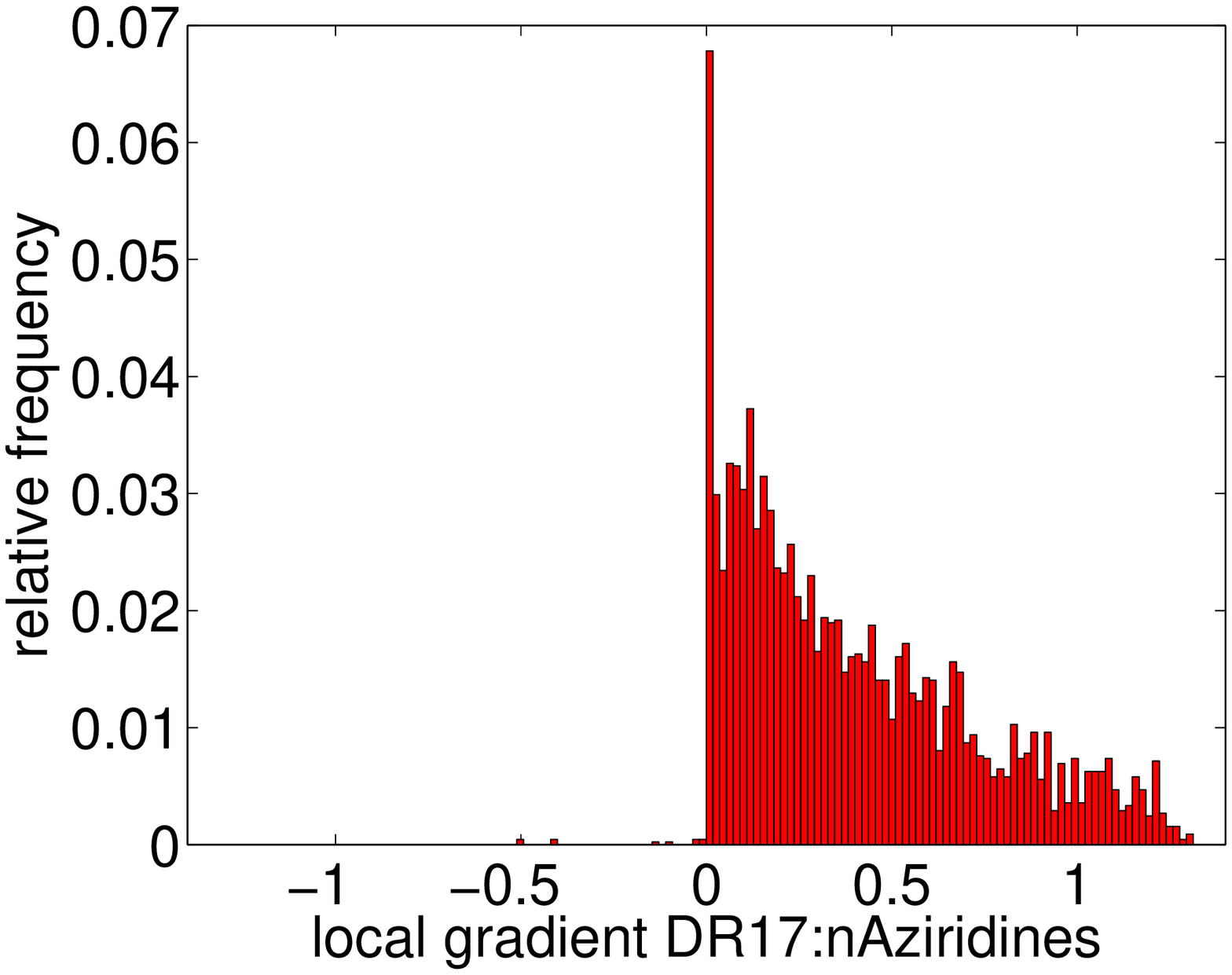}}
\subfigure[azide]{\includegraphics[width=0.3\textwidth,keepaspectratio]{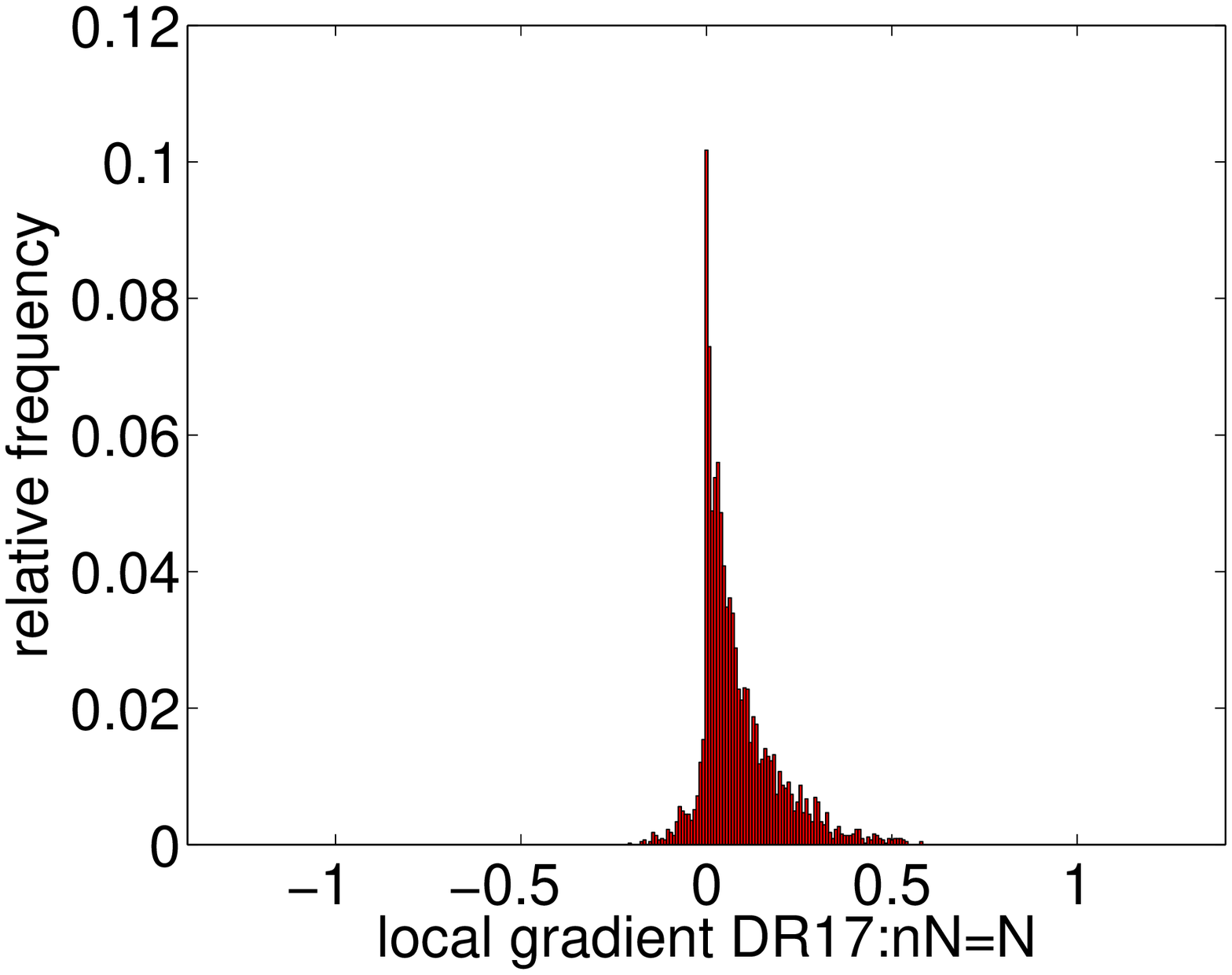}}
\subfigure[aromatic hydroxylamine]{\includegraphics[width=0.3\textwidth,keepaspectratio]{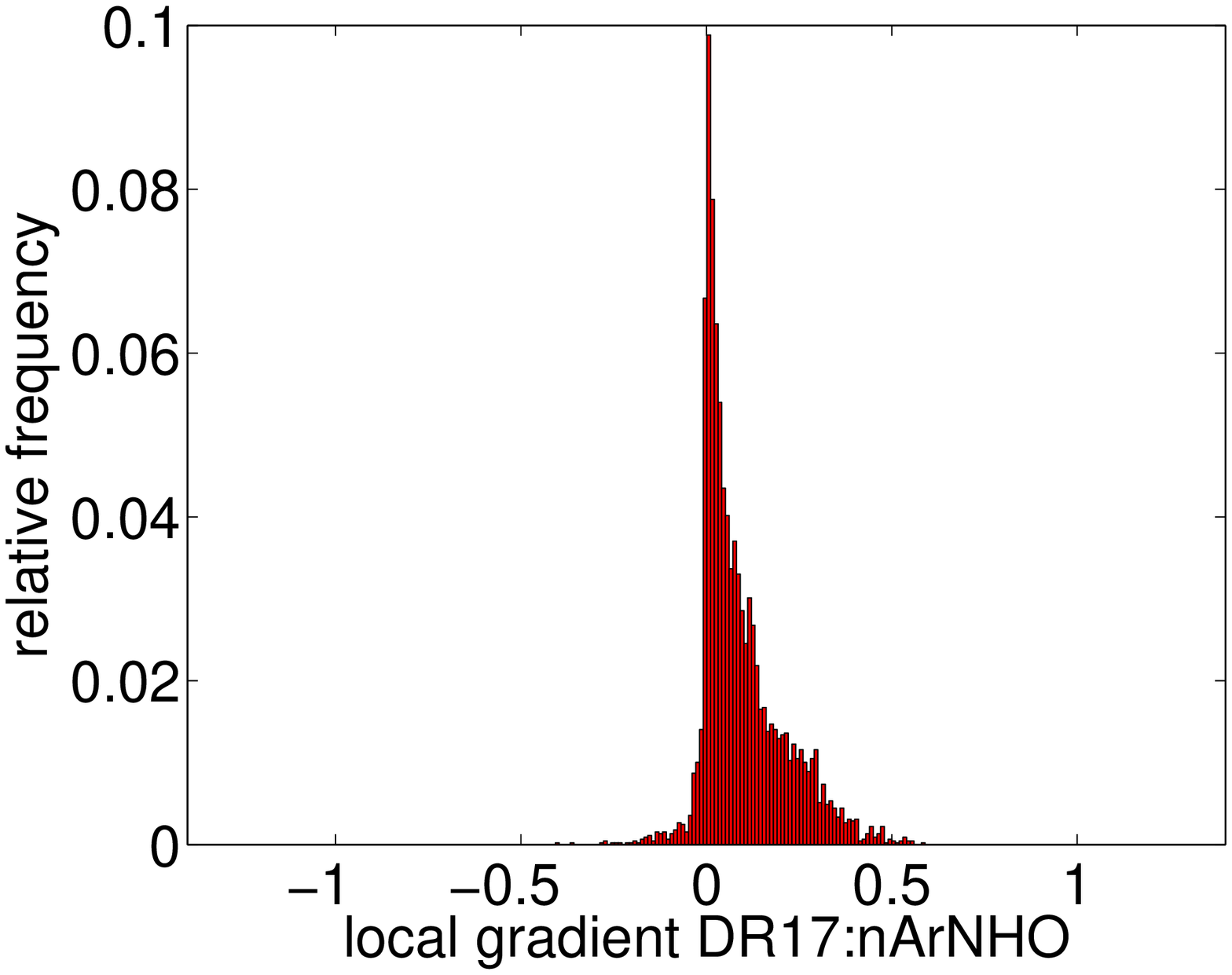}}
\subfigure[aliphatic halide]{\includegraphics[width=0.3\textwidth,keepaspectratio]{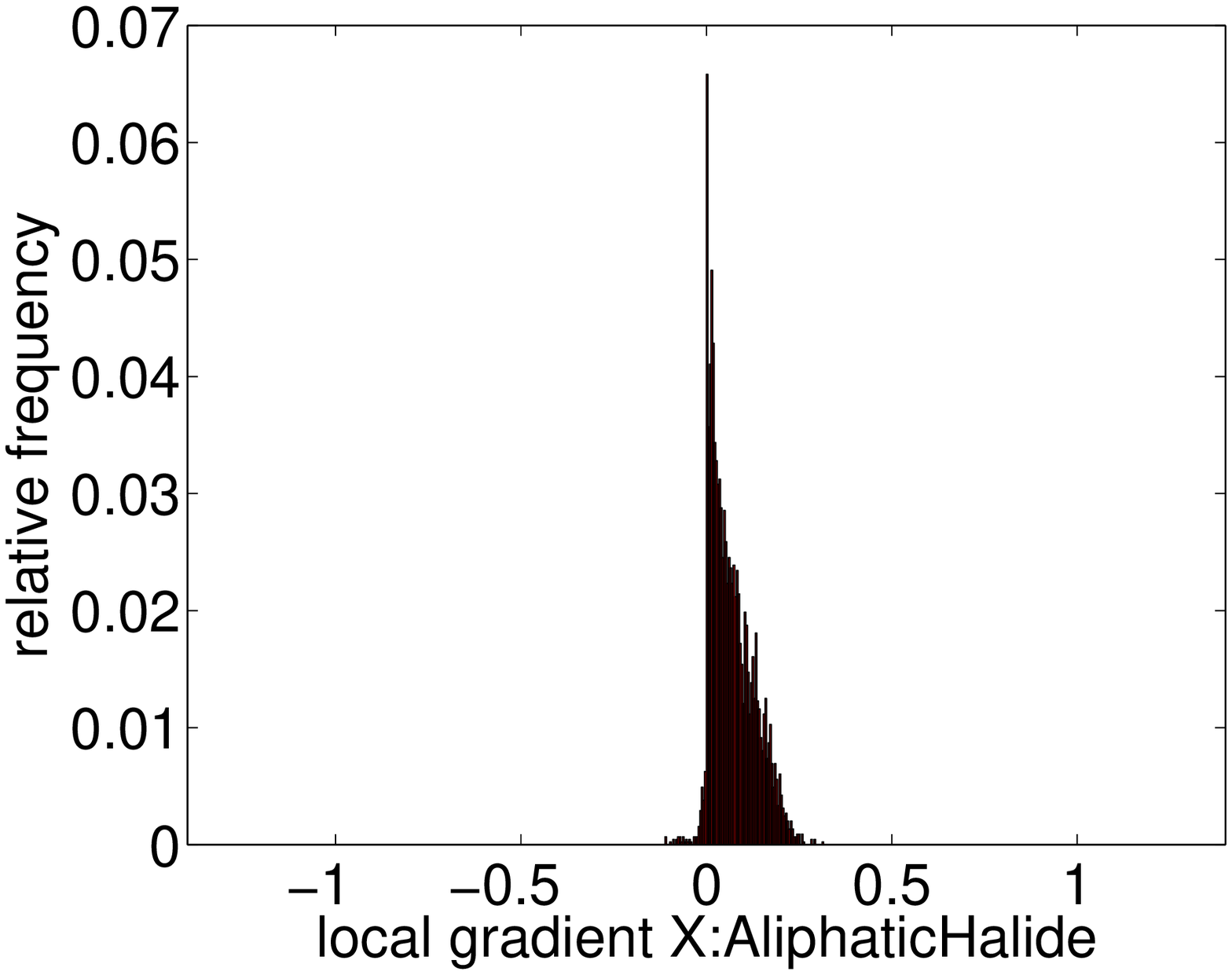}}
\caption{Distribution of local importance of selected features across the test set of 4512 compounds. Nine out of ten known toxicophores \citep*{Kazius05} indeed exhibit positive local gradients.}
\label{fig:tox_gpc_all_tox}
\end{figure}

\begin{figure}
\centering
\subfigure[sulfonamide]{\includegraphics[width=0.3\textwidth,keepaspectratio]{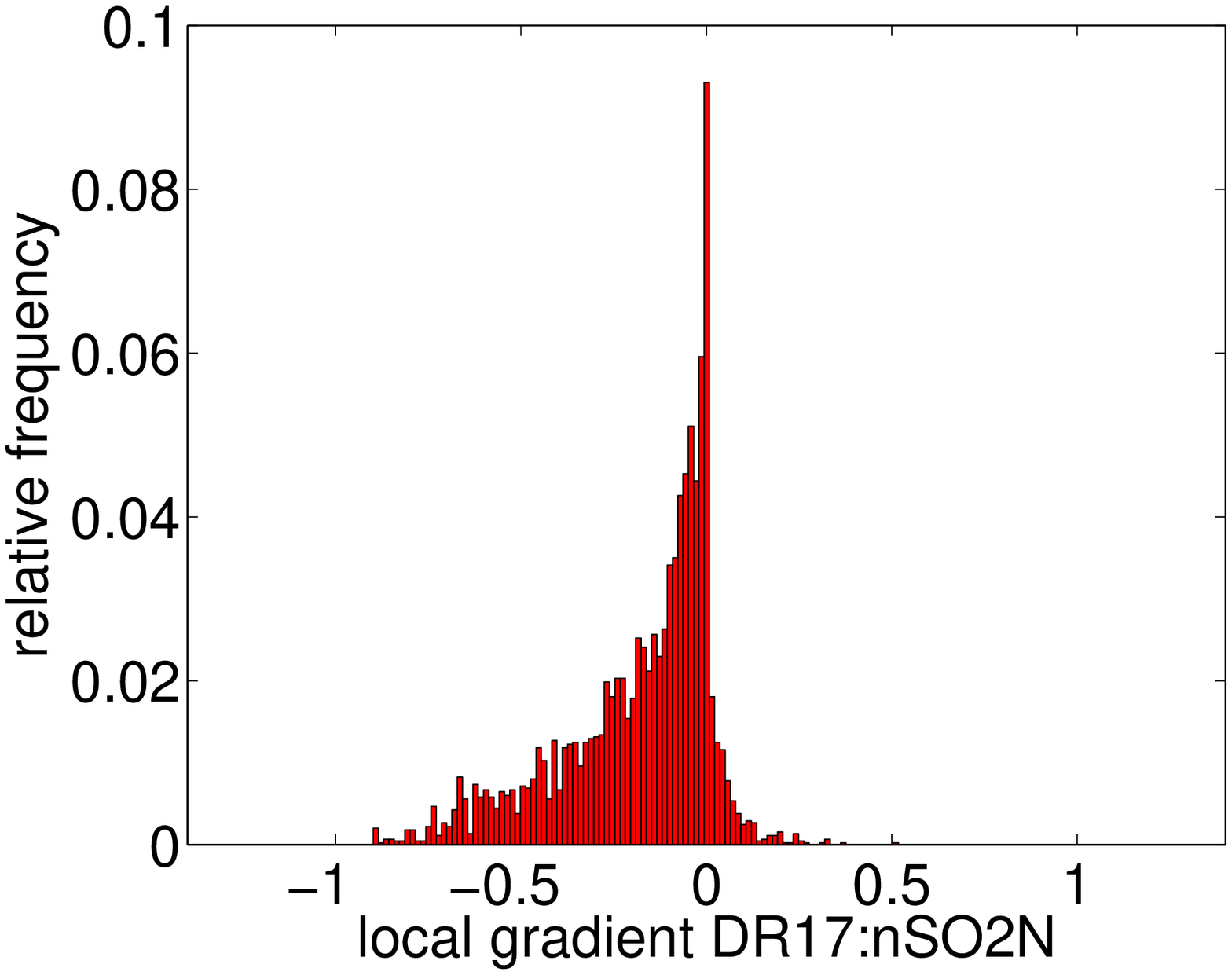}}
\subfigure[sulfonic acid]{\includegraphics[width=0.3\textwidth,keepaspectratio]{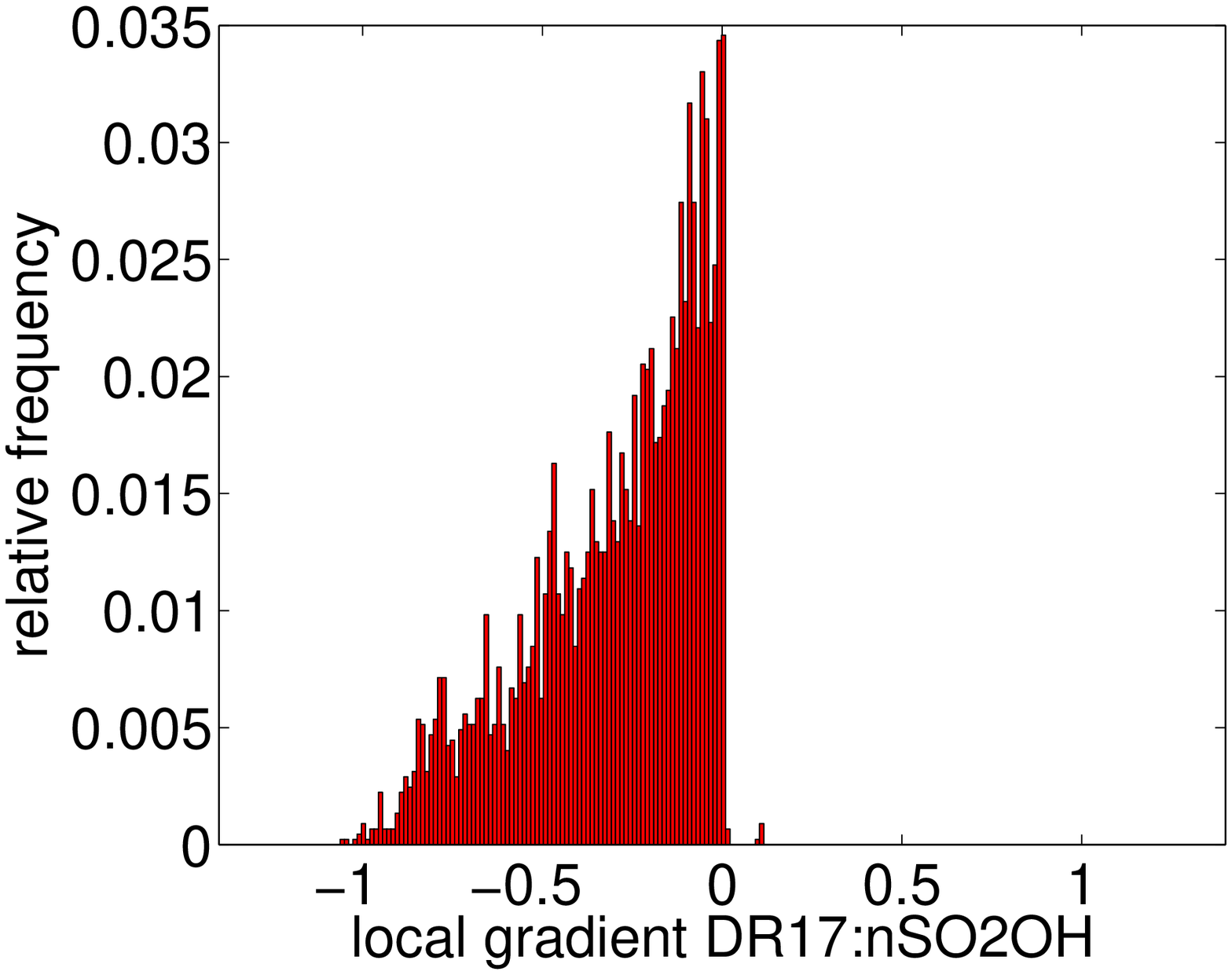}}
\subfigure[arylsulfonyl]{\includegraphics[width=0.3\textwidth,keepaspectratio]{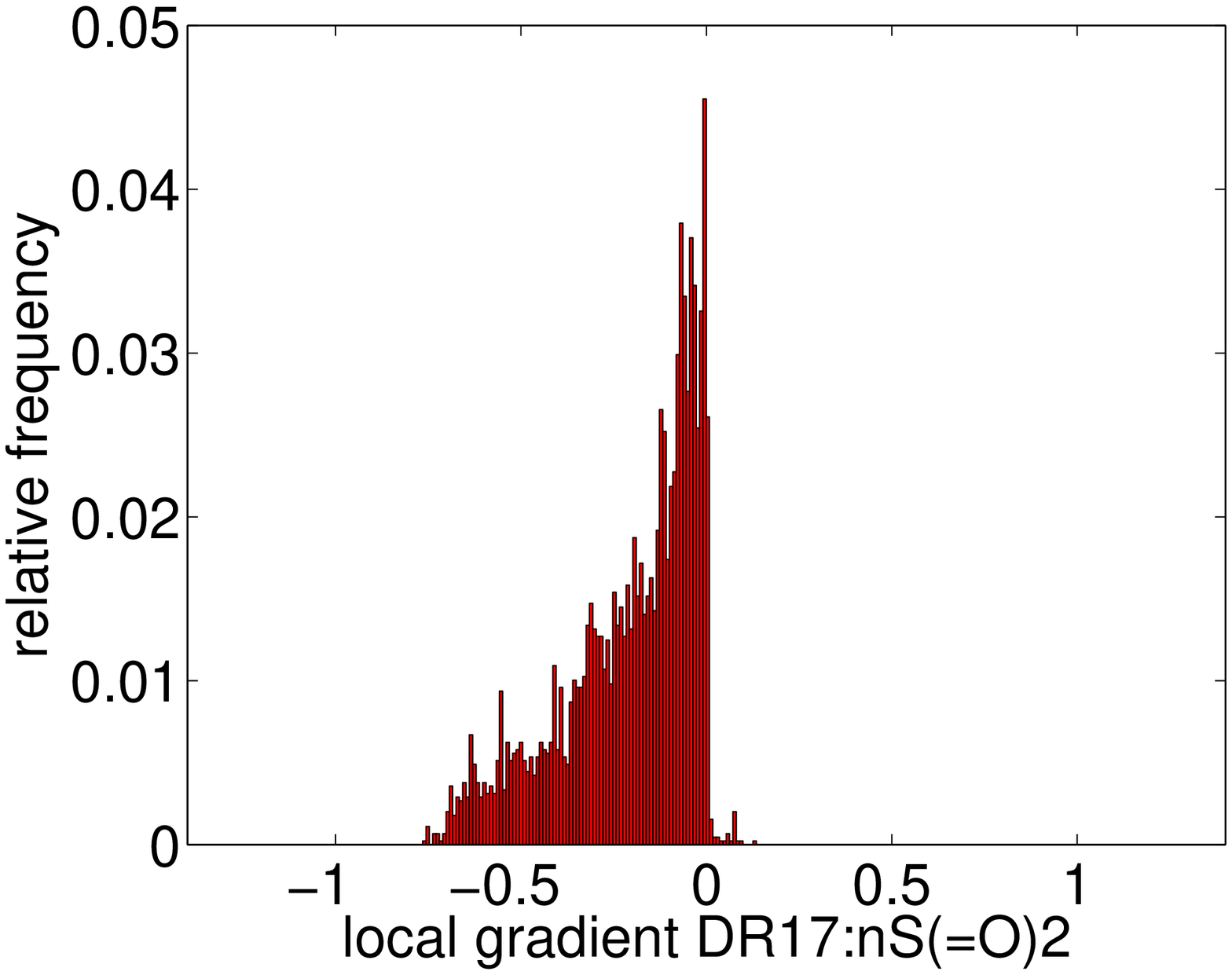}}
\subfigure[aliphatic carboxylic acid]{\includegraphics[width=0.3\textwidth,keepaspectratio]{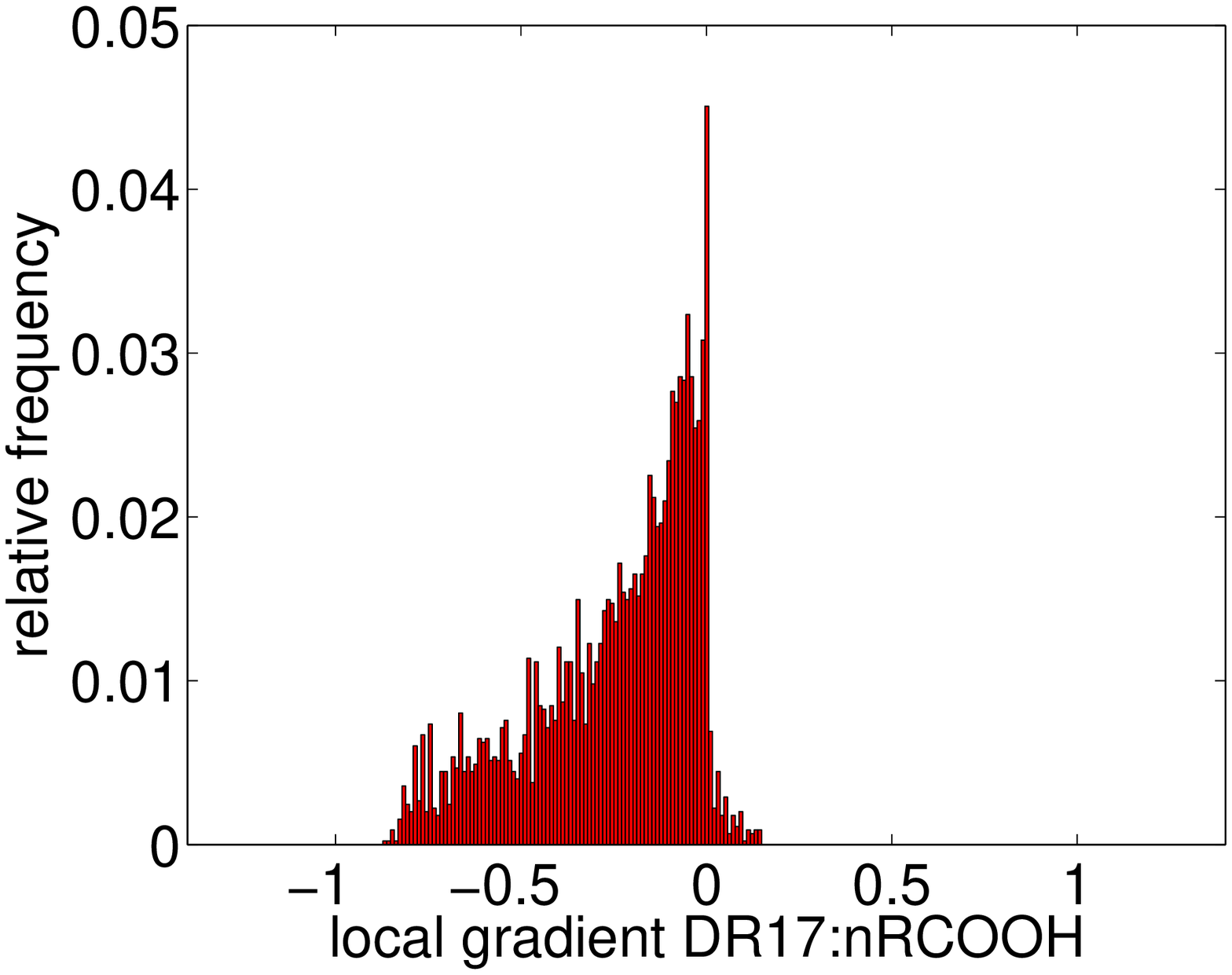}}
\subfigure[aromatic carboxylic acid]{\includegraphics[width=0.3\textwidth,keepaspectratio]{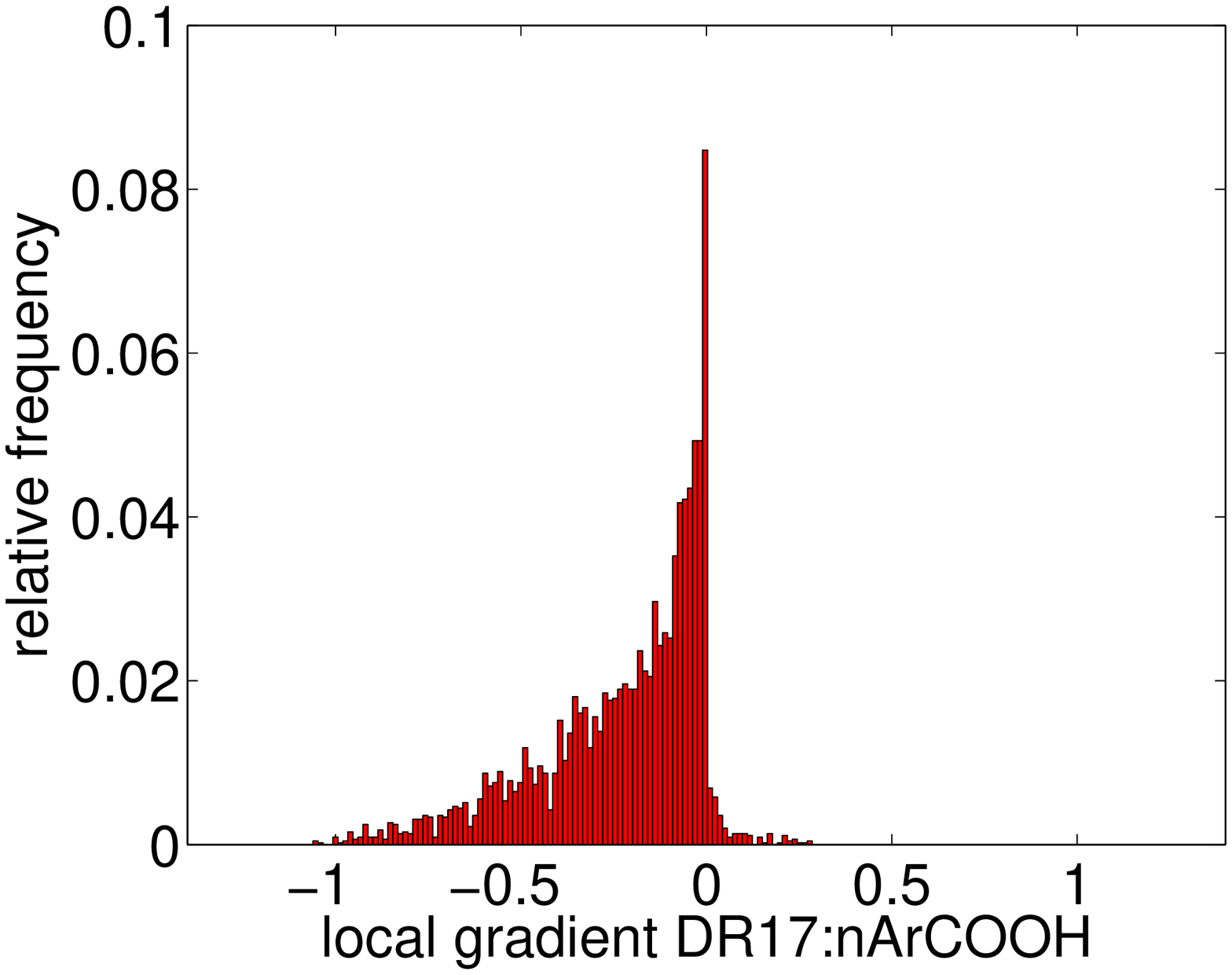}}
\caption{Distribution of local importance of selected features across the test set of 4512 compounds. All five known detoxicophores exhibit negative local gradients}
\label{fig:tox_gpc_all_detox}
\end{figure}

In Figures \ref{fig:tox_gpc_all_tox} and \ref{fig:tox_gpc_all_detox} we show the distribution of the local importance of selected features across the test set:
For each input feature we generate a histogram of local importance values, as indicated by its corresponding entry in the explanation vector of each of the 4512 test compounds. 
%
The features examined in Figure \ref{fig:tox_gpc_all_tox} are counts of substructures known to cause mutagenicity.  We show all approved ``specific toxicophores'' introduced by \citet*{Kazius05} that are also represented in the \textsc{Dragon} set of features.
The features shown in Figure \ref{fig:tox_gpc_all_detox} are known to detoxify certain toxicophores \citep*[again see][]{Kazius05}. With the exception of \ref{fig:tox_gpc_all_tox}(e) the toxicophores also have a toxifying influence according to our GPC prediction model. Feature \ref{fig:tox_gpc_all_tox}(e) seems to be mostly irrelevant for the prediction of the GPC model on the test points. In contrast the detoxicophores show overall negative influence on the prediction outcome of the GPC model.
Modifying the test compounds by adding toxicophores will increase the probability of being mutagenic as predicted by the GPC model while adding detoxicophores will decrease this predicted probability. 


So we have seen that the conclusions drawn from our explanation vectors agree with established knowledge about toxicophores and detoxicophores.  While this is reassuring, such a sanity check required existing knowledge about which compounds are toxicophores and detoxicophores and which are not.  Thus it is interesting to ask, whether we also could have \emph{discovered} that knowledge from the explanation vectors.  To answer this question
 we ranked all 142 features by the means of their local gradients\footnote{Tables resulting from this ranking are made available as a supplement to this paper and can be downloaded from the journals website.}. Clear trends result: 9 out of 10 known toxicophores can be found close the top of the list (mean rank of 19). The only exception (rank 81) is the aromatic nitrosamine feature.\footnote{This finding agrees with the result obtained by visually inspecting Figure \ref{fig:tox_gpc_all_tox}(e). We found that only very few compounds with this feature are present in the data set. Consequently, detection of this feature is only possible if enough of these few compounds are included in the training data. This was not the case in the random split used to produce the results presented above.}
This trend is even stronger for the detoxicophores: The mean rank of these five features is 138 (out of 142), i.e. they consistently exhibit the largest negative local gradients. Consequently, the established knowledge about toxicophores and detoxicophores could indeed have been \textit{discovered} using our methodology.

In the following paragraph we will discuss steroids\footnote{Steroids are natural products and occur in humans, animals and plants. They have a characteristic backbone containing four fused carbon-rings. Many hormones important to the development of the human body are steroids, including androgens, estrogens, progestagens, cholesterol and natural anabolics. These have been used as starting points for the development of many different drugs, including the most reliable contraceptives currently on the market.}
as an example of an important compound class for which the meaning of features differs from this global trend, so that local explanation vectors are needed to correctly identify relevant features.

\begin{figure}
\centering
\subfigure[epoxide feature: steroid vs. non-steroid]{\includegraphics[width=0.49\textwidth,keepaspectratio]{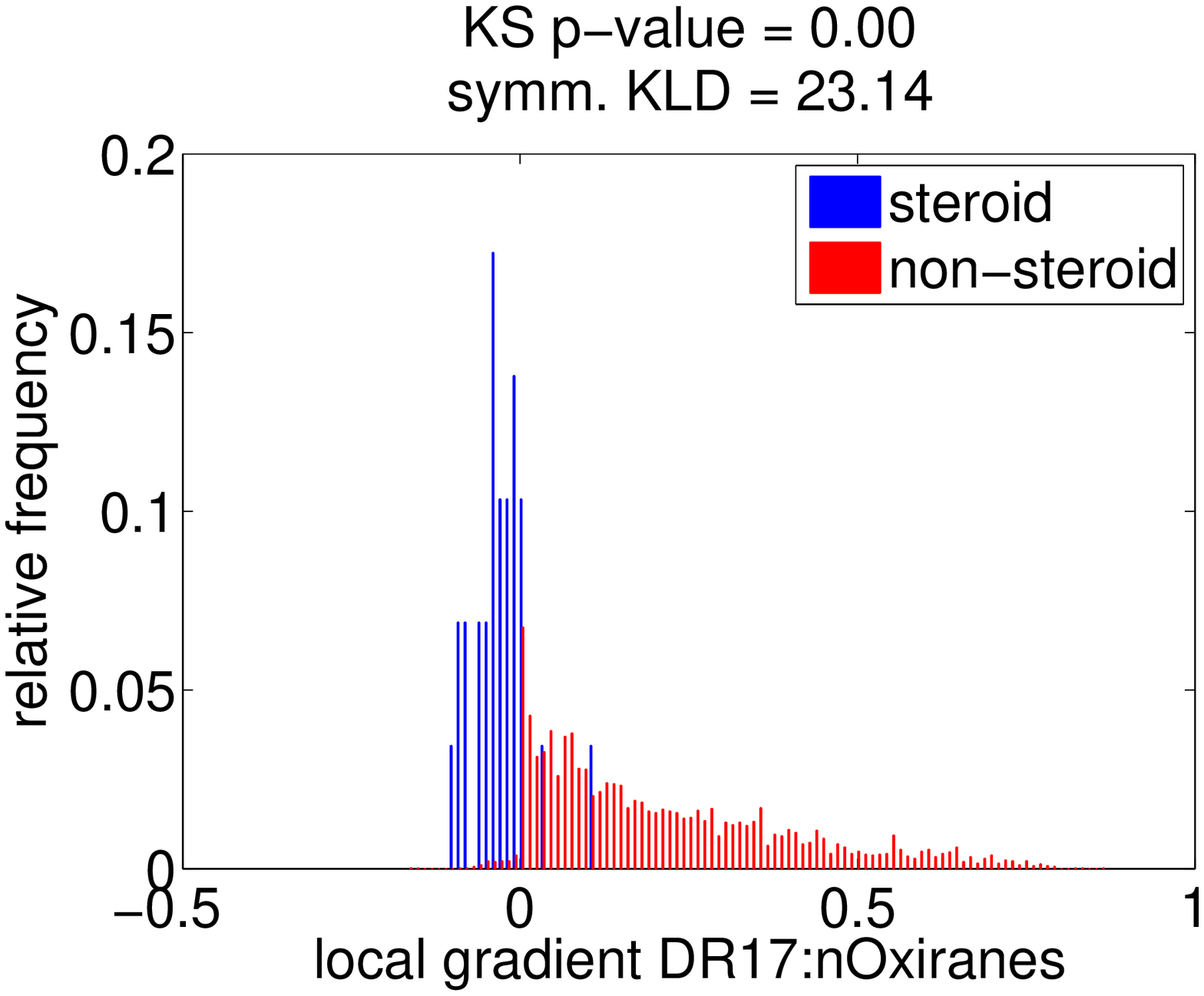}}
\subfigure[epoxide feature: random compounds vs. the rest]{\includegraphics[width=0.49\textwidth,keepaspectratio]{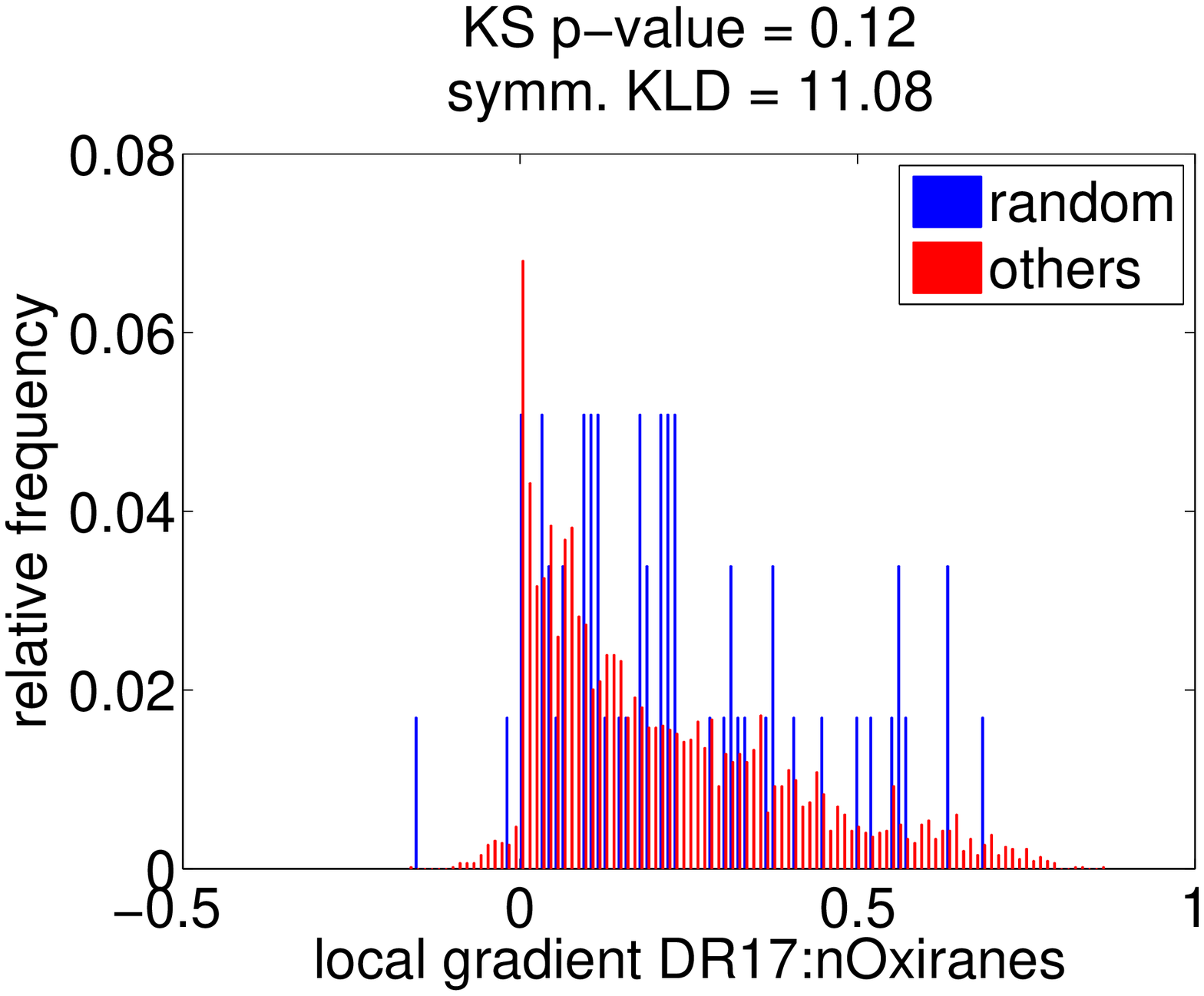}}
\subfigure[aliphatic nitrosamine feature: steroid vs. non-steroid]{\includegraphics[width=0.49\textwidth,keepaspectratio]{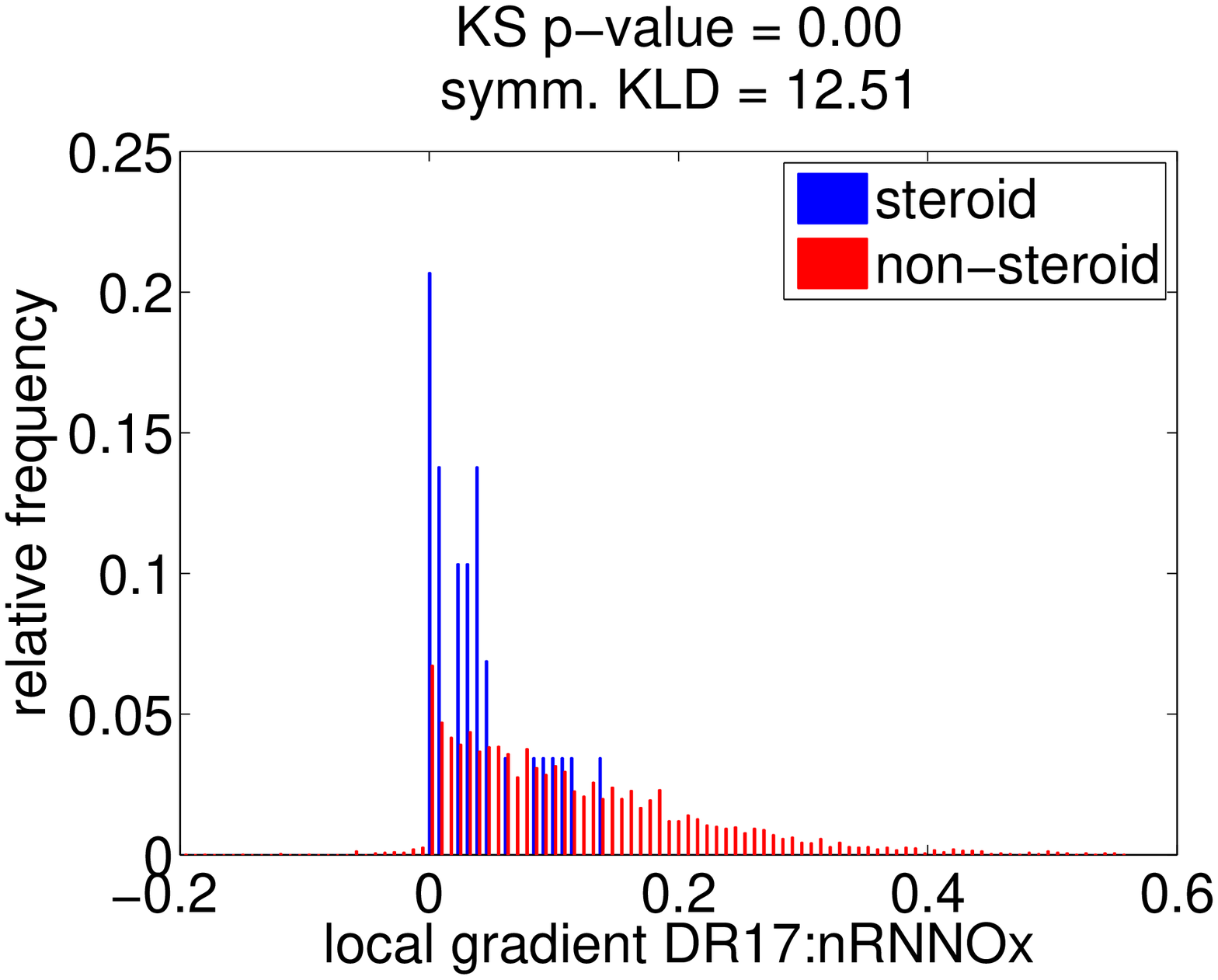}}
\subfigure[aliphatic nitrosamine feature: random compounds vs. the rest]{\includegraphics[width=0.49\textwidth,keepaspectratio]{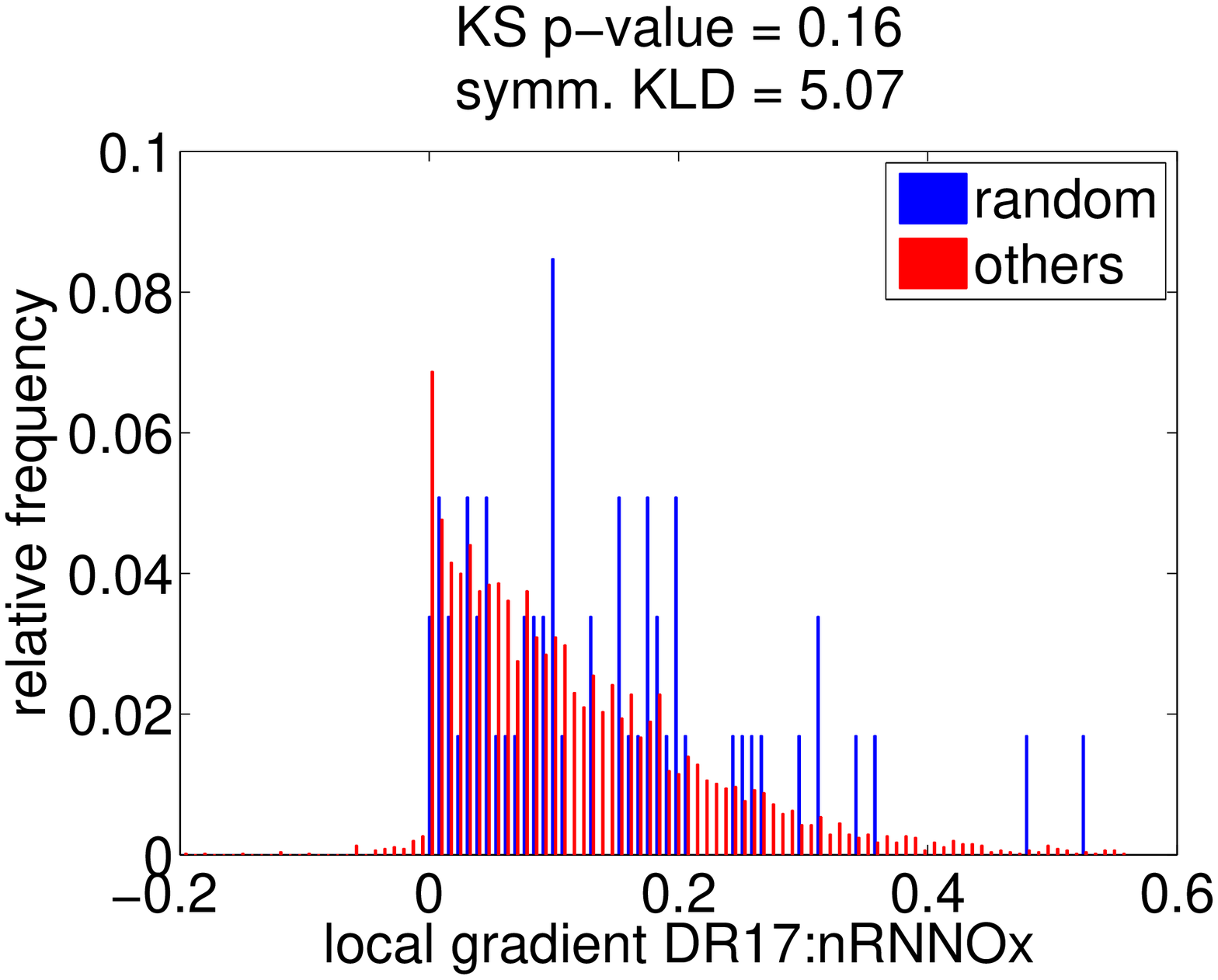}}
\caption{The local distribution of feature importance to steroids and random non-steroid compounds significantly differs for two known toxicophores. The small local gradients found for the steroids (shown in blue) indicate that the presence of each toxicophore is irrelevant to the molecules toxicity. For non-steroids (shown in red) the known toxicophores indeed exhibit positive local gradients.}
\label{fig:tox_gpc_steroid}
\end{figure}

Figure \ref{fig:tox_gpc_steroid} displays the difference in relevance of epoxide (a) and aliphatic nitrosamine (c) substructures for the predicted mutagenicity of steroids and non-steroid compounds. For comparison we also show the distributions for compounds chosen at random from the test set (b,d). 
Each subfigure contains two measures of (dis-)similarity for each pair of distributions. The p-value of the Kolmogorov-Smirnoff test (KS) gives the probability of error when rejecting the hypothesis that both relative frequencies are drawn from the same underlying distribution. The symmetrized Kullback-Leibler divergence (KLD) gives a metric of the distance between the two distributions.\footnote{Symmetry is achieved by averaging the two Kullback-Leibler divergences: $\frac{KL(P1,P2)+KL(P2,P1)}{2}$, cf. \cite*{Johnson00symmetrizingthe}. To prevent zero-values in the histograms which would lead to infinite KL distances, an $\varepsilon > 0$ has been added to each bin count.}
While containing epoxides generally tends to make molecules mutagenic (see discussion above), we do not observe this effect for steroids: In Figure \ref{fig:tox_gpc_steroid}(a), almost all epoxide containing non-steroids exhibit positive gradients, thereby following the global distribution of epoxide containing compounds as shown in Figure \ref{fig:tox_gpc_all_tox}(f). In contrast, almost all epoxide containing steroids exhibit gradients just below zero. ``Immunity'' of steroids to the epoxide toxicophore is an established fact and has first been discussed by \citet*{Glatt_Jung_Oesch_1983}.
This peculiarity in chemical space is clearly exhibited by the local explanation given by our approach. For aliphatic nitrosamine, the situation in the GPC model is less clear but still the toxifying influence seems to be less in steroids than in many other compounds. To our knowledge, this phenomenon has not yet been discussed in the pharmaceutical literature.

In conclusion, we can learn from the explanation vectors that:
\begin{itemize}
\item toxicophores tend to make compounds mutagenic (class 1)
\item detoxicophores tend to make compounds non-mutagenic (class 0)
\item steroids are immune to the presence of some toxicophores (epoxide, possibly also aliphatic nitrosamine)
\end{itemize}

%% file: related_work.tex
\section{Related Work}
\label{sec:related_work}

Assigning potentially different explanations to individual data points 
distinguishes our approach from conventional feature extraction
methods that extract global features that are relevant for all data points,
i.e.~those features that allow to achieve a small overall prediction
error.  Our notion of explanation is not related to the prediction
error, but only to the label provided by the prediction
algorithm.  Even though the error is large, our framework is able to
answer the question {\em why} the algorithm has decided on a data point
the way it did.

The explanation vector proposed here is similar in spirit to
sensitivity analysis which is common to various areas
of information science. A classical example is the outlier sensitivity
in statistics \citep*{Hampel}.  In this case, the effects of
removing single data points on estimated parameters are evaluated by
an influence function.  If the influence for a data point is
significantly large, it is detected as an outlier and should be
removed for the following analysis. In regression problems, leverage
analysis is a procedure along similar lines. It detects leverage points
which have potential to give large impact on the estimate of the
regression function.  In contrast to the influential points
(outliers), removing a leverage sample may not actually change the
regressor, if its response is very close to the predicted value.
E.g.~for linear regression the samples whose inputs are far from the
mean are the leverage points. Our framework of explanation vectors considers a different view. It describes the influence of {\em moving} single
data points locally and it thus answers the question which directions are
locally most influential to the prediction. The explanation
vectors are used for extracting sensitive features which are relevant
to the prediction results, rather than detecting/eliminating the
influential samples.


In recent decades, explanation of results by expert systems have
been an important topic in the AI community. Especially, for those
based on Bayesian belief networks, such explanation is crucial in
practical use. In this context sensitivity analysis has also been used
as a guiding principle \citep*{Horvitz}. There the influence is evaluated
by removing a set of variables (features) from evidences
and the explanation is constructed from those variables which affect
inference (relevant variables).  For example, \citet*{Suermondt92} measures the cost of omitting a single
feature $E_i$ by the cross-entropy
\[
  H^{-}(E_i) = H(p(D|E); P(D|E\backslash E_i)\,) =
  \sum_{j=1}^N P(d_j|E) \log \frac{P(d_j|E)}{p(d_j|E\backslash E_i)},
\]
where $E$ denotes evidences and $D = \left(d_1,\dots,d_N\right)^T$ is the target variable.  The cost
of a subset $F \subset E$ can be defined similarly.  This line of
research is more connected to our work, because explanation can
depend on the assigned values of the evidences $E$, and is thus 
local.

Similarly \citet*{RobKon08} and \citet*{StrKon08} try to explain the decision of trained kNN-, SVM- and ANN-models for individual instances by measuring the difference in their prediction with sets of features omitted. The cost of omitting features is evaluated as the information difference, the log-odds ratio or the difference of probabilities between the model with knowledge about all features and with omissions respectively. To know what the prediction would be without the knowledge of a certain feature the model is retrained for every choice of features whose influence is to be explained. To save the time of combinatorial training \citet*{RobKon08} propose to use neutral values which have to be estimated by a known prior distribution of all possible parameter values. As a theoretical framework for considering feature interactions, \citet*{StrKon08} propose to calculate the differences between model predictions for every choice of feature subset.

For multi-layer perceptrons \citet*{FerCle02} measure the importance of individual input variables on clusters of test points. Therefore the change in the model output is evaluated for the change of a single input variable in a chosen interval while all other input variables are fixed. \citet*{DBLP:conf/f-egc/LemaireF07} use a similar approach on an instance by instance basis. By considering each input variable in turn there is no way to measure input feature interactions on the model output \citep*[see][]{LeCunBottou98}.

The principal differences between our approach and these frameworks are:
(i) We consider continuous features and no structure among them is required, while some other frameworks start from binary features and may require discretization steps with the need to estimate parameters for it.
(ii) We allow changes in any direction, i.e. any weighted combination of variables, while
other approaches only consider one feature at a time or the omission of a set of variables.

%% file: discussion.tex
\section{Discussion}
\label{sec:discussion}

\begin{figure}
  \centering
  \includegraphics[width=.7\textwidth]{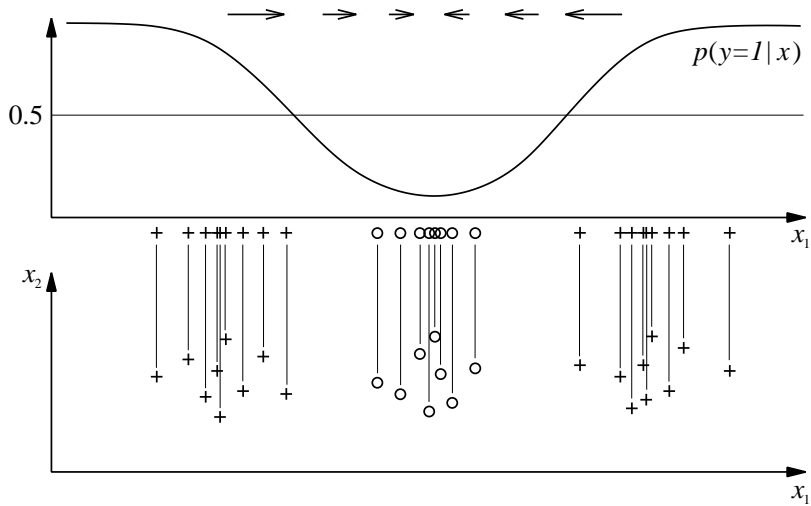}
  \caption{$\zeta(x)$ is the zero vector in the middle of the cluster
  in the middle.}
  \label{fig:zero}
\end{figure}

By now we have shown that our methods for calculating / estimating explanation
vectors are useful in a variety of situations. In the following we
discuss their limitations.
\paragraph{What can we do, if the derivative is zero?}  This situation
is depicted in Figure~\ref{fig:zero}.  In the lower panel we see a
two-dimensional data set consisting of three clusters.  The middle
cluster has a different class than the clusters on the left and on the
right.  Relevant for the classification is only the horizontal
coordinate (i.e.~$x_1$).  The upper panel shows the projected data and
a representative slice of $\zeta(x)$.  However, the explanation $\zeta(x)$ for the center point of the middle cluster
is the zero vector, because at that point $p(Y\eq 1|X\eq x)$ is
maximal.  What can we do in such situations?  Actually, the
(normalized) explanation vector is derived from the following
optimization problem for finding the locally most influential
direction: $\mathop{\rm argmax}_{\| \varepsilon \| = 1} \left\{
  p(Y\neq g^*(x_0)| X = x_0+\varepsilon) - p(Y\neq g^*(x_0)| X = x_0)
\right\}$.  In case that the first derivative of the above criterion
is zero, its Taylor expansion starts from the second order term, which
is a quadratic form in its Hessian matrix.  In the example data set
with three clusters, the explanation vector is constant along the
second dimension.  The most {\em interesting} direction is given by
the eigenvector corresponding to the largest eigenvalue of the
Hessian.  This direction will be in our example along the first
dimension.  Thus, we can learn from the Hessian that the first
coordinate is relevant for the classification, but we do not obtain an
orientation for it.  Instead it means that both directions (left and
right) will influence the classification.
However, if the conditional distribution $P(Y=1\given X=x)$ is flat in
some regions, no meaningful explanation can be obtained by the 
gradient-based approach with the remedy mentioned above. 
Practically, by using Parzen
window estimators with larger widths, the explanation vector can
capture coarse structures of the classifier at the points which are not so far 
from the boarders. In \ref{sec:appendix_estimating_fit} we give an illustration of this point. In the future, we would like to work on global
approaches, e.g. based on distances to the boarders, or extensions
of the approach by \cite*{RobKon08}. Since these
procedures are expected to be computationally demanding, our proposal is
useful in practice, in particular for probabilistic classifiers.

\paragraph{Does our framework generate different explanations for different
  prediction models?} When using the local gradient of the model prediction directly as in Definition \ref{def:ev_cls} and Section \ref{sec:tox_gpc}, the explanation follows the given model precisely by definition. For the estimation framework this depends on whether the different classifiers
classify the data differently.  In that case the explanation vectors
will be different, which makes sense, since they should explain the
classifier at hand, even if its estimated labels were not all correct.
On the other hand, if the different classifiers agree on all labels,
the explanation will be exactly equal.

\paragraph{Which implicit limitations do analytical gradients inherit from Gaussian Process models?}
A particular phenomenon can be observed at the boundaries of the training data: Far from the training data, Gaussian Process Classification models predict a probability of 0.5 for the positive class. When querying the model in an area of the feature space where predictions are negative, and one approaches the boundaries of the space populated with training data, explanation vectors will point away from any training data and therefore also away from areas of positive prediction. This behavior can be observed in Figure \ref{fig:toy_gpc}(d), where unit length vectors indicate the direction of explanation vectors. In the right hand side corner, arrows point away from the triangle. However, we can see that the length of these vectors is so small, that they are not even visible in Figure \ref{fig:toy_gpc}(c).
Consequently, this property of GPC models does not pose a restriction for identifying the locally most influential features by investigating the features with the highest absolute values in the respective partial derivatives, as shown in Section \ref{sec:tox_gpc}.

\paragraph{Stationarity of the data.}
Since explanation vectors are defined as local gradients of the model prediction (see Definition \ref{def:ev_cls}), no assumption on the data is made: The local gradients follow the predictive model in any case.
If, however, the model to be explained assumes stationarity of the data, the explanation vectors will inherit this limitation and reflect any shortcomings of the model (e.g. when the model is applied to non-stationary data).
Our method for estimating explanation vectors, on the other hand, assumes stationarity of the data. 

When modeling data that is in fact non-stationary, appropriate measures to deal with such data sets should be taken. One option is to separate the feature space into stationary and non-stationary parts using Stationary Subspace Analysis as introduced by \citet*{Bunau09}. For further approaches to data set shift see \citet*{CovShiftPaul2007}, \citet*{Sugiyama07} and the book by \citet*{DatasetShift2009}.

%% file: conclusion.tex
\section{Conclusion}
\label{sec:conclusion}
This paper proposes a method that sheds light into the black boxes of nonlinear classifiers.  In other words, we introduce a method that can explain the local decisions taken by arbitrary (possibly) nonlinear classification algorithms.  In a nutshell, the estimated explanations are local gradients that characterize how a data point has to be moved to change its predicted label.  For models where such gradient information cannot be calculated explicitly, we employ a probabilistic approximate mimic of the learning machine to be explained.

To validate our methodology we show how it can be used to draw new conclusions on how the various Iris flowers in Fisher's famous data set are different from each other and how to identify the features with which certain types of digits 2 and 8 in the USPS data set can be distinguished.  Furthermore, we applied our method to a challenging drug discovery problem.  The results on that data fully agree with existing domain knowledge, which was not available to our method.  Even local peculiarities in chemical space (the extraordinary behavior of steroids) was discovered using the local explanations given by our approach.

Future directions are two-fold: First we believe that our method will find its way into the tool boxes of practitioners who not only want to automatically classify their data but who also would like to understand the learned classifier. Thus using our explanation framework in computation biology \citep*[see][]{SonZiePhiRae08} and
in decision making experiments in psychophysics \citep*[e.g.][]{Kienzle09} 
seems most promising. The second direction is to generalize our approach to other prediction problems such as regression.

%% file: acknowledgments.tex
\subsection*{Acknowledgments}
This work was supported in part by the FP7-ICT Programme of the European Community, under the PASCAL2 Network of Excellence, ICT-216886 and by DFG Grant MU 987/4-1. We would like to thank Andreas Sutter, Antonius Ter Laak, Thomas Steger-Hartmann and Nikolaus Heinrich for publishing the Ames mutagenicity data set \citep*{OurTox}.

%% file: appendix.tex
\section{Appendix}
\label{sec:appendix}

\subsection{Illustration of direct local gradients}
\label{sec:appendix_direct}
In the following we give some illustrative examples of our method to explain models using local gradients. Since the explanation is derived directly from the respective model, it is interesting to investigate its acuteness depending on different model parameters and in instructive scenarios. We examine the effects that local gradients exhibit when choosing different kernel functions, when introducing outliers, and when the classes are not linearly separable locally.

\subsubsection{Choice of kernel function}
\label{sec:appendix_direct_kernel}
\begin{figure}[ht]
\centering
\subfigure[linear model\label{fig:toy_gpc_comparison_linear_model}]{\includegraphics[width=0.49\textwidth, keepaspectratio]{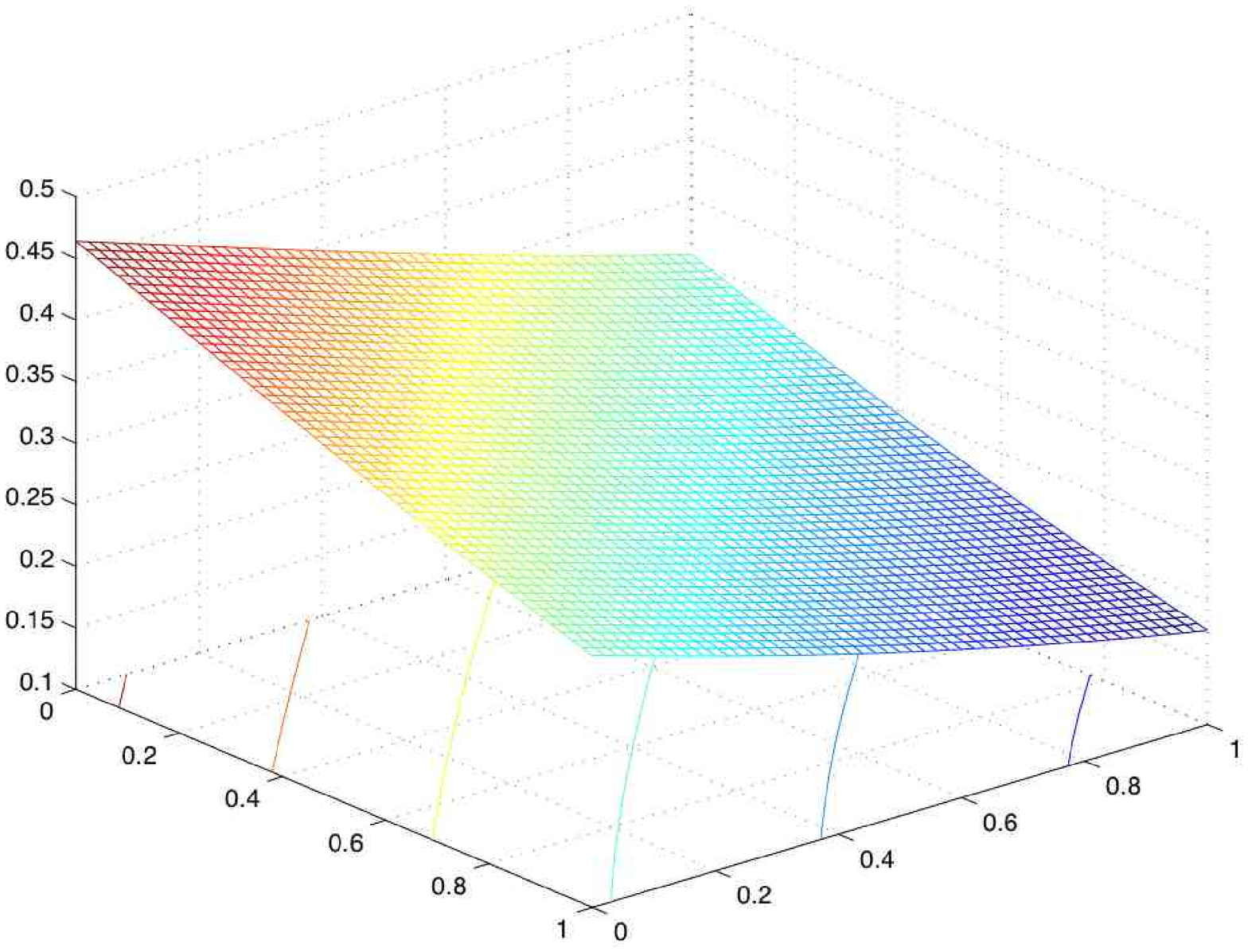}}
\subfigure[linear explanation\label{fig:toy_gpc_comparison_linear_explanation}]{\includegraphics[width=0.49\textwidth, keepaspectratio]{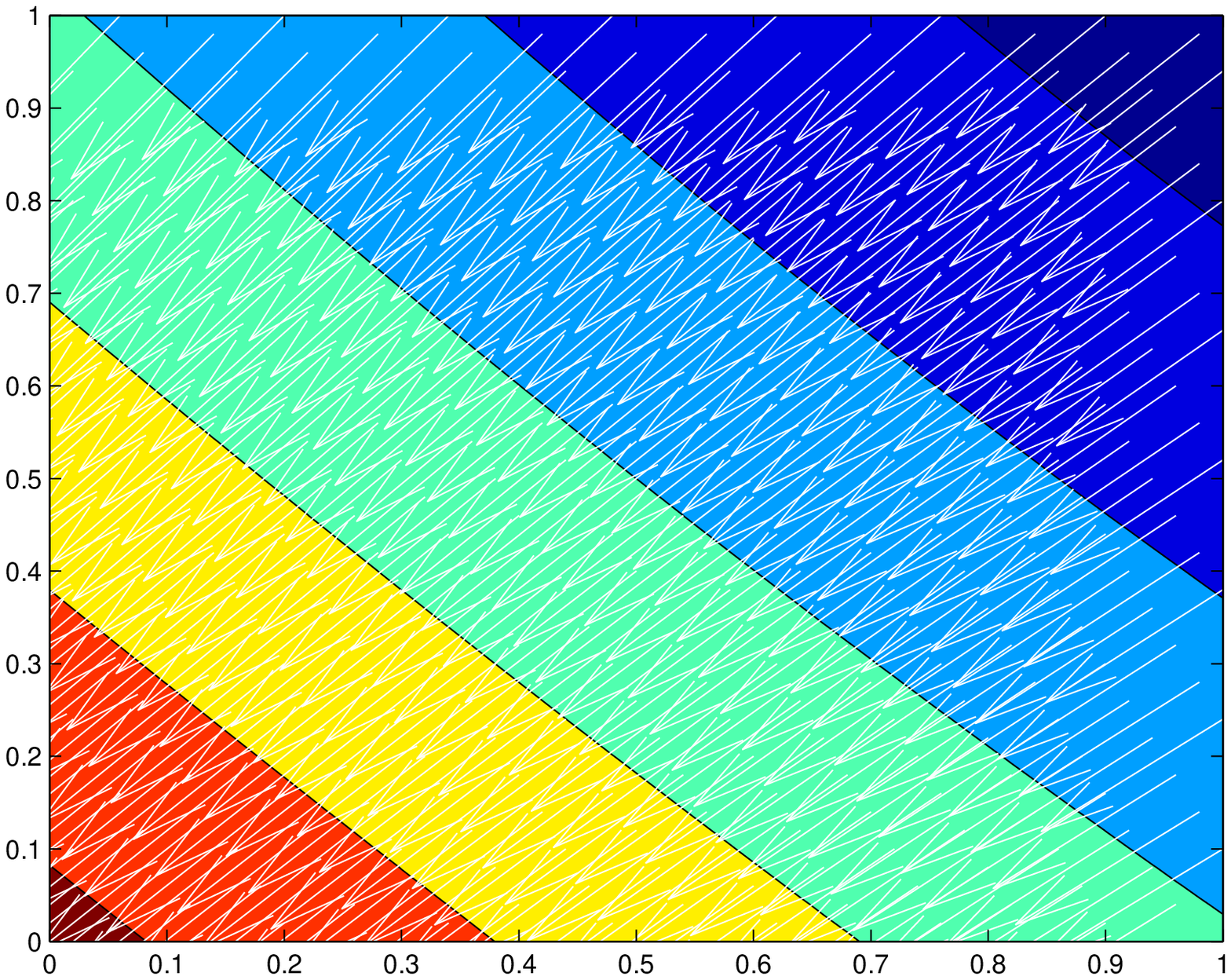}}
\subfigure[rational quadratic model\label{fig:toy_gpc_comparison_ratquad_model}]{\includegraphics[width=0.49\textwidth, keepaspectratio]{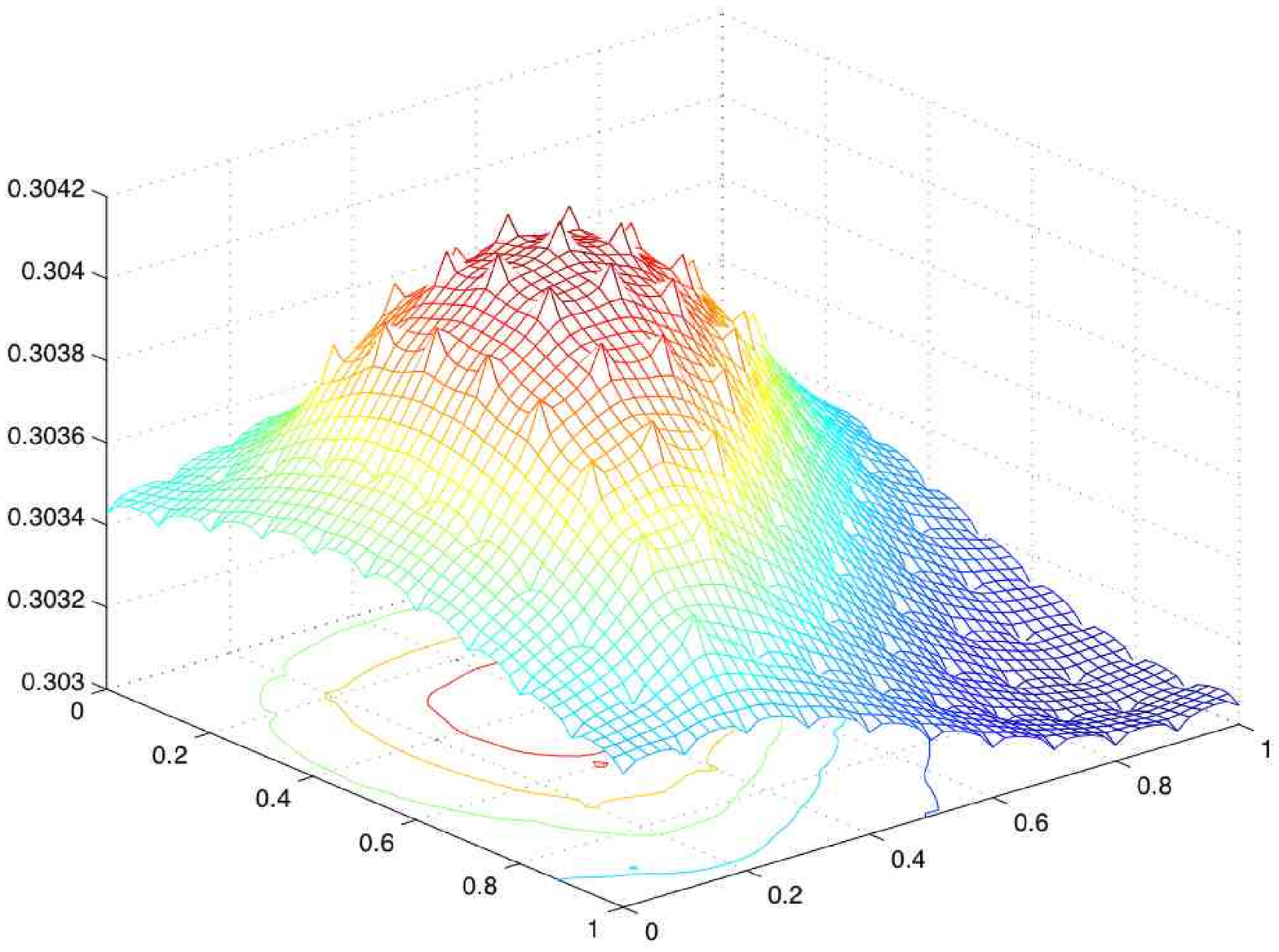}}
\subfigure[rational quadratic explanation\label{fig:toy_gpc_comparison_ratquad_explanation}]{\includegraphics[width=0.49\textwidth, keepaspectratio]{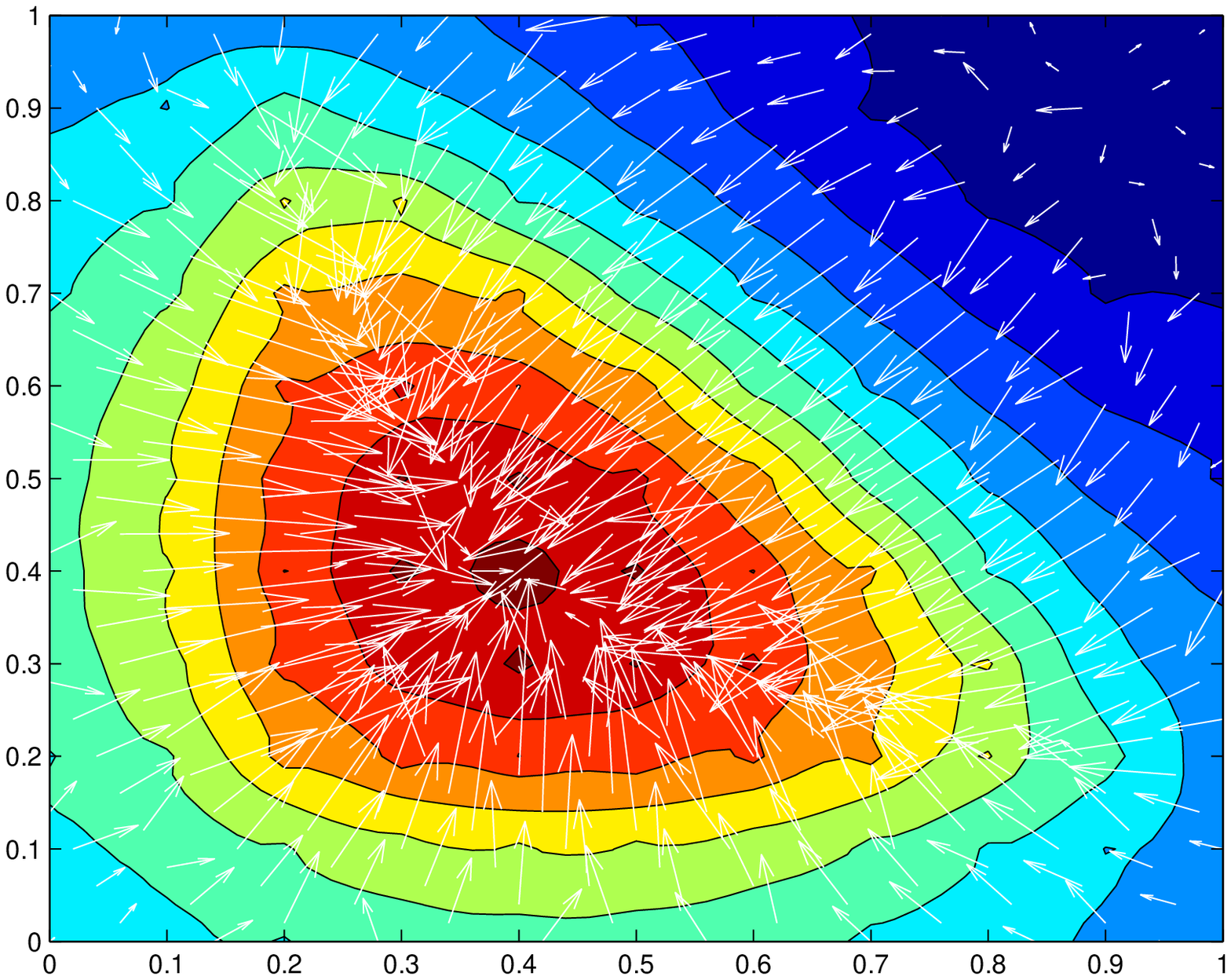}}
\caption{The effect of different kernel functions to the local gradient explanations}
\label{fig:toy_gpc_comparison}
\end{figure}

Figure \ref{fig:toy_gpc_comparison} shows the effect of different kernel functions on the triangle toy data from Figure \ref{fig:toy_gpc} in Section \ref{sec:definition}. The following observations can be made:
\begin{itemize}
 \item In any case note that the local gradients explain the model, which in turn may or may not capture the true situation.
\item In Subfigure \ref{fig:toy_gpc_comparison_linear_model} the linear kernel leads to a model which fails to capture the non-linear class separation. This model misspecification is reflected by the explanations given for this model in Subfigure \ref{fig:toy_gpc_comparison_linear_explanation}.
\item The rational quadratic kernel is able to more accurately model the non-linear separation. In Subfigure \ref{fig:toy_gpc_comparison_ratquad_model} a non-optimal degree parameter has been chosen for illustrative purposes. For other parameter values the rational quadratic kernel leads to similar results as the RBF kernel function used in Figure \ref{fig:toy_gpc}.
\item The explanations in Subfigure \ref{fig:toy_gpc_comparison_ratquad_explanation} obtained for this model show local perturbations at the small ``bumps'' of the model but the trends towards the positive class are still clear. As previously observed in Figure \ref{fig:toy_gpc}, the explanations make clear that both features interact at the corners and on the hypotenuse of the triangle class.
\end{itemize}

\subsubsection{Outliers}
\label{sec:appendix_direct_outliers}
\begin{figure}[ht]
\centering
\subfigure[outliers in classes\label{fig:toy_gpc_outliers_object}]{\includegraphics[width=0.49\textwidth, keepaspectratio]{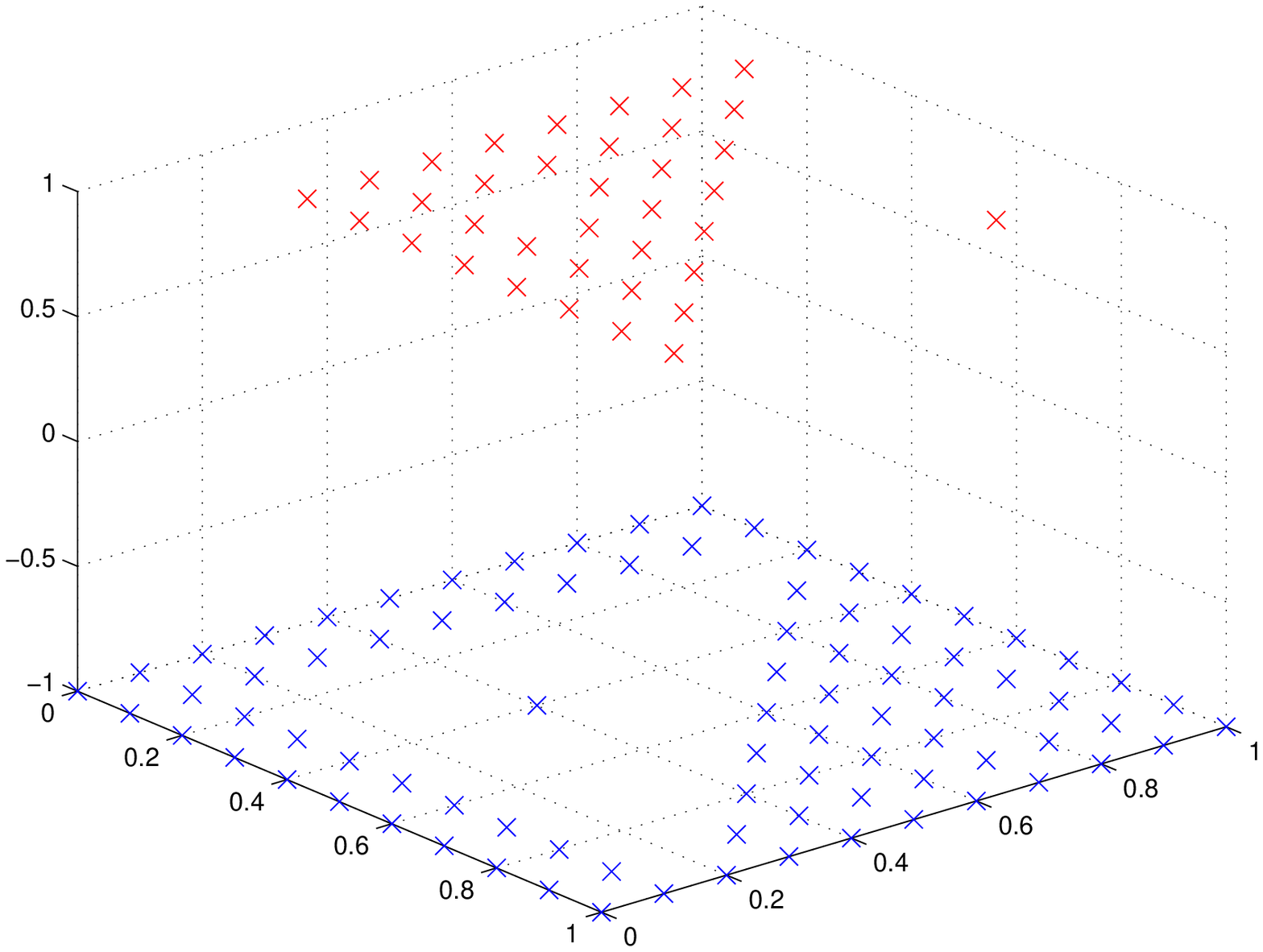}}
\subfigure[outliers in model\label{fig:toy_gpc_outliers_model}]{\includegraphics[width=0.49\textwidth, keepaspectratio]{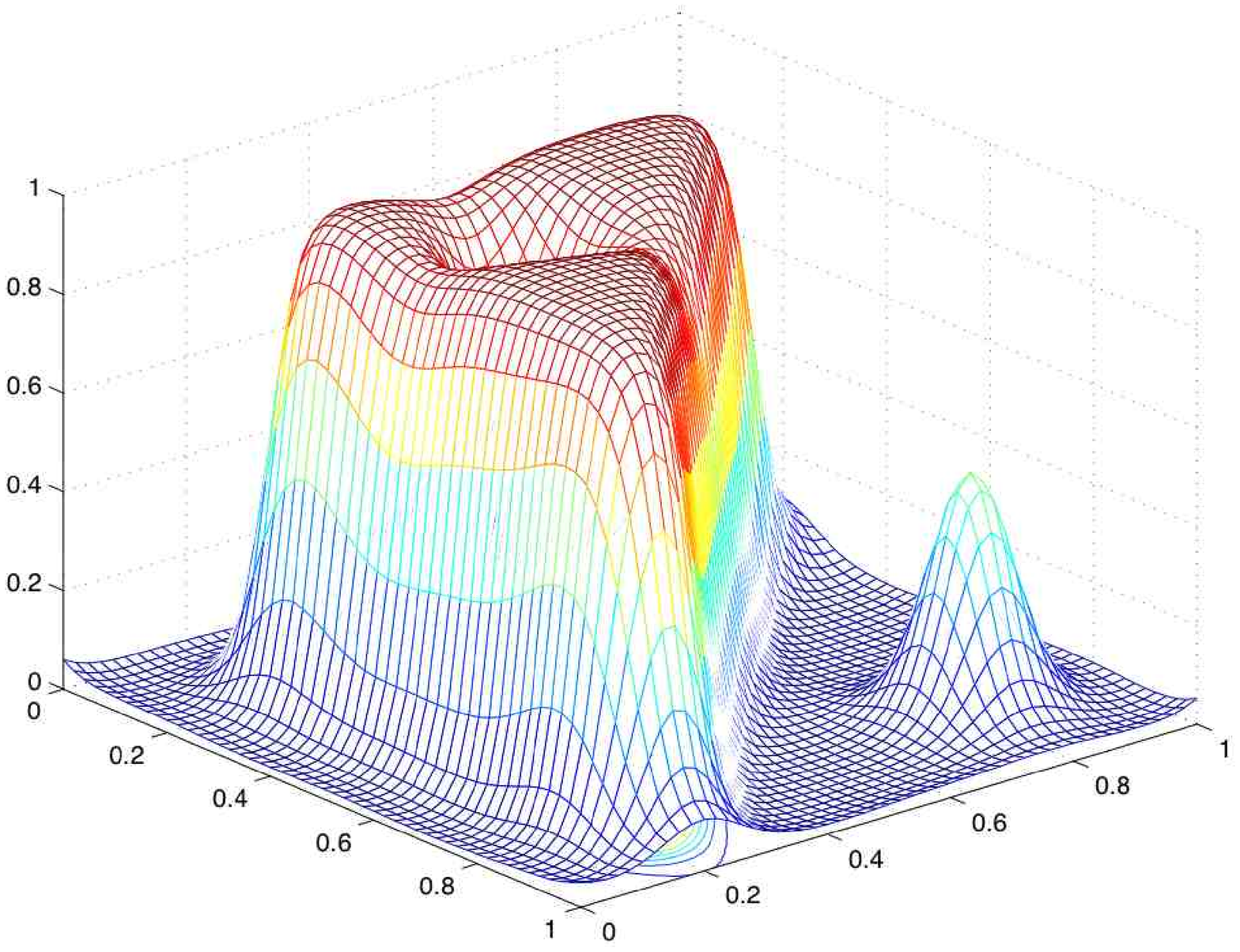}}
\subfigure[outlier explanation\label{fig:toy_gpc_outliers_explanation}]{\includegraphics[width=0.49\textwidth, keepaspectratio]{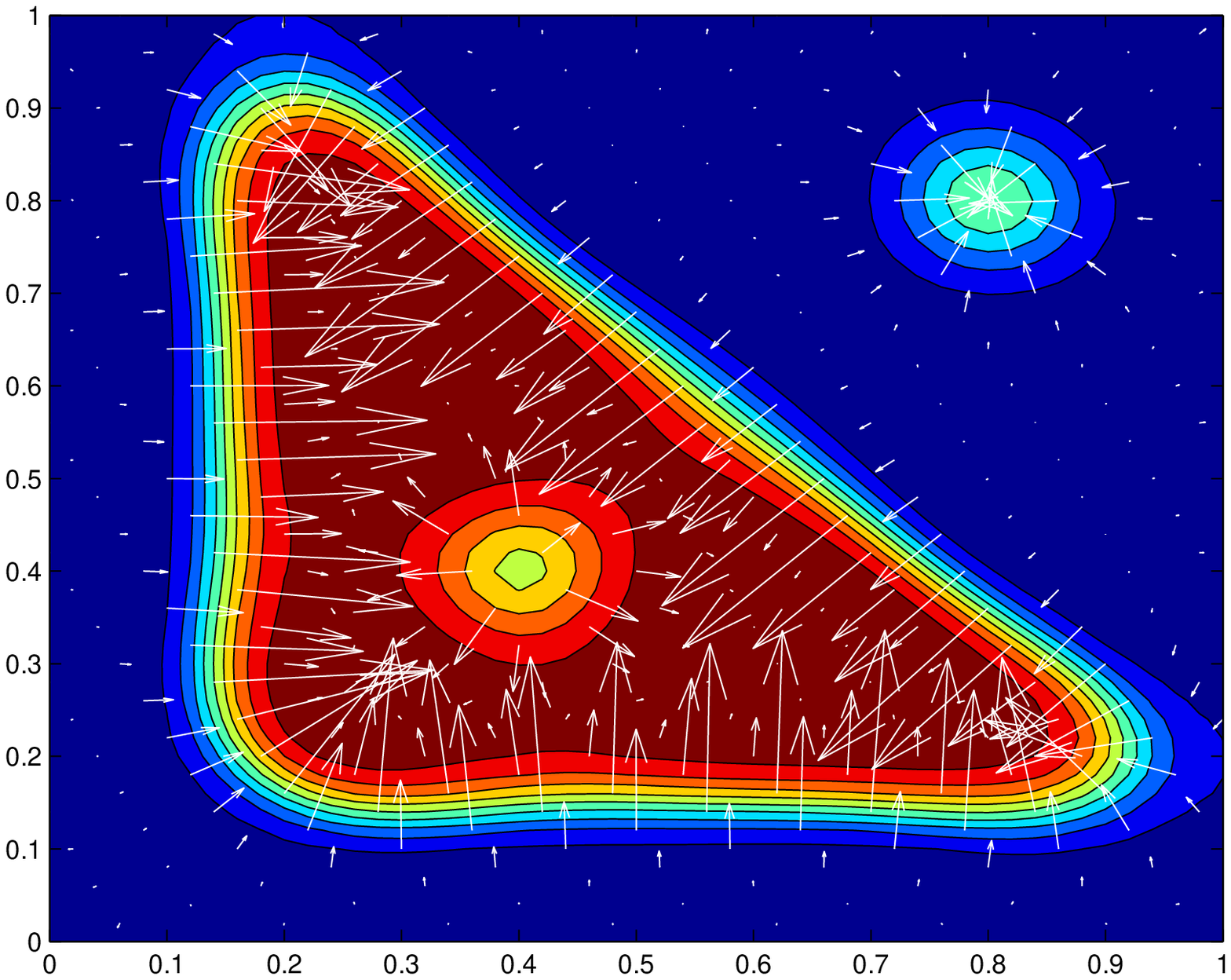}}
\caption{The effect of outliers to the local gradient explanations}
\label{fig:toy_gpc_outliers}
\end{figure}

In Figure \ref{fig:toy_gpc_outliers} the effects of two outliers in the classification data to GPC with RBF kernel are shown. Once more, note that the local gradients explain the model, which in turn may or may not capture the true situation. The size of the region affected by the outliers depends on the kernel width parameter. We consider the following items:

\begin{itemize}
\item Local gradients are in the same way sensitive to outliers as the model which they try to explain. Here a single outlier deforms the model and with it the explanation which may be extracted from it.
\item Being derivatives the sensitivity of local gradients to a nearby outlier is increased over the sensitivity of the model prediction itself.
\item Thus the local gradient of a point near an outlier may not reflect a true explanation of the features important in reality. Nevertheless it is the model here which is wrong around an outlier in the first place.
\item The histograms in the Figures \ref{fig:tox_gpc_all_tox}, \ref{fig:tox_gpc_all_detox}, and \ref{fig:tox_gpc_steroid} in Section \ref{sec:tox_gpc} show the trends of the respective features in the distribution of all test points and are thus not affected by single outliers.
\end{itemize}

To compensate for the effect of outliers to the local gradients of points in the affected region we propose to use a sliding window method to smooth the gradients around each point of interest. Thus for each point use the mean of all local gradients in the hypercube centered at this point and of appropriate size. This way the disrupting effect of an outlier is averaged out for an appropriately chosen window size.

\subsubsection{Local non-linearity}
\label{sec:appendix_direct_nonlinear}
\begin{figure}[ht]
\centering
\subfigure[locally non-linear object\label{fig:toy_gpc_nonlinear_object}]{\includegraphics[width=0.49\textwidth, keepaspectratio]{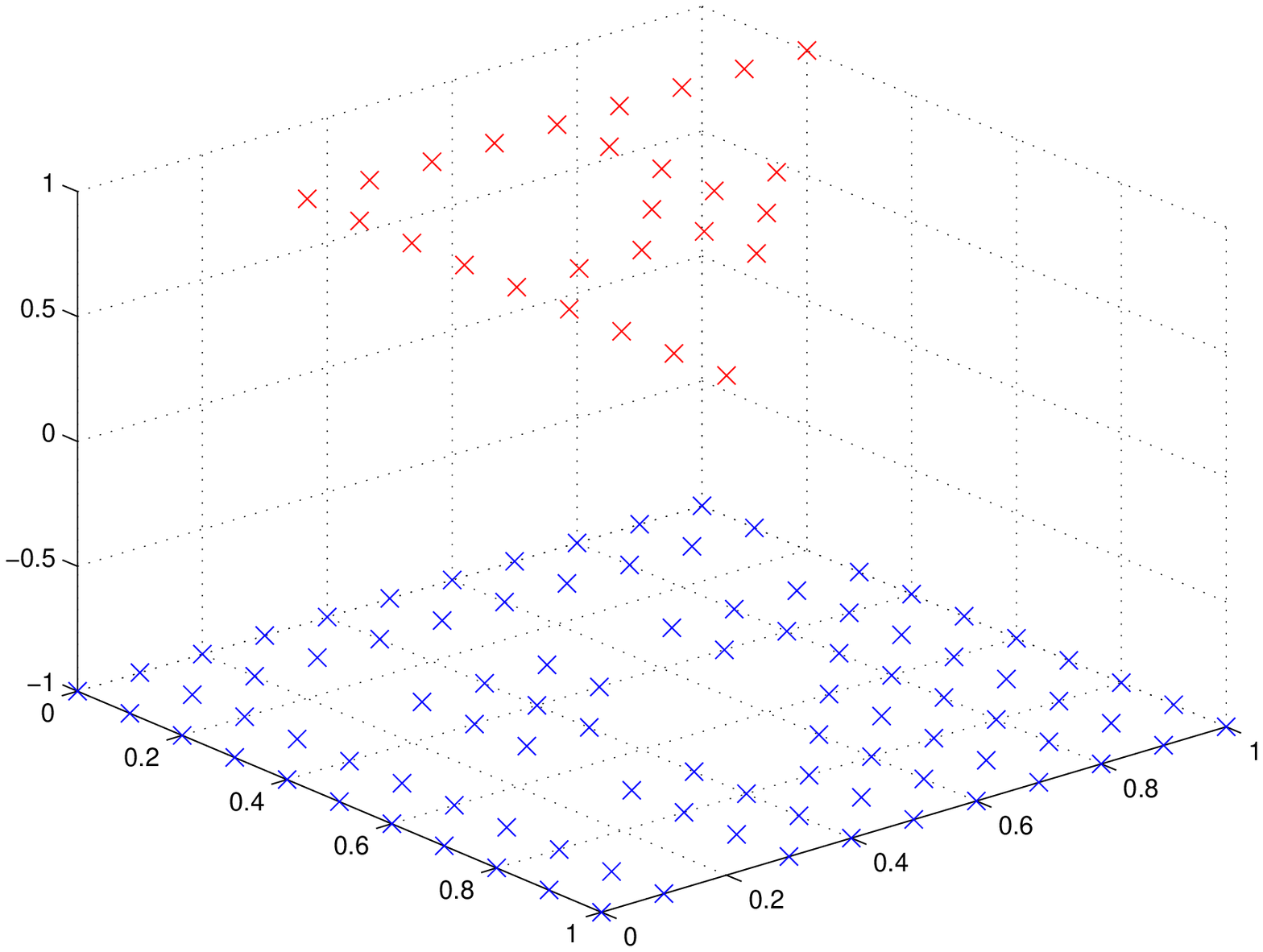}}
\subfigure[locally non-linear model\label{fig:toy_gpc_nonlinear_model}]{\includegraphics[width=0.49\textwidth, keepaspectratio]{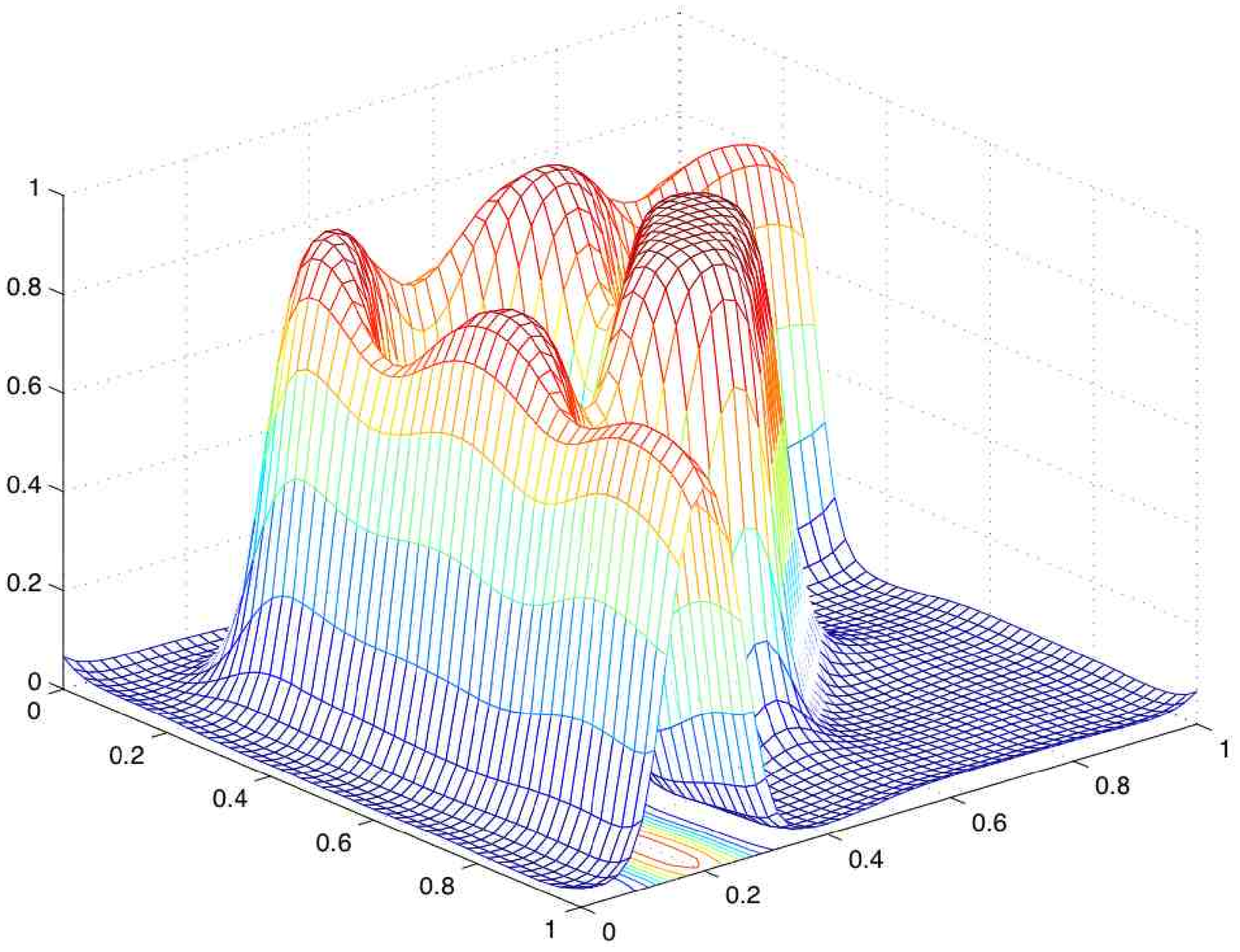}}
\subfigure[locally non-linear explanation\label{fig:toy_gpc_nonlinear_explanation}]{\includegraphics[width=0.49\textwidth, keepaspectratio]{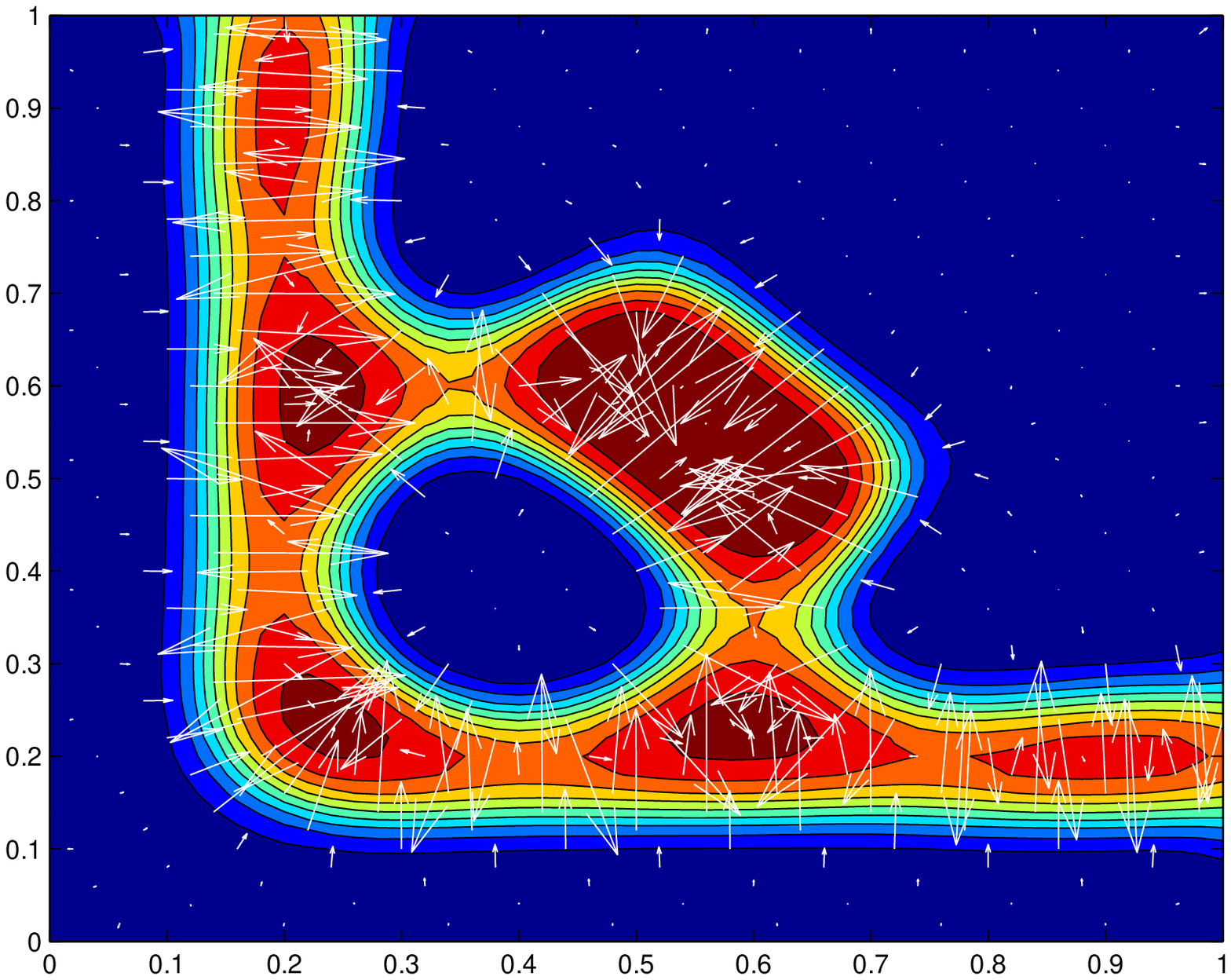}}
\caption{The effect of local non-linearity to the local gradient explanations}
\label{fig:toy_gpc_nonlinear}
\end{figure}

The effect of locally non-linear class boundaries in the data is shown in Figure \ref{fig:toy_gpc_nonlinear} again for GPC with an RBF kernel. The following points can be observed:
\begin{itemize}
\item All the non-linear class boundaries are accurately followed by the local gradients
\item The circle shaped region of negative examples surrounded by positive ones shows the full range of feature interactions towards the positive class
\item On the ridge of single positive instances the model introduces small valleys which are reflected by the local gradients
\end{itemize}

\subsection{Estimating by Parzen window}
\label{sec:appendix_estimating}
Finally we elaborate on some details of our estimation approach of local gradients by Parzen window approximation. First we give the derivation to obtain the explanation vector and second we examine how the explanation varies with the goodness of fit of the Parzen window method.

\subsubsection{Derivation of explanation vectors}
\label{sec:appendix_estimating_derivation}
These are more details on the derivation of Eq.~(\ref{eq:17}). We use the index set $I_c = \{i~|~g(x_i) = c\}$:
\begin{align*}
  &\frac{\partial}{\partial x} k_\sigma(x) = -\frac{x}{\sigma^2}k_\sigma(x) \\
  &\frac{\partial}{\partial x} \hat{p}_\sigma(x, y\neq c)
  =  \frac{1}{n}\sum_{i \notin I_c} k_\sigma(x-x_i)\frac{-(x-x_i)}{\sigma^2}\\
  &\frac{\partial}{\partial x} \hat{p}_\sigma(y\neq c|x) \\
  &= \frac{\Big(\sum_{i \notin I_c}k(x-x_i)\Big)\Big(\sum_{i=1}^n k(x-x_i)(x-x_i)\Big)}%
  {\sigma^2 \;\big(\sum_{i=1}^n k(z-x_i)\big)^2}\\
  &-\frac{\Big(\sum_{i \notin I_c}k(x-x_i) (x-x_i)\Big)\Big(\sum_{i=1}^n k(x-x_i)\Big)}%
  {\sigma^2 \;\big(\sum_{i=1}^n k(z-x_i)\big)^2}\\
  &= \frac{\Big(\sum_{i \notin I_c}k(x-x_i)\Big)\Big(\sum_{i \in I_c} k(x-x_i)(x-x_i)\Big)}
  {\sigma^2 \;\big(\sum_{i=1}^n k(z-x_i)\big)^2}\\
  &-\frac{\Big(\sum_{i \notin I_c}k(x-x_i) (x-x_i)\Big)\Big(\sum_{i \in I_c} k(x-x_i)\Big)}
  {\sigma^2 \;\big(\sum_{i=1}^n k(z-x_i)\big)^2}
\end{align*}
and thus for the index set $I_{g(z)} = \{i~|~g(x_i) = g(z)\}$
\begin{align*}
  &\hat{\zeta}(z) = \left.\frac{\partial}{\partial x}
    \;\hat{p}(y\neq g(z)\given x)\right\rvert_{x=z}\\
  &= \frac{\Big(\sum_{i \notin I_{g(z)}}k(z-x_i)\Big)\Big(\sum_{i \in I_{g(z)}} k(z-x_i)(z-x_i)\Big)}%
  {\sigma^2 \;\big(\sum_{i=1}^n k(z-x_i)\big)^2}\\
  &-\frac{\Big(\sum_{i \notin I_{g(z)}}k(z-x_i) (z-x_i)\Big)\Big(\sum_{i \in I_{g(z)}} k(z-x_i)\Big)}
  {\sigma^2 \;\big(\sum_{i=1}^n k(z-x_i)\big)^2}
\end{align*}

\subsubsection{Goodness of fit by Parzen window}
\label{sec:appendix_estimating_fit}
In our estimation framework the quality of the local gradients depends on the approximation of the classifier we want to explain by Parzen windows for which we can calculate the explanation vectors as given by Definition \ref{eq:17}.

\begin{figure}[ht]
\centering
\subfigure[SVM model\label{fig:toy_svm_parzen_model}]{\includegraphics[width=0.49\textwidth, keepaspectratio]{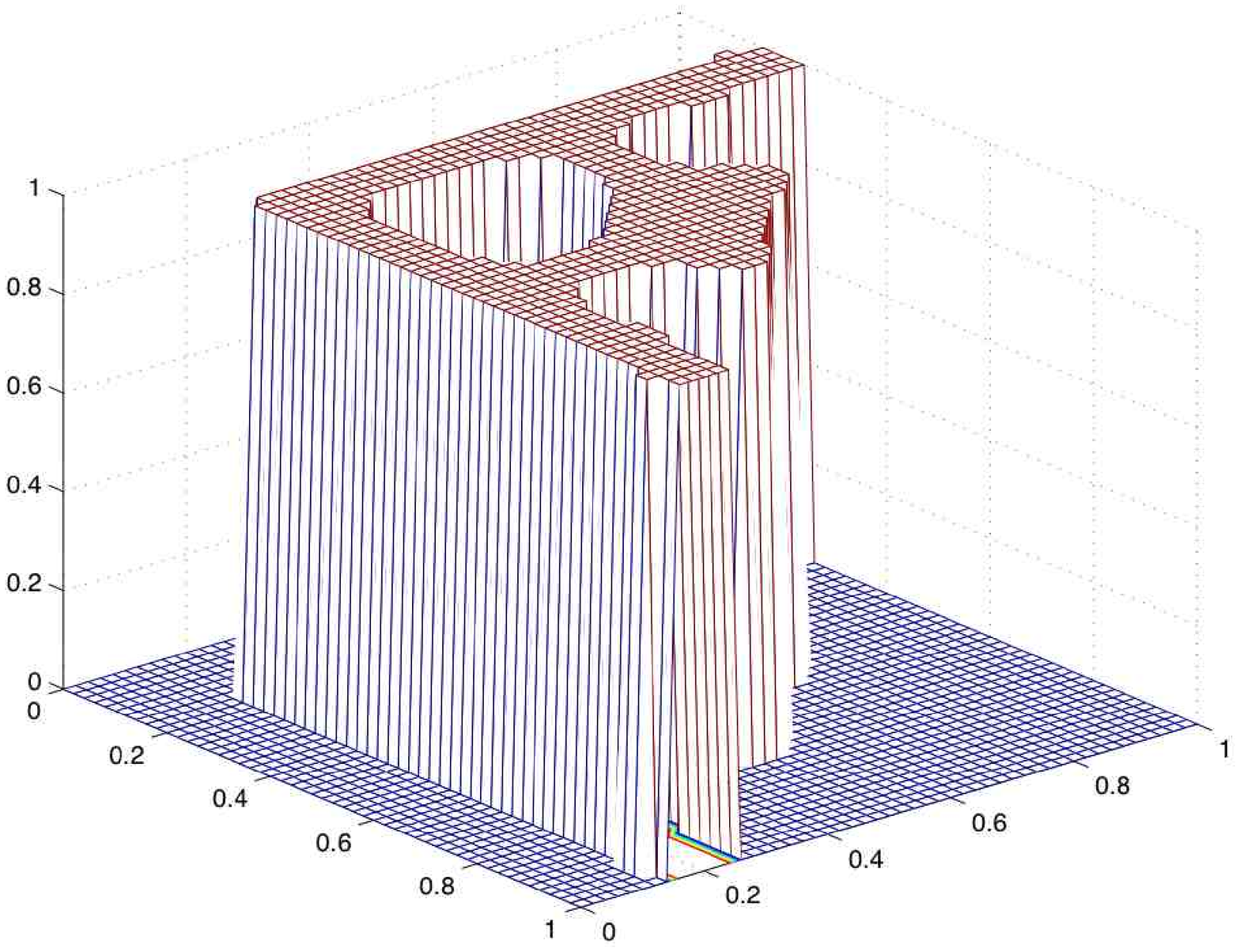}}
\subfigure[estimated explanation with $\sigma = 0.00069$\label{fig:toy_svm_parzen_explanation_local}]{\includegraphics[width=0.49\textwidth, keepaspectratio]{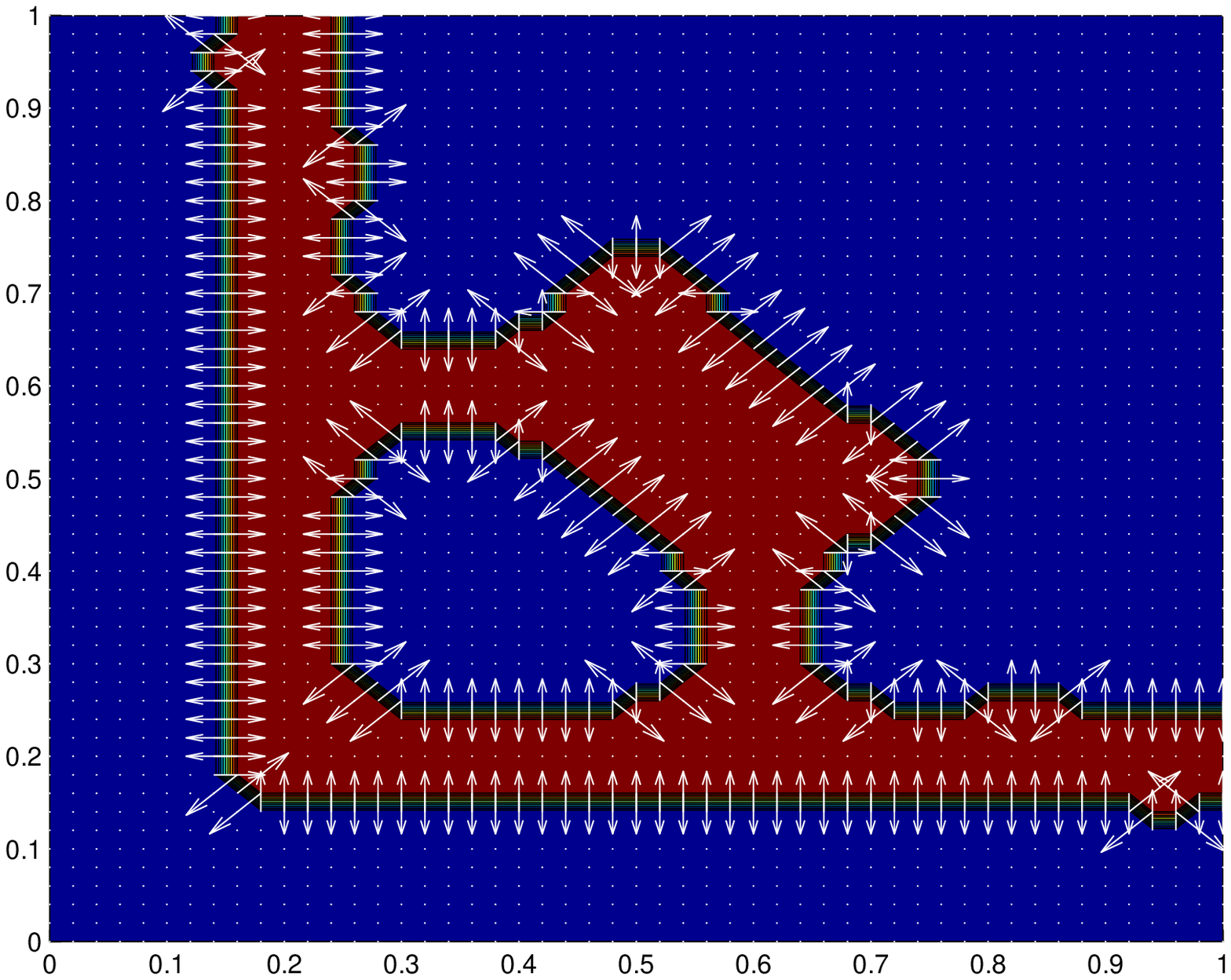}}
\subfigure[estimated explanation with $\sigma = 0.1$\label{fig:toy_svm_parzen_explanation_best}]{\includegraphics[width=0.49\textwidth, keepaspectratio]{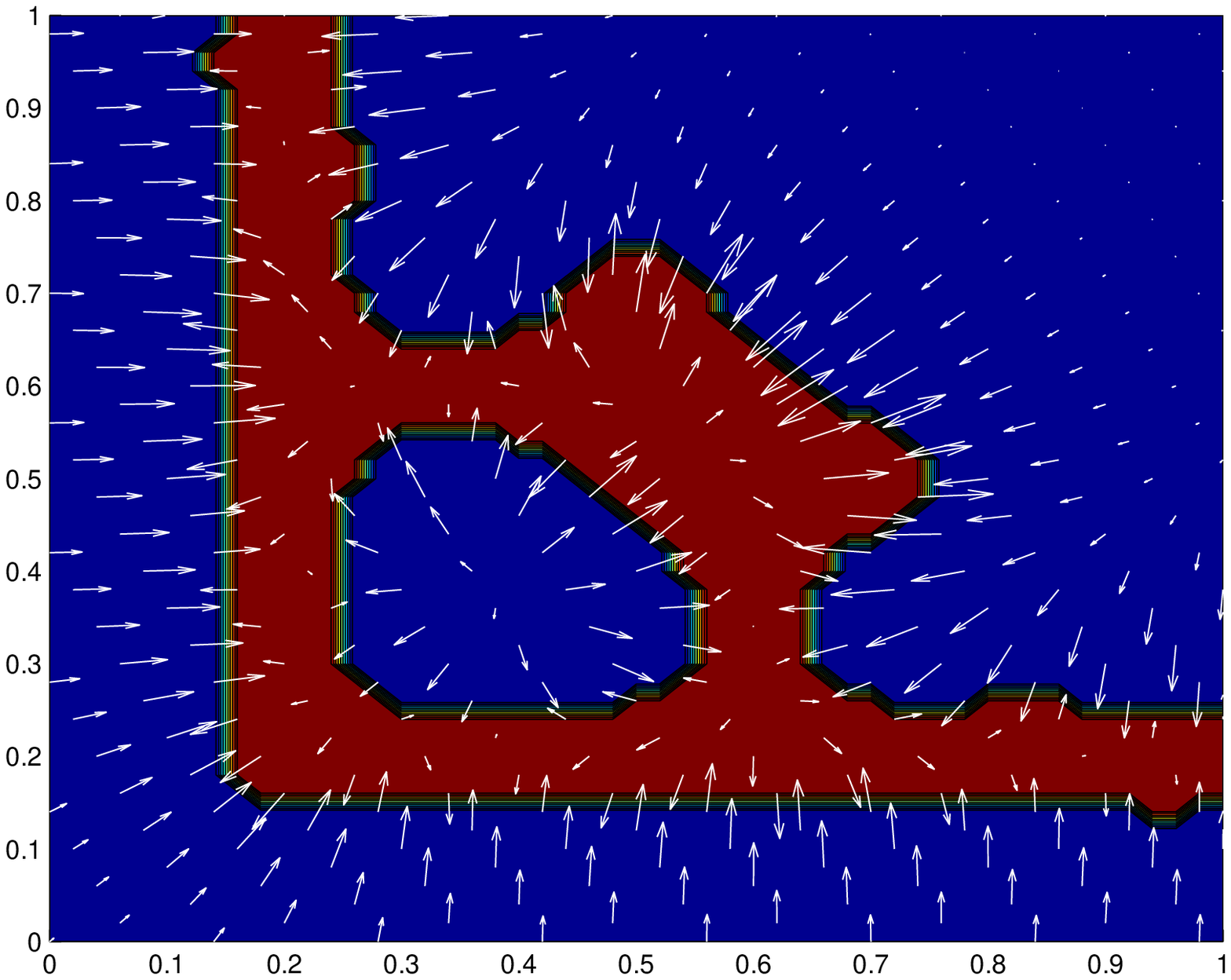}}
\subfigure[estimated explanation with $\sigma = 1.0$\label{fig:toy_svm_parzen_explanation_global}]{\includegraphics[width=0.49\textwidth, keepaspectratio]{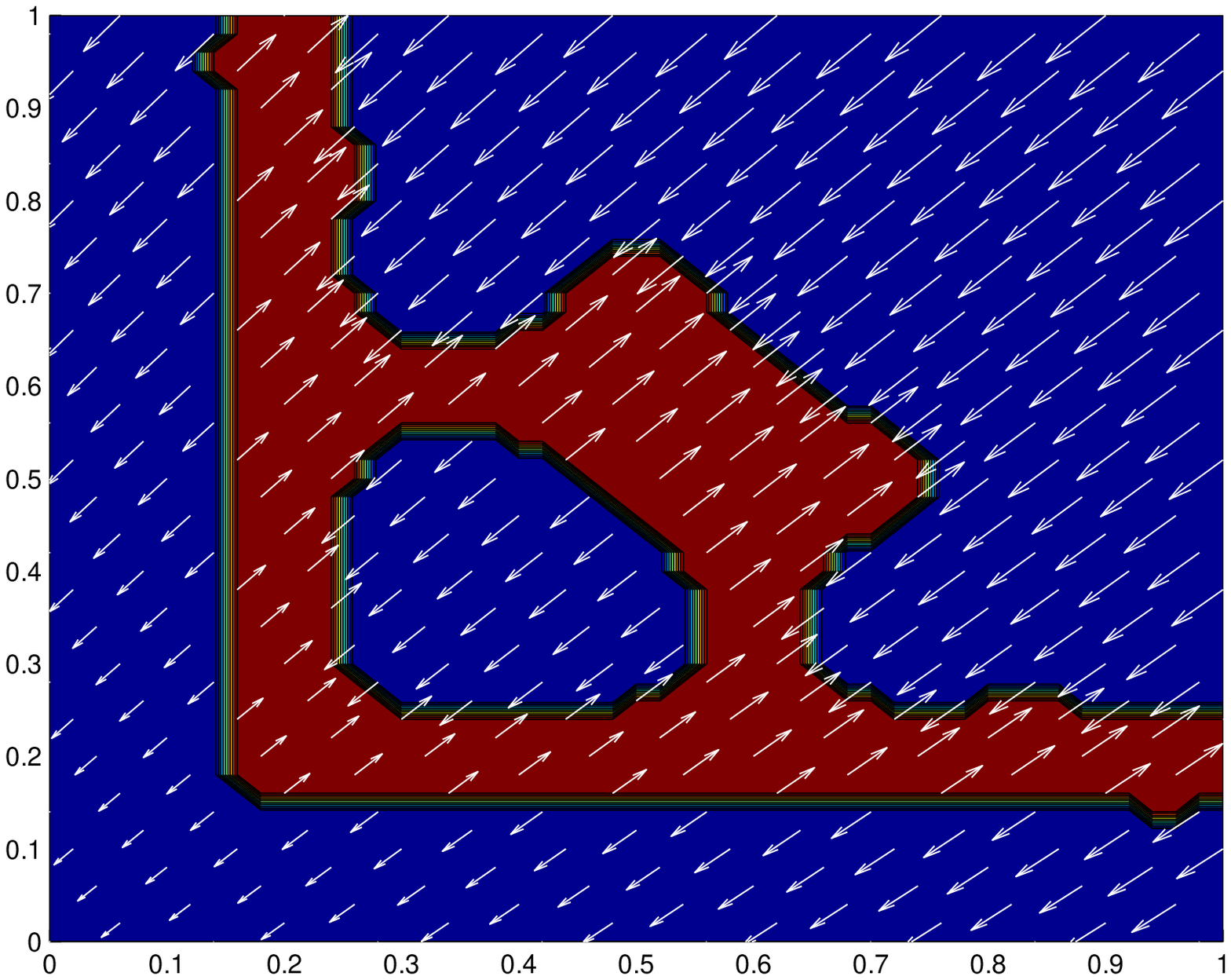}}
\caption{Good fit of Parzen window approximation affects the quality of the estimated explanation vectors}
\label{fig:toy_svm_parzen}
\end{figure}

Figure \ref{fig:toy_svm_parzen_model} shows an SVM model trained on the classification data from Figure \ref{fig:toy_gpc_nonlinear_object}. The local gradients estimated for this model by different Parzen window approximations are depicted in Subfigures \ref{fig:toy_svm_parzen_explanation_local}, \ref{fig:toy_svm_parzen_explanation_best}, and \ref{fig:toy_svm_parzen_explanation_global}. We observe the following points:

\begin{itemize}
\item The SVM model was trained with $C = 10$ and using an RBF kernel of width $\sigma = 0.01$
\item In Subfigure \ref{fig:toy_svm_parzen_explanation_local} a small window width has been chosen by minimizing the mean absolute error over the validation set of labels predicted by the SVM classifier. Thus we obtain explaining local gradients on the class boundaries but zero vectors in the inner class regions. While this resembles the piecewise flat SVM model most accurately it may be more useful practically to choose a larger width to obtain non-zero gradients pointing to the borders in this regions as well. For a more detailed discussion of zero gradients see Section \ref{sec:discussion}.
\item A larger width practically useful in this example is shown in Subfigure \ref{fig:toy_svm_parzen_explanation_best}. Here the local gradients in the inner class regions point to the other class as well.
\item For a too large window width in Subfigure \ref{fig:toy_svm_parzen_explanation_global} the approximation fails to obtain local gradients which closely follow the model. Here only two directions are left and the gradients for the blue class on the left and on the bottom point in the wrong direction.
\end{itemize}